\documentclass[final]{cvpr}

\usepackage{times}
\usepackage{epsfig}
\usepackage{graphicx}
\usepackage{amsmath}
\usepackage{amssymb}
\usepackage{dsfont}
\usepackage{subfigure}

\usepackage[table,xcdraw]{xcolor}
\usepackage{colortbl}

\usepackage{multirow}
\usepackage{arydshln}

\usepackage[pagebackref=true,breaklinks=true,colorlinks,bookmarks=false]{hyperref}
\newcommand{\revision}[1]{\textcolor{black}{#1}}
\newcommand{\revised}[1]{\textcolor{black}{#1}}
\newcommand{\rev}[1]{\textcolor{black}{#1}}

\def\cvprPaperID{1906} 
\def\confYear{CVPR 2021}

\begin{document}

\title{Continual Semantic Segmentation via Repulsion-Attraction of Sparse and Disentangled Latent Representations}

\author{Umberto Michieli and Pietro Zanuttigh\\
University of Padova, Department of Information Engineering\\
{\tt\small \{umberto.michieli, zanuttigh\}@dei.unipd.it}
}

\maketitle

\begin{abstract}
Deep neural networks suffer from the major limitation of catastrophic forgetting old tasks when learning new ones. In this paper we focus on class incremental continual learning in semantic segmentation, where new categories are made available over time while previous training data is not retained. The proposed continual learning scheme shapes the latent space to reduce forgetting whilst improving the recognition of novel classes. Our framework is driven by three novel components which we also combine on top of existing techniques effortlessly. First, prototypes matching enforces latent space consistency on old classes, constraining the encoder to produce similar latent representation for previously seen classes in the subsequent steps. Second, features sparsification allows to make room in the latent space to accommodate novel classes. Finally, contrastive learning is employed to cluster features according to their semantics while tearing apart those of different classes. Extensive evaluation on the Pascal VOC2012 and ADE20K datasets demonstrates the effectiveness of our approach, significantly outperforming state-of-the-art methods.
\end{abstract}


\section{Introduction}
\label{sec:introduction}

Semantic segmentation is a  challenging computer vision  problem 
with many real-world applications ranging from robot sensing, to autonomous driving, video surveillance, virtual reality, and many others. For most applications, continuously improving the set of classes to be distinguished is a fundamental requirement.
 Current state-of-the-art semantic segmentation approaches are typically based on auto-encoder structures and on fully convolutional models \cite{long2015fully} that are trained in a single-shot requiring all the dataset to be available at once. Indeed, existing architectures are not designed to incrementally update their inner classification model to accommodate new categories. This issue is well-known for deep neural networks and it is called \textit{catastrophic forgetting} \cite{mccloskey1989catastrophic,french1999catastrophic,goodfellow2013empirical}, as deep architectures fail to update their parameters for learning new categories while preserving good performance on the old ones.

\begin{figure}
\centering
\includegraphics[trim={0cm 12cm 19.65cm 0cm}, clip, width=\linewidth]{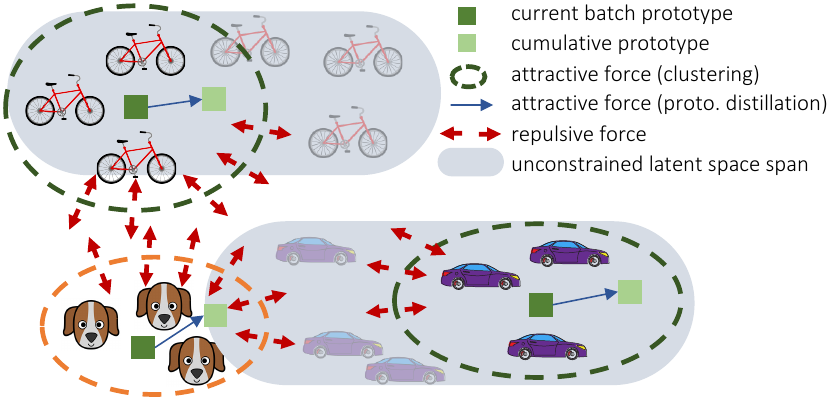}
\caption{\revised{Our} continual learning scheme is driven by 3 main  components: latent contrastive learning, prototypes matching and features sparsity. Latent representations of old classes are preserved via prototypes matching and clustering, whilst also making room for accommodating new classes via sparsity and repulsive force of contrastive learning. The decoder preserves previous knowledge via output-level distillation. In the figure, bike and cars represent old classes and leave more space to new classes (the dog) thanks to the novel constraints (green dotted ovals versus gray-filled ovals).}
\label{fig:graphical_abstract}
\end{figure}

\revised{Continual learning has been widely studied in image classification} \cite{kirkpatrick2017overcoming,li2018learning} and object detection \cite{shmelkov2017incremental,li2019rilod}, while has been tackled only recently in the semantic segmentation field \cite{michieli2019incremental,tasar2018incremental,cermelli2020modeling,klingner2020class}. 
\revised{In this paper, we investigate class-incremental continual learning in semantic segmentation.} Differently from the majority of previous approaches both in image classification \cite{li2018learning,rebuffi2017icarl,castro2018end} and semantic segmentation \cite{tasar2018incremental,michieli2019incremental,cermelli2020modeling,klingner2020class}, we do not mainly or solely rely on output-level knowledge distillation. 
In this work, we focus on latent space organization which has been only marginally investigated in the current literature, and we empirically prove it to be complementary to other existing techniques. The main idea is depicted in Fig.~\ref{fig:graphical_abstract}, where some of the latent space constraints are introduced. 
First, a prototype matching is devised to enforce features extraction consistency on old classes between the cumulative prototype  computed using all previous samples and the current prototype (\ie, the prototype computed on the current batch only). In other words, we force the encoder to produce similar latent representations for previously seen classes in the new steps. 
Second, a features sparsification constraint makes room in the latent space to accommodate novel classes. To further regularize the latent space, we introduce an attraction-repulsion rule similar in spirit to the recent advancements in contrastive learning. Finally, to enforce the decoder to preserve discriminability on previous categories during classification, we employ a targeted output-level distillation.


Although continual semantic segmentation has only been faced recently, it already comes with different experimental protocols depending on how the incremental data are considered (see Section~\ref{subsec:protocols}): namely, \textit{sequential} (new images are labeled with both new and old classes), \textit{disjoint} (new images are labeled with only new classes, old classes are assigned to the background) and \textit{overlapped} (new images are labeled with only new classes, images are repeated across training steps with different semantic maps associated to them). In this paper we devise a common framework which allows to tackle all these scenarios and can be applied in combination with previous techniques, which has never been attempted before. We evaluate on standard semantic segmentation datasets, like Pascal VOC2012 \cite{pascalvoc2012} and ADE20K \cite{zhou2017scene}, in many scenarios. 

Summing up, the main contributions of this work are: 1) \revised{We investigate class-incremental learning in semantic segmentation, providing a common framework for different experimental protocols. 2) We explore the latent space organization and we propose complementary techniques with respect to the existing ones.} 3) We propose novel knowledge preservation techniques based on prototypes matching, contrastive learning and features sparsity. 4) We benchmark our approach on standard semantic segmentation datasets 
 outperforming state-of-the-art continual learning methods.

\section{Related Work}
\label{sec:related}

\textbf{Continual Learning.} 
Deep learning models are prone to \textit{catastrophic forgetting} \cite{goodfellow2013empirical,kemker20178measuring,parisi2019continual}, \ie, training a model with new information interferes with previously learned knowledge and typically greatly degrades performance. 
This phenomenon has been widely studied in image classification task and most of the current techniques fall into the following categories \cite{delange2019continual,parisi2019continual}: regularization approaches \cite{chaudhry2018riemannian,kirkpatrick2017overcoming,zenke2017continual,dhar2018learning,li2018learning}, dynamic architectures \cite{xiao2014error,wang2017growing,li2019learn}, parameter isolation \cite{fernando2017pathnet,serra2018overcoming,mallya2018packnet} and replay-based methods \cite{wu2018incremental,ostapenko2019learning,shin2017continual,hou2019learning}. 
Regularization-based approaches are by far the most widely employed and mainly come in two flavours, \ie, penalty computing and knowledge distillation \cite{hinton2015distilling}. Penalty computing approaches \cite{zenke2017continual,kirkpatrick2017overcoming,kirkpatrick2017overcoming} protect important weights inside the models to prevent forgetting. Knowledge distillation \cite{schwarz2018progress,wu2018incremental,li2018learning,dhar2018learning} relies on a teacher (old) model transferring or remembering knowledge related to previous tasks to a student model which is trained to learn also additional tasks.
Parameter isolation approaches \cite{mallya2018packnet,mallya2018piggyback} reserve a subset of weights for a specific task to avoid degradation. 
Dynamic architectures \cite{wang2017growing,li2019learn} grow new branches for new tasks. 
Replay-based models exploit stored \cite{castro2018end,hou2019learning,rebuffi2017icarl} or generated  \cite{wu2018incremental,ostapenko2019learning,shin2017continual} examples during the learning process of new tasks.

\textbf{Continual Semantic Segmentation.}
Nowadays, deep learning architectures have achieved outstanding results in semantic segmentation \cite{garcia2018survey,guo2018review}. Current approaches are based on fully convolutional models \cite{long2015fully,chen2017rethinking,chen2018deeplab,zhao2017pyramid,yu2017dilated} and exploit various techniques to cope with multi-scale and spatial dependency. All these approaches, however, require training data and segmentation maps to be available at once (\ie, \textit{joint} setting) and they experience catastrophic forgetting if new tasks (\eg, new classes to learn) are made available sequentially \cite{michieli2019incremental}.
Hence, it emerged the need for continual approaches specifically targeted to solve the semantic segmentation task \cite{ozdemir2019extending,tasar2018incremental,michieli2019incremental,michieli2020knowledge,klingner2020class,cermelli2020modeling}. Earlier works focus on the continual semantic segmentation problem in specific scenarios, \eg, in medical imaging \cite{ozdemir2019extending} or remote sensing \cite{tasar2018incremental}, extending standard image-level classification methods. More recently, standard semantic segmentation datasets and targeted methods have been proposed. In \cite{michieli2019incremental,michieli2020knowledge} an exploration on knowledge distillation techniques is proposed to alleviate forgetting: the authors designed output-level and features-level distillation losses coupled with 
freezing the encoder's weights. 
Klingner \etal \cite{klingner2020class} extend previous work not requiring old labels during the incremental steps and proposing class importance weighting to emphasize gradients on difficult classes.
Cermelli \etal \cite{cermelli2020modeling} study the distribution shift of the background class when it incorporates previous and/or future classes (\textit{disjoint} and \textit{overlapped} protocols, respectively). \revised{Background shift is addressed via unbiased versions of cross entropy and output-level knowledge distillation losses together with an unbiased weight initialization rule for the classifier.}
\revised{Nevertheless, previous works neglect accurate investigation of the latent space in continual learning.}

\textbf{Latent Space Organization.}
\revised{The analysis of the latent space organization is becoming crucial towards understanding and improvement of classification models \cite{xian2016latent,peng2019domain}.
Recently, some attention has been devoted to latent regularization in continual image classification} \cite{achille2018life,aljundi2018selfless,javed2019meta}.\\
Besides this, one of the emerging paradigms is constrastive learning applied to visual representations. 
Dating back to \cite{hadsell2006dimensionality}, these approaches learn representations by contrasting positive against negative pairs and have been recently re-discovered for deep learning. \revised{Many works use a memory bank to store the instance class representation vector \cite{wu2018unsupervised,zhuang2019local,tian2019contrastive,he2020momentum,misra2020self,chen2020simple}, while some others explore the usage of in-batch negative samples instead} \cite{doersch2017multi,ye2019unsupervised,ji2019invariant,khosla2020supervised}. The contrastive learning objective proposed in this work moves from opposition of positive and negative pairs and also recalls features clustering (if features belong to the same class) and separation (if features belong to different classes), \revised{which has been recently applied to adapt semantic segmentation models across domains \cite{kang2019contrastive,liang2019distant,toldo2021unsupervised}.}\\
Prototypes-based regularizing terms gained a great interest and, in particular, have been largely used in the literature of few-shot learning \cite{dong2018few,wang2019panet,tian2020generalized}, to learn prototypical representations of each category, and domain adaptation, to enforce orthogonality \cite{pinheiro2018unsupervised,wu2019improving} or centroid matching \cite{xie2018learning,deng2019cluster}.\\
Finally, to minimize the interference among features we drive them to be channel-wise sparse. Only limited attention has been given on sparsity for deep learning architectures \cite{aljundi2018selfless}; however, some prior techniques exist for domain adaptation on linear models exploiting sparse codes on a shared dictionary between the domains \cite{shekhar2013generalized,zhang2015domain}.

\revised{Our work is the first combining together contrastive learning, sparsity and prototypes matching to regularize latent space for segmenting new categories over time.}

\section{Problem Definition and Setups}
\label{sec:problem}


Before presenting the proposed strategies, we first introduce the semantic segmentation task, which assigns a class to each pixel in an image.
We denote the input image space with $\mathcal{X} \in \mathbb{R}^{H\times W \times 3}$ with spatial dimensions $H$ and $W$, the set of classes (or categories) with $\mathcal{C}=\left\lbrace c_i \right\rbrace_{i=0}^{C-1}$ and the output space with $\mathcal{Y} \in \mathcal{C}^{H \times W}$ (\ie, the segmentation map). 
Given a training set $\mathcal{T} = \left\lbrace \left( \mathbf{x}_n , \mathbf{y}_n \right)  \right\rbrace_{n=1}^N$, where $\left( \mathbf{x}_n , \mathbf{y}_n \right) \in \mathcal{X} \times \mathcal{Y}$, we aim at finding a map $M$ from the input space to a pixel-wise class probability vector $M: \mathcal{X} \mapsto \mathbb{R}^{H\times W \times C}$. Then, the output segmentation mask is computed as $\hat{\mathbf{y}}_n = \mathrm{arg}\max_{c\in \mathcal{C}} M(\mathbf{x}_n)[h,w,c]$, where $h=1,..,H$, $w=1,...,W$ and  $M (\mathbf{x}_n)[h,w,c]$ is the probability for class $c$ in pixel $(h,w)$. Nowadays, $M$ is typically some auto-encoder model made by an encoder $E$ and a decoder $D$ (\ie, $M = E \circ D$). \revision{We call $\mathbf{F}_n = E(\mathbf{x}_n)$ the feature map of $\mathbf{x}_n$, and $\mathbf{y}^*_n$ the downsampled segmentation map matching the spatial dimensions of $\mathbf{F}_n$.}

In the standard supervised setting it is assumed that the training set $\mathcal{T}$ is available at once and the model is learned in one shot. In the continual learning scenario, instead, training is achieved over multiple iterations each carrying a novel category to learn and a subset of the training data. More formally, at each learning step $k$ the previous label set $\mathcal{C}_{k-1}$ is expanded with a set of novel classes $\mathcal{S}_k$ forming a new label set $\mathcal{C}_k =\mathcal{C}_{k-1} \cup \mathcal{S}_k$. Additionally, a new training subset $\mathcal{T}_k \subset \mathcal{X} \times \mathcal{C}_k$ is made available and used to update the previous model into a new model $M_{k}$. Step $k=0$ consists of a standard supervised training performed with only a subset of training data and classes. As in the standard incremental class learning scenario, we assume the different sets of new classes to be disjoint with the exception of the peculiar background class $c_0$, \ie, $\mathcal{S}_i \cap \mathcal{S}_j = \left\lbrace c_0 \right\rbrace$.

\subsection{Experimental Protocols}
\label{subsec:protocols}
Despite being quite a recent field, continual learning in semantic segmentation already comes in different flavors.
\textbf{Sequential:} this setup has been proposed in \cite{michieli2019incremental,michieli2020knowledge}. Each learning step contains a unique set of images, whose pixels belong to classes seen either in the current or in the previous learning steps. At each step, labels for pixels of both old and novel classes are present.\\
\textbf{Disjoint:} this setup has been proposed in \cite{cermelli2020modeling}. At each learning step, the unique set of images is identical to the \textit{sequential} setup. The difference with respect to the \textit{sequential} setup lies in the set of labels. At each step, only labels for pixels of novel classes are present, while the old ones are labeled as background in the ground truth.\\
\textbf{Overlapped:} this setup moves from the work of \cite{shmelkov2017incremental} for object detection and has been adapted to   semantic segmentation in \cite{cermelli2020modeling}. Each training step contains all the images that have at least one pixel of a novel class, with only the novel classes annotated while the rest is set to background. Differently from the other settings, here images may contain pixels of classes that will be learned in  future learning steps, but they are labeled as background in the current step. 
\section{Method}
\label{sec:contSS}

\begin{figure*}[ht]
\centering
\includegraphics[trim={0cm 7.8cm 5cm 0cm}, clip, width=0.95\textwidth]{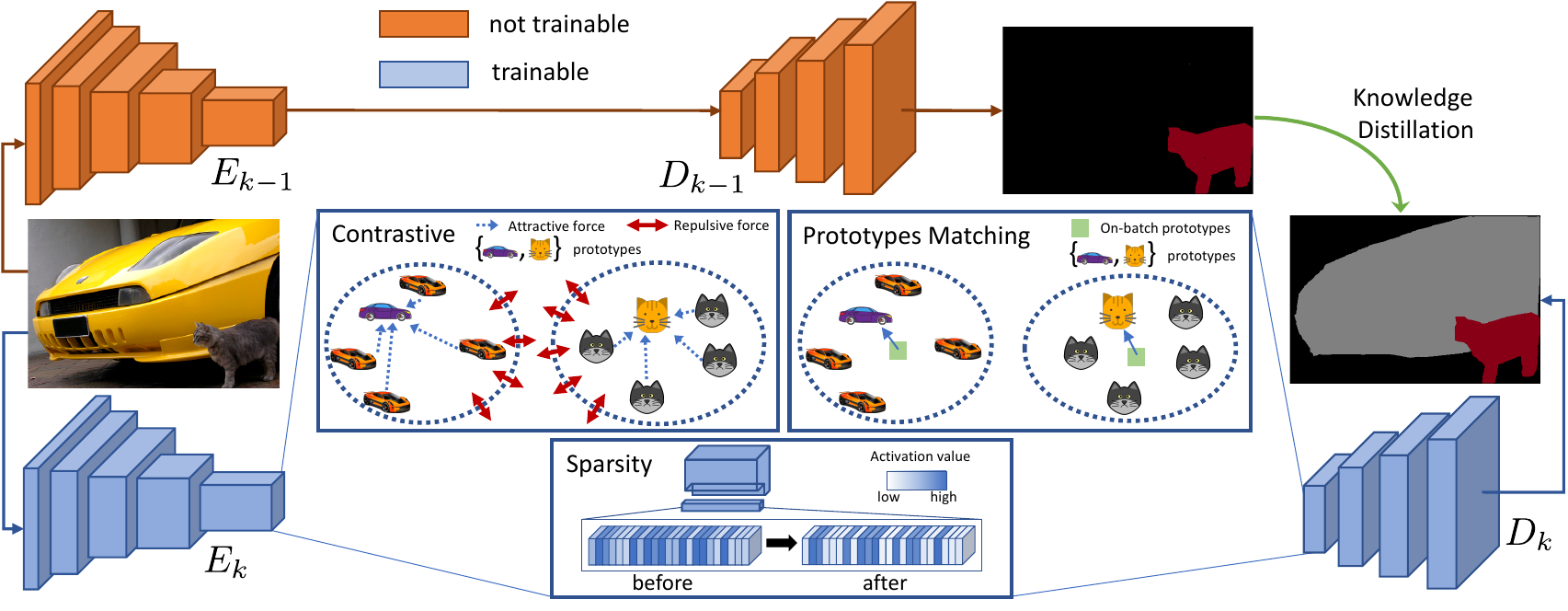}
\caption{Overview of the proposed approach, with an old class (\textit{cat}) and a new class (\textit{car}). Latent representations of old classes are preserved over time via prototypes matching and clustering, whilst also making room for accommodating new classes via sparsity and repulsive force in contrastive learning. The decoder is constrained to act as in previous steps on previous classes via output-level distillation.}
\label{fig:architecture}
\end{figure*}
In this section, we provide a detailed description of the core modules of the proposed method. Our approach leverages a contrastive learning objective applied over the 
feature representations, with novel prototypes matching and sparsity constraints.
Specifically, features repulsion and attraction based on the semantic classes are enforced by grouping together features of the same class, while simultaneously pushing away those of different categories.
We further regularize the distribution of latent representations by the joint application of prototypes matching and sparsity. While prototypes matching seeks for an invariant  representation of the features extracted for the  old classes, the sparsity objective encourages a lower volume of active feature channels from latent representations (\ie, it concentrates the energy of features on few dimensions) to free up space for new classes. 

An overall scheme of our approach 
is shown in Fig.~\ref{fig:architecture}: \revised{the training objective is given by the combination of a cross-entropy loss  ($\mathcal{L}_{ce}$) with the proposed modules.  $\mathcal{L}_{ce}$ is the usual cross-entropy loss for all the classes except for the background.} The ground truth of the background, indeed, is not directly compared with its probabilities, 
but with the probability of having either an old class 
or the background in the current model \cite{cermelli2020modeling}. 
Formally, at step $k$ the background probabilities $M (\mathbf{x}_n)[h,w,c_0]$ are replaced by $\sum_{c \in \mathcal{C}_{k-1} } M (\mathbf{x}_n)[h,w,c]$. 
The rationale behind this is that the background class could incorporate statistics of previous classes in both the disjoint and overlapped protocols. 

The other components are a prototypes matching target ($\mathcal{L}_{pm}$), a contrastive learning objective ($\mathcal{L}_{cl}$) and a sparsity constraint ($\mathcal{L}_{sp}$), which will be detailed in the following sections.
The training objective is then computed as:
\begin{equation}
\mathcal{L}_{tot}' = \mathcal{L}_{ce} + \lambda_{pm} \cdot \mathcal{L}_{pm} + \lambda_{cl} \cdot \mathcal{L}_{cl} + \lambda_{sp} \cdot \mathcal{L}_{sp} 
\label{eq:tot1}
\end{equation}
%
%
%
where the $\lambda$ parameters  balance the multiple losses and have been tuned using a validation set (see Section \ref{sec:training}).
Our aim is to seek for disentangled latent representations characterized by semantic-driven regularization and to show that this approach can achieve comparable or superior results with respect to standard regularization methods (\eg, output-level knowledge distillation). We further integrate the proposed framework with an output-level knowledge distillation objective \cite{michieli2020knowledge} and we show that its effect is highly not overlapping,  achieving increased accuracy. 
The training objective comprising an unbiased output-level distillation module is defined as:
\begin{equation}
\mathcal{L}_{tot} = \mathcal{L}_{tot}' + \lambda_{kd} \cdot \mathcal{L}_{kd}
\label{eq:tot2}
\end{equation}

\subsection{Prototypes Matching}
\label{subsec:prototypes}
Prototypes (\ie, class-centroids) are vectors that are representative of each category that appears in the dataset. During training, the features extracted by the encoder 
contribute in forming the latent prototypical representation of each class. To preserve the geometrical structure of the features of old classes we apply prototypes matching. Current prototypes $\hat{\mathbf{p}}_c$ (\ie, computed on the current batch of images) are forced to be placed close to their representation learned from the previous steps $\mathbf{p}_c $. We use the Frobenius norm $||\cdot||_F$ as metric distance \cite{snell2017prototypical,oreshkin2018tadam,wang2019panet}. More formally:
\begin{equation}
\mathcal{L}_{pm} =  \frac{1}{| \mathcal{C}_{k-1}  |}
||
\mathbf{p}_c - \hat{\mathbf{p}}_c
||_F
\ \ \ \ \\  \
c \in \mathcal{C}_{k-1}
\label{eq:L_pm}
\end{equation}

The prototypes are computed in-place with a running average updated at each training step with supervision. At training step $t$ with batch $\mathcal{B}$ of $B$ images, the prototypes are updated for a generic class $c$ as:
\begin{equation}
\mathbf{p}_c \!\left[ t \right] \!=\!\!
\frac{1}{B t} \! \!
\left(\! \!
B \! \left( t\!-\!1\! \right) \!\mathbf{p}_c \!\left[ t \! - \! 1\right] \!+\!\!\!
\sum_{\mathbf{x}_n \in \mathcal{B}}
\!\!\!\frac{ \sum_{\mathbf{f}_i \in \mathbf{F}_n} \!\! \mathbf{f}_i \mathds{1}\!\left[  y_i^* \!= \!c   \right]}
{| \mathds{1}\left[  \mathbf{y}_n^* = c  \right] |} 
\!\!\right)  
\label{eq:prototypes}
\end{equation}
\revised{initialized to $\mathbf{p}_c \!\left[ 0 \right] \! = \! \mathbf{0} \, \forall c$.  $\mathbf{f}_i \! \in \! \mathbf{F}_n$  is a generic feature vector and $y_i^*$ the corresponding pixel in  $\mathbf{y}_n^*$, 
${\mathds{1}\left[  \mathbf{y}_n^* = c  \right]}$  indicates the pixels in $\mathbf{y}_n^*$ associated to $c$ and $|\cdot |$ denotes cardinality.} 
 
We update the prototypes only when we have ground truth labels for that class to avoid incorporating the mutable statistics of the background class:  we exclude the background from the incremental steps in the disjoint protocol (as it could contain old classes) and in the overlapped scenario (as it could contain old and \revised{future} classes).

For the current batch $\mathcal{B}$ of an incremental training stage, the current (or in-batch) prototypes $\hat{\mathbf{p}}_c[t]$ are computed as:
\begin{equation}
\displaystyle
\mathbf{\hat{\mathbf{p}}}_c[t] = \!\frac{1}{B} \sum_{\mathbf{x}_n \in \mathcal{B}}
\begin{cases}
 \frac{  \sum_{\mathbf{f}_i \in \mathbf{F}_n} \mathbf{f}_i \mathds{1}\left[  y_i^* = c   \right] }{|\mathds{1}\left[  \mathbf{y}_n^* = c   \right]|} \quad \mathrm{if\ sequential} \\
   \frac{  \sum_{\mathbf{f}_i \in \mathbf{F}_n} \mathbf{f}_i \mathds{1}\left[  \hat{z}_i^* = c   \right] }{|\mathds{1}\left[  \hat{\mathbf{z}}_n^* = c   \right]|} \quad \mathrm{otherwise} \\
\end{cases} 
\label{eq:prototypes_current}
\end{equation}
where $\hat{\mathbf{z}}_n^*$ (with pixels $\hat{z}_i^*$) is a \revision{pseudo-labeled} segmentation map computed from the ground truth data by replacing the background region  with the prediction from the previous model, since in the disjoint and overlapped protocols old classes are labeled as background. \revision{The difference between \eqref{eq:prototypes} and \eqref{eq:prototypes_current} lies in the usage of pseudo-labels:  we use them in \eqref{eq:prototypes_current} to compute prototypes for old classes in the current batch since we may not have any label for them, but we avoid to use them in \eqref{eq:prototypes}, since there is no need to update prototypes computed using the ground truth at previous steps with data from less reliable pseudo-labels.}

\subsection{Contrastive Learning}
\label{subsec:contrastive}

The second component is similar to recent contrastive learning \cite{chen2020simple,tian2019contrastive} and clustering \cite{kang2019contrastive,liang2019distant} approaches to constraint the latent space organization. 
The underlying idea is to structure the latent space in order to have features of the same category clustered near their prototype and at the same time to force prototypes to be far one from the other. We argue that this organization helps also in continual learning to mitigate forgetting and to facilitate the addition of novel classes, as features are clustered and there is more separation between the clusters.
In formal terms, the constraint is defined by a loss $\mathcal{L}_{cl}$ made of an attractive term $\mathcal{L}^{a}_{cl}$ and a repulsive term $\mathcal{L}^{r}_{cl}$, as follows:
\begin{align}
\mathcal{L}_{cl}^a  &= \frac{1}{|c_j \! \in \! \mathbf{y}_n^*\! |} 
\sum_{c_j \in \mathbf{y}_n^*} \sum_{\mathbf{f}_i \in \mathbf{F_n}}
|\!|  \!  \left(  \! \mathbf{f}_i \!\!  - \! \mathbf{p}_{c_j}  \!\right)  \! \mathds{1}\! \left[   y_i^* \! = \! c _j \right] \!     \!  |\!|_{\!F} 
\label{eq:attractive}
\\
\mathcal{L}_{cl}^r \! & =  \! \! 
\frac{1}{|c_j \! \in \! \mathbf{y}_n^*\! |} \! 
\sum_{c_j \in \mathbf{y}_n^*} 
\sum_{\substack{c_k \in \mathbf{y}_n^* \\ c_k \neq c_j}} \! \! \frac{1}{|\!|  \hat{\mathbf{p}}_{c_j} \! \! - \! \! \hat{\mathbf{p}}_{c_k}  |\!|_{\!F} } 
\label{eq:repulsive}
\end{align}

The  objective is composed of two terms: $\mathcal{L}^{a}_{cl}$ measures how close features are from their respective centroids and $\mathcal{L}^{r}_{cl}$ how spaced out prototypes corresponding to different semantic classes are. Hence, the effect provided by the loss minimization is twofold: firstly, feature vectors from the same class are tightened around class feature centroids; secondly, features from separate classes are subject to a repulsive force applied to feature centroids, moving them apart.

\subsection{Features Sparsity}
\label{subsec:sparsity}
To enforce the regularizing effect brought by contrastive learning, we introduce a further feature-wise objective on the latent space.
We propose a sparsity loss to decrease the number of active feature channels of latent vectors.
First, to give the same importance to all classes, we normalize each feature vector with respect to the maximum value any of the feature channels for that particular class assumes, \ie:
\begin{equation}
\mathbf{\bar{f}}_i =  
\frac{\mathbf{f}_i}
{    \max_{\substack{g_{j,l} \in \mathbf{g}_j   \\  y_j^* = y_i^*}} g_{j,l} }
\ \ \ \ \ \ \ \ \ \
\mathbf{f}_i , \mathbf{g}_j \in \mathbf{F}_n 
\label{eq:features_normalization}
\end{equation}
%
%

\rev{We design the sparsity constraint as the ratio between the sum of exponentials and the linear sum of the elements of each feature vector:} 
\begin{equation}
\mathcal{L}_{sp} =  \frac{1}{|       \mathbf{f}_{i} \in \mathbf{F}_n     |}
\sum_{\mathbf{f}_{i} \in \mathbf{F}_n}
\frac{\sum_j \exp\left(\bar{f}_{i,j}\right)}
{ \sum_j \bar{f}_{i,j}}
\label{eq:L_spa}
\end{equation}
   
While the contrastive learning objective forces features to lie within tight semantically-consistent well-distanced clusters, the sparsity constraint aims at narrowing down the set of mid-range spurious activations with the aim of letting room for the representation of upcoming classes. In other words, by constraining features of the same classes to be tightly clustered and to be spaced apart from features of other classes and sparse, we can preserve 
geometrical space (few active channels) and expressiveness (division in well-separated clusters) for the latent representation of future classes. Empirically, we found entropy-based minimization \revised{methods in the latent space \cite{vu2019advent}  to be less reliable for our task.} 
\rev{In the \textit{Supplementary Material} we show some empirical insights on the regularization effects achieved by this constraint.}

\subsection{Output-Level Knowledge Distillation}
\label{subsec:knowledge}
The last component of our work is an output-level knowledge distillation which we show to be complementary to the previously introduced strategies.
 Indeed, we add knowledge distillation on top of all the other components to transfer knowledge from the old model's classifier to the current one. While previous constraints regularize the latent space achieving simultaneously an invariant features extraction with respect to previous steps and an easier addition of novel categories, output-level knowledge distillation directly acts on the classifier, to preserve its discriminative ability regarding old classes.
In particular, we start from the preliminary considerations of \cite{michieli2019incremental,michieli2020knowledge} and we employ the unbiased distillation proposed in \cite{cermelli2020modeling} as natural extension to the case in which the background may contain other categories. 
In this case we avoid to re-normalize the probabilities from the previous step and, instead, we compare the background probability from the previous step with 
the probability of having either a new class or the background (this accounts for the fact that the background  in the previous steps may include samples of the new classes, see \cite{cermelli2020modeling}).


\section{Training Procedure}
\label{sec:training}

To train and benchmark our approach we resort to two publicly available datasets following \cite{shmelkov2017incremental,michieli2019incremental,michieli2020knowledge,cermelli2020modeling}.
The \textbf{Pascal VOC 2012} \cite{pascalvoc2012} contains $10582$ images in the training split and $1449$ in the validation split (that we used for testing, as done by all competing works being the test set not publicly available). 
Each pixel of each image is assigned to one semantic label chosen among $21$ different classes ($20$ plus the background). 
The \textbf{ADE20K} \cite{zhou2017scene} is a large-scale dataset of $22210$ images, $2000$ of which form the validation split. The typical benchmark defined in \cite{zhou2017scene} includes $150$ classes,  representing both stuff (\eg, sky, building) and object classes (\eg, bottle, chair), differently from VOC 2012.

The proposed strategy is agnostic to the backbone architecture. For the experimental evaluation of all the compared methods we use a standard Deeplab-v3+ \cite{deeplabv3plus2018} architecture with  ResNet-101 \cite{he2016deep} as  backbone (differently from \cite{cermelli2020modeling} for wider reproducibility)
 with output stride of $16$. The backbone has been initialized using a pre-trained model on ImageNet \cite{deng2009imagenet} \revised{(see the \textit{Supplementary Material}
 for a detailed discussion of the impact of different pre-training strategies)}. We optimize the network weights following \cite{chen2017rethinking} with SGD and with same learning rate policy, momentum and weight decay. The first learning step involves an initial learning rate of $10^{-2}$, which is decreased to $10^{-3}$ for the following steps as done in \cite{shmelkov2017incremental,cermelli2020modeling}. The learning rate is decreased with a polynomial decay rule with power $0.9$.  In each learning step we train the models with a batch size of $8$ for $30$ epochs for Pascal VOC 2012 and a batch size of $4$ for $60$ epochs for ADE20K. Following \cite{chen2017rethinking}, we crop the images to $512\times 512$ during both training and validation and we apply the same data augmentation (\ie, random scaling the input images of a factor from $0.5$ to $2.0$ and random left-right flipping during training). In order to set the hyper-parameters of each method, we follow the same continual learning protocol of \cite{delange2019continual,cermelli2020modeling}, i.e, we used $20\%$ of the training set as validation and we report the results on the original validation set of the datasets.
We use Pytorch to develop and train all the models on a  NVIDIA 2080 Ti GPU. \revision{The code is available at:} {\small \url{https://lttm.dei.unipd.it/paper_data/SDR/}}.

\section{Experimental Results}
\label{sec:results}

We evaluate the performance of our method (denoted in the tables with \textbf{SDR}, \ie, Sparse and Disentangled Representations) against some state-of-the-art continual learning frameworks. We report as a lower limit the performance of the na\"ive fine-tuning approach (FT), which consists in training the model on the newly available training data with no additional provisions, while the upper limit is given by the offline single-shot training (offline) on the whole dataset $\mathcal{T}$ and on all the classes at once. Then, we compare with 3 recent continual semantic segmentation schemes:  ILT \cite{michieli2019incremental}, which combines latent and output level knowledge distillation, CIL \cite{klingner2020class}, which adds class importance weighting to output-level knowledge distillation, and MiB \cite{cermelli2020modeling}, which deals with the background distribution shift and proposes an unbiased weight initialization rule.
We also report the results on LwF \cite{li2018learning} (together with its single-headed version LwF-MC \cite{rebuffi2017icarl}), that according to \cite{cermelli2020modeling} is the best performing continual image classification algorithm when adapted to semantic segmentation.
%
%
For a fair comparison, all the methods have been re-trained with a standard Deeplab-v3+ \cite{deeplabv3plus2018} architecture with ResNet-101 \cite{he2016deep} as backbone.

\begin{table*}[t]
  \centering
  \caption{\revision{mIoU on multiple incremental scenarios and protocols on VOC2012. Best in \textbf{bold}, runner-up \underline{underlined}. \dag: results from \cite{cermelli2020modeling}.}}
  \setlength{\tabcolsep}{1.18pt}
  \footnotesize
    \begin{tabular}{l|ccc|ccc|ccc||ccc|ccc|ccc||ccc|ccc|ccc}
     & \multicolumn{9}{c||}{19-1}                                             & \multicolumn{9}{c||}{15-5}                                             & \multicolumn{9}{c}{15-1} \\
          & \multicolumn{3}{c|}{sequential} & \multicolumn{3}{c|}{disjoint} & \multicolumn{3}{c||}{overlapped} & \multicolumn{3}{c|}{sequential} & \multicolumn{3}{c|}{disjoint} & \multicolumn{3}{c||}{overlapped} & \multicolumn{3}{c|}{sequential} & \multicolumn{3}{c|}{disjoint} & \multicolumn{3}{c}{overlapped} \\
          Method & old   & new   & all   & old   & new   & all   & old   & new   & all   & old   & new   & all   & old   & new   & all   & old   & new   & all   & old   & new   & all   & old   & new   & all   & old   & new   & all \\\hline
    FT    & 63.4  & 21.2  & 61.4  & 35.2  & 13.2  & 34.2  & 34.7  & 14.9  & 33.8  & 62.0  & 38.1  & 56.3  & 8.4   & 33.5  & 14.4  & 12.5  & 36.9  & 18.3  & 49.0  & 17.8  & 41.6  & 5.8   & 4.9   & 5.6   & 4.9   & 3.2   & 4.5 \\\hdashline
    LwF  \cite{li2018learning} & 67.2  & 26.4  & 65.3  & 65.8  & 28.3  & 64.0  & 62.6  & 23.4  & 60.8  & 68.0  & 43.0  & 62.1  & 39.7  & 33.3  & 38.2  & 67.0  & 41.8  & 61.0  & 33.7  & 13.7  & 29.0  & 26.2  & \underline{15.1}  & 23.6  & 24.0  & \underline{15.0}  & 21.9 \\
    LwF-MC \cite{rebuffi2017icarl} & 49.2  & 0.9   & 46.9  & 38.5  & 1.0   & 36.7  & 37.1  & 2.3   & 35.4  & 70.6  & 19.5  & 58.4  & 41.5  & 25.4  & 37.6  & 59.8  & 22.6  & 51.0  & 12.1  & 1.9   & 9.7   & 6.9   & 2.1   & 5.7   & 6.9   & 2.3   & 5.8 \\
    ILT \cite{michieli2019incremental}  & 64.3  & 22.7  & 62.3  & 66.9  & 23.4  & 64.8  & 50.2  & \underline{29.2}  & 49.2  & 71.3  & \textbf{47.8}  & 65.7  & 31.5  & 25.1  & 30.0  & 69.0  & 46.4  & 63.6  & 49.2  & \textbf{30.3}  & \textbf{48.3}  & 6.7   & 1.2   & 5.4   & 5.7   & 1.0   & 4.6 \\
    CIL \cite{klingner2020class}  & 64.1  & 22.8  & 62.1  & 62.6  & 18.1  & 60.5  & 35.1  & 13.8  & 34.0  & 63.8  & 39.8  & 58.1  & 42.6  & 35.0  & 40.8  & 14.9  & 37.3  & 20.2  & 52.4  & \underline{22.3}  & 45.2  & 33,3  & \textbf{15.9}  & 29.1  & 6.3   & 4.5   & 5.9 \\
    MiB\dag  \cite{cermelli2020modeling} & -  & -  & -  & 69.6 & 25.6 & 67.4 & \underline{70.2} & 22.1 & \underline{67.8} & - & - & - & 71.8 & 43.3 & 64.7 & \underline{75.5} & 49.4 & 69.0 & - & - & - & 46.2 & 12.9 & 37.9 & 35.1 & 13.5 & 29.7 \\
    MiB  \cite{cermelli2020modeling} & 68.2  & \underline{31.9}  & 66.5  & 67.0  & 26.0  & 65.1  & 69.6  & 23.8  & 67.4  & 73.0  & 44.4  & 66.1  & 47.5  & 34.1  & 44.3  & 73.1  & 44.5  & 66.3  & 35.7  & 11.0  & 29.8  & 39.0  & 15.0  & 33.3  & 44.5  & 11.7  & 36.7 \\\hdashline
    SDR (ours)  & \underline{68.4}  & \textbf{35.3}  & \underline{66.8}  & \underline{69.9}  & \textbf{37.3}  & \underline{68.4}  & 69.1  & \textbf{32.6}  & 67.4  & \underline{73.6}  & \underline{46.7}  & \underline{67.2}  & \underline{73.5}  & \textbf{47.3}  & \underline{67.2}  & 75.4  & \textbf{52.6}  & \underline{69.9}  & \textbf{58.5}  & 10.1  & 47.0  & \underline{59.2}  & 12.9  & \underline{48.1}  & \underline{44.7}  & \textbf{21.8}  & \underline{39.2} \\
    SDR + MiB & \textbf{70.6}  & 24.8  & \textbf{68.5}  & \textbf{70.8}  & \underline{31.4}  & \textbf{68.9}  & \textbf{71.3}  & 23.4  & \textbf{69.0}  & \textbf{74.6}  & 43.8  & \textbf{67.3}  &\textbf{ 74.6}  & \underline{44.1}  & \textbf{67.3}  & \textbf{76.3}  & \underline{50.2}  & \textbf{70.1}  & \underline{58.1}  & 11.8  & \underline{47.1}  & \textbf{59.4 } & 14.3  & \textbf{48.7 } & \textbf{47.3}  & 14.7  & \textbf{39.5} \\\hdashline
    offline & 75.5  & 73.5  & 75.4  & 75.5  & 73.5  & 75.4  & 75.5  & 73.5  & 75.4  & 77.5  & 68.5  & 75.4  & 77.5  & 68.5  & 75.4  & 77.5  & 68.5  & 75.4  & 77.5  & 68.5  & 75.4  & 77.5  & 68.5  & 75.4  & 77.5  & 68.5  & 75.4 \\
    \end{tabular}%
  \label{tab:voc_quantitative}%
\end{table*}

\newcommand{\imgsize}{18.6mm}
\begin{figure*}[htbp]
\setlength{\tabcolsep}{0.7pt} 
\renewcommand{\arraystretch}{0.6}
\centering
\begin{tabular}{ccccccccccc}
  
   \rotatebox{90}{\hspace{+4ex}19-1} &  
   \includegraphics[width=\imgsize]{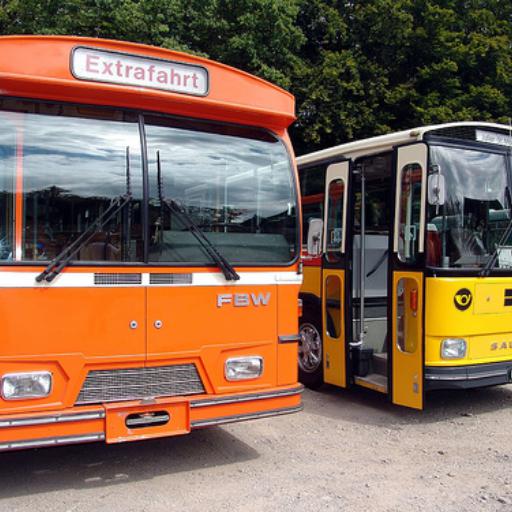} &
   \includegraphics[width=\imgsize]{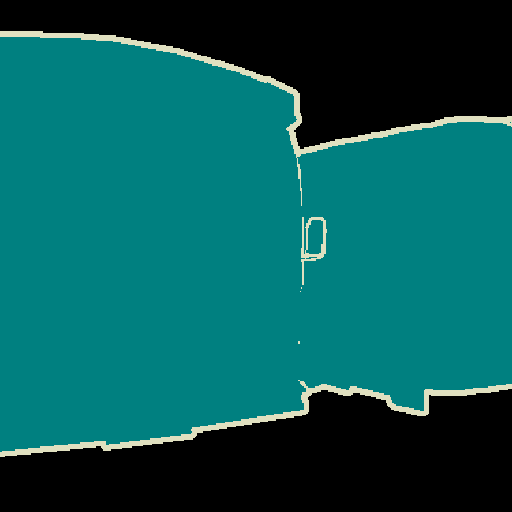} & 
   \includegraphics[width=\imgsize]{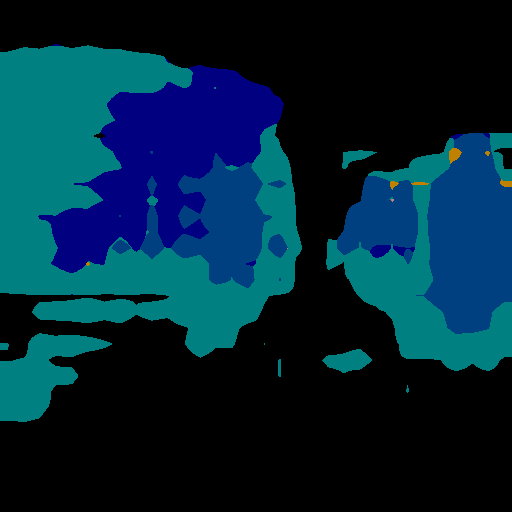} & 
   \includegraphics[width=\imgsize]{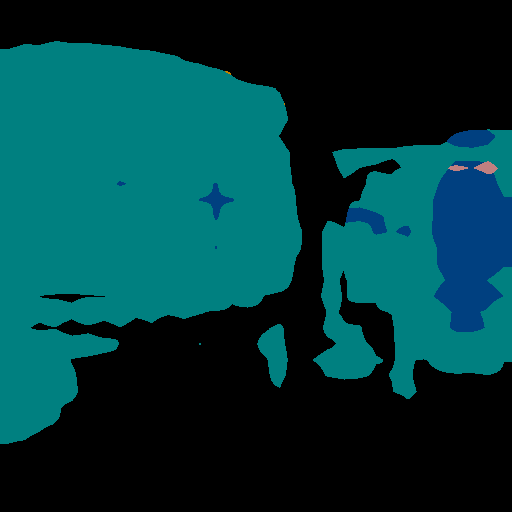} & 
   \includegraphics[width=\imgsize]{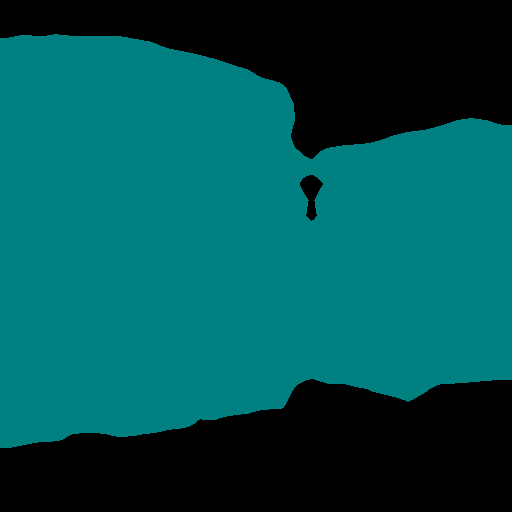} &
   \includegraphics[width=\imgsize]{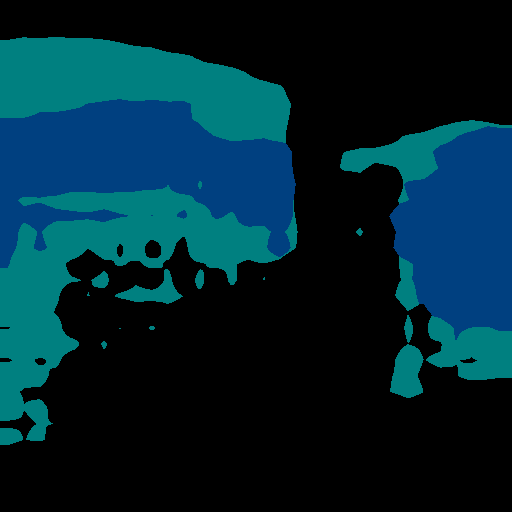} &
   \includegraphics[width=\imgsize]{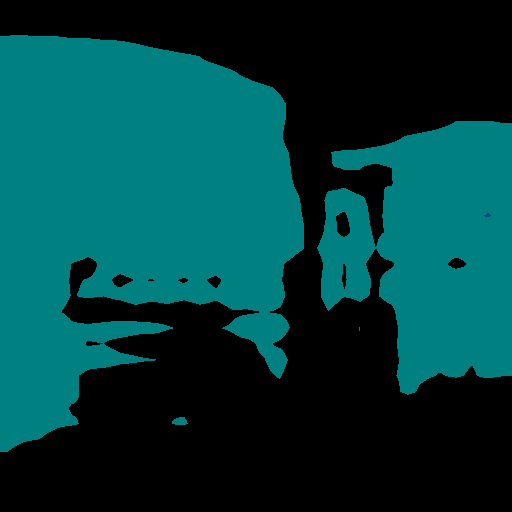} &
   \includegraphics[width=\imgsize]{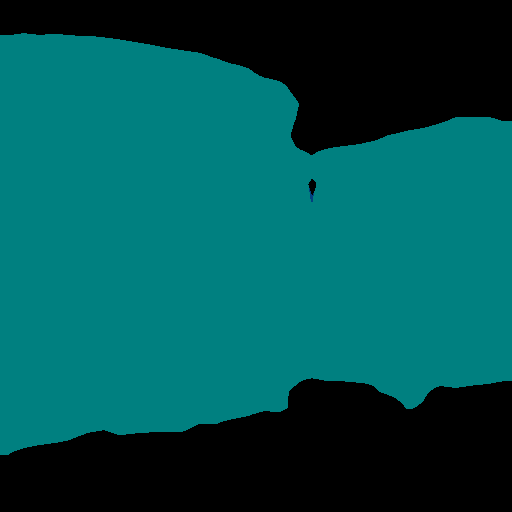} &
   \includegraphics[width=\imgsize]{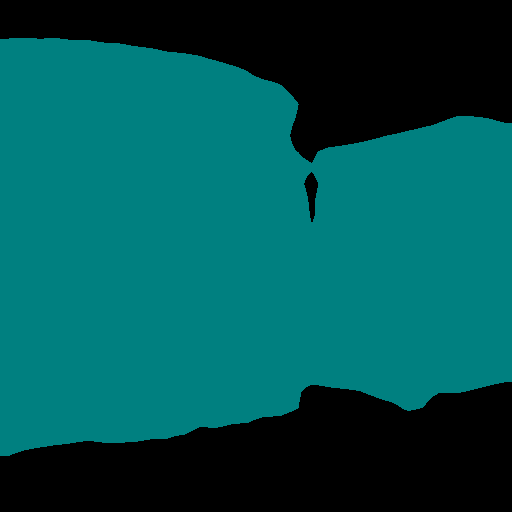} \\ 
   
      \rotatebox{90}{\hspace{+4ex}15-5} &  
   \includegraphics[width=\imgsize]{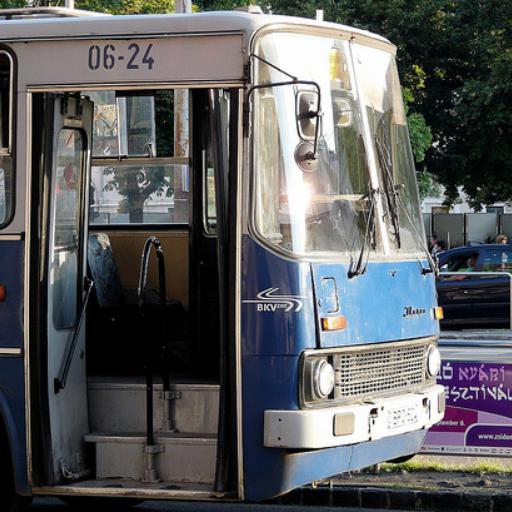} &
   \includegraphics[width=\imgsize]{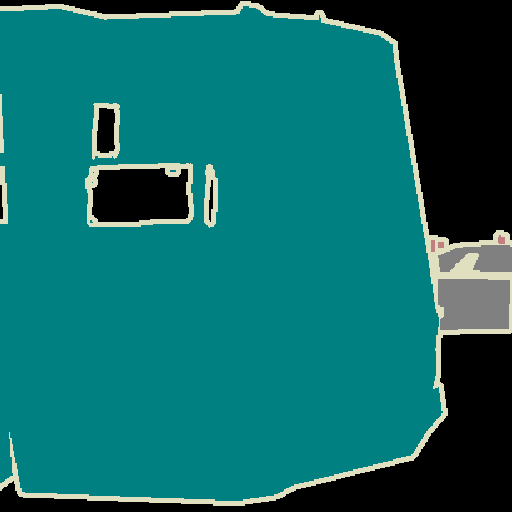} & 
   \includegraphics[width=\imgsize]{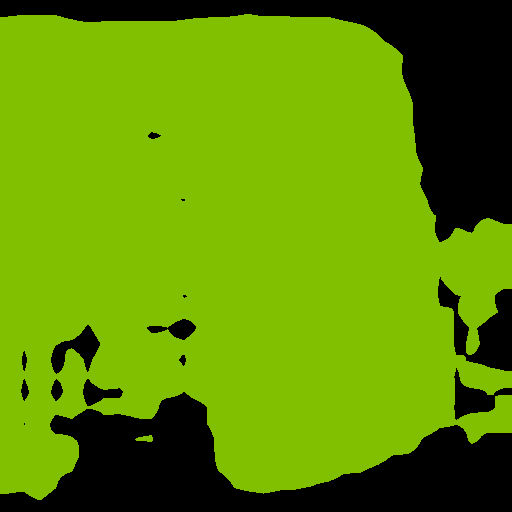} & 
   \includegraphics[width=\imgsize]{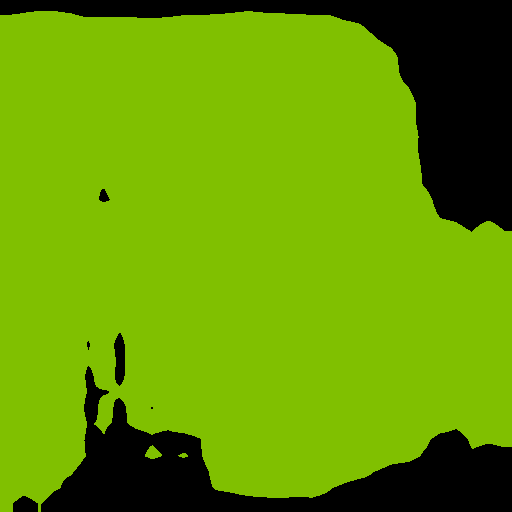} & 
   \includegraphics[width=\imgsize]{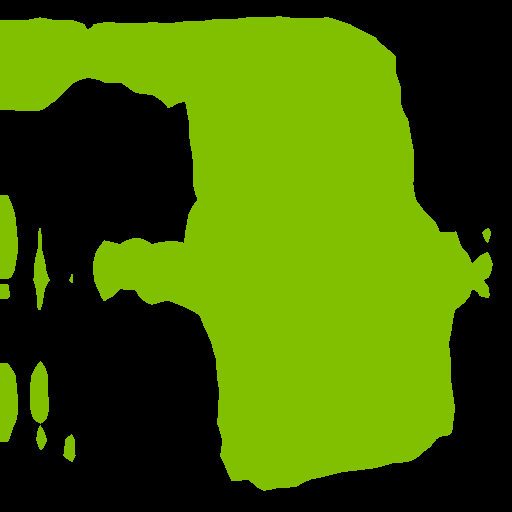} &
   \includegraphics[width=\imgsize]{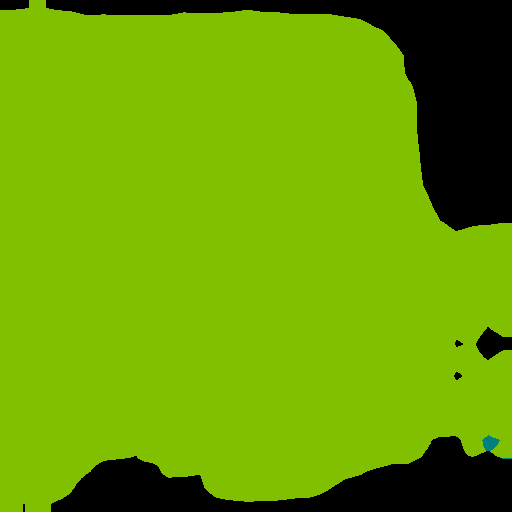} &
   \includegraphics[width=\imgsize]{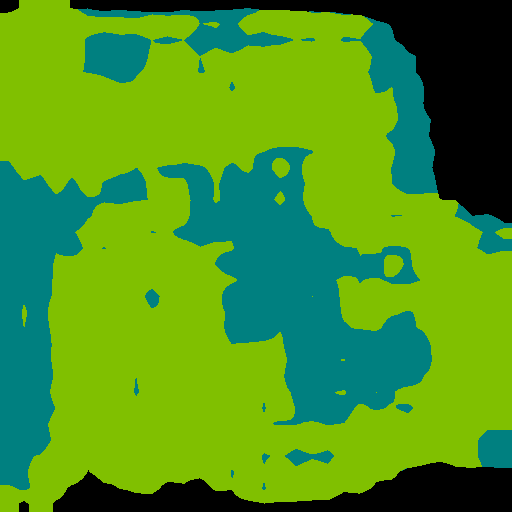} &
   \includegraphics[width=\imgsize]{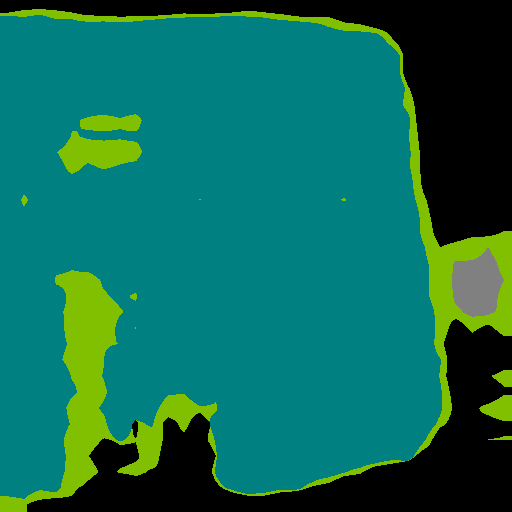} &
   \includegraphics[width=\imgsize]{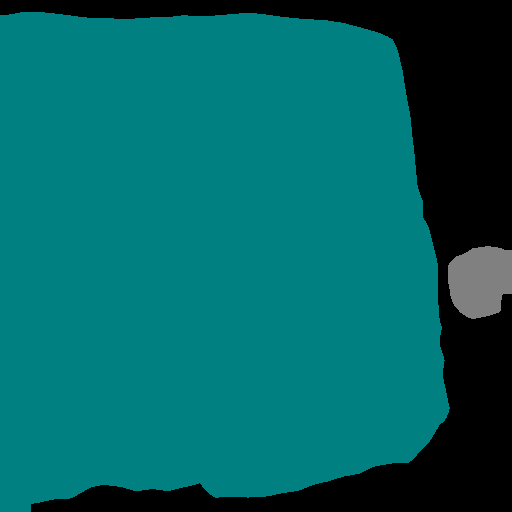} \\

         \rotatebox{90}{\hspace{+4ex}15-1} &  
   \includegraphics[width=\imgsize]{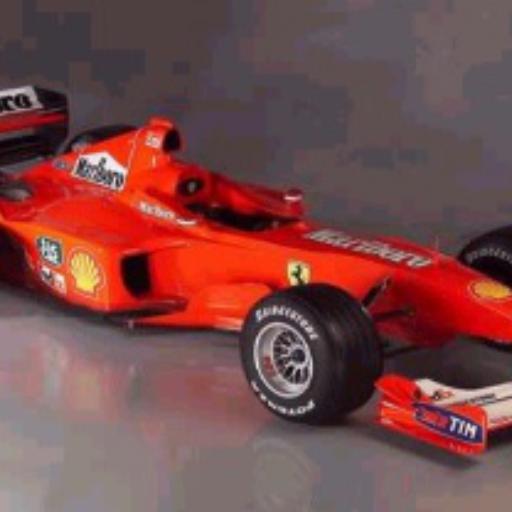} &
   \includegraphics[width=\imgsize]{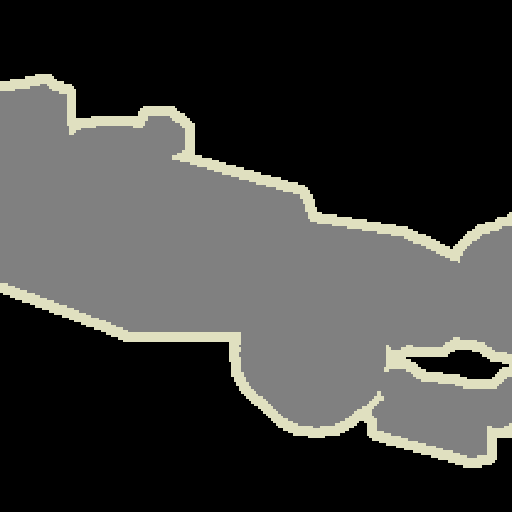} & 
   \includegraphics[width=\imgsize]{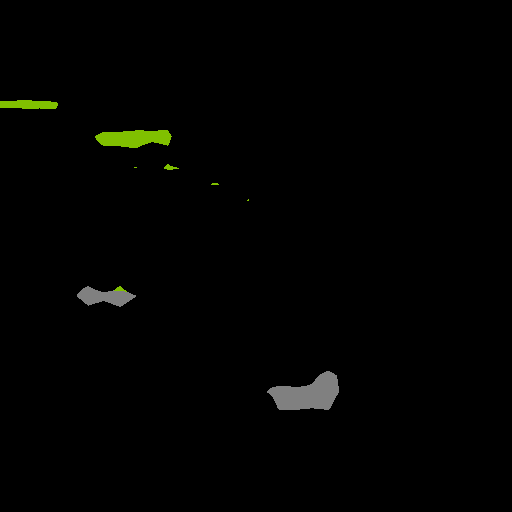} & 
   \includegraphics[width=\imgsize]{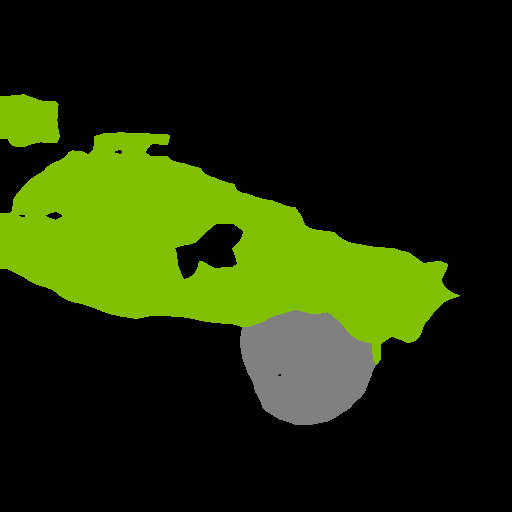} & 
   \includegraphics[width=\imgsize]{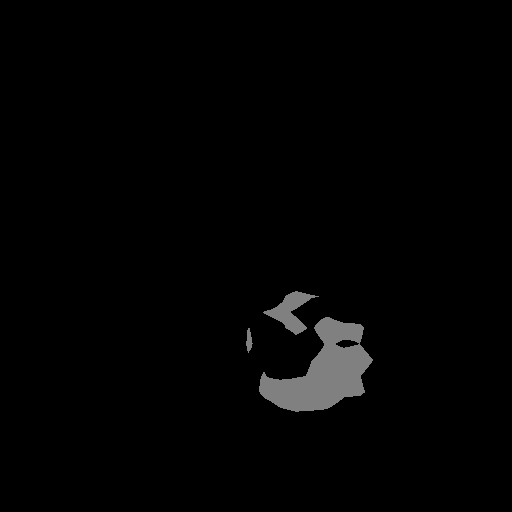} &
   \includegraphics[width=\imgsize]{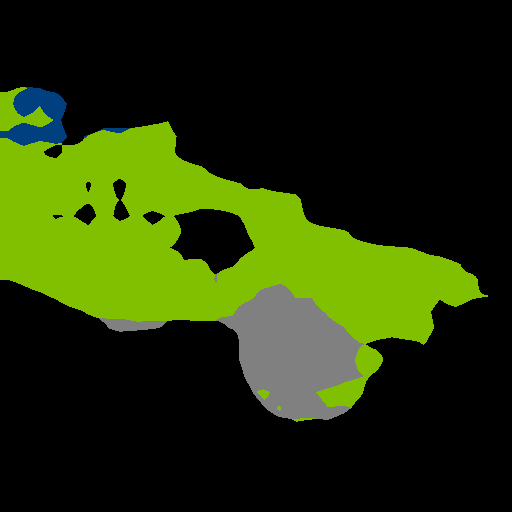} &
   \includegraphics[width=\imgsize]{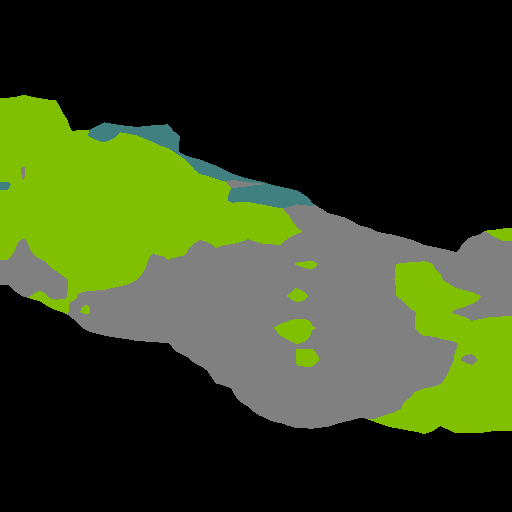} &
   \includegraphics[width=\imgsize]{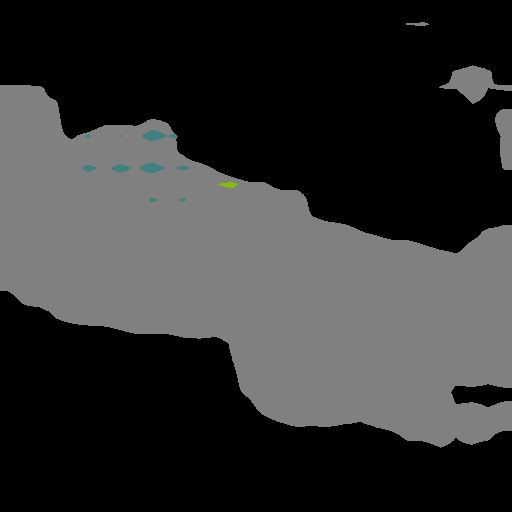} &
   \includegraphics[width=\imgsize]{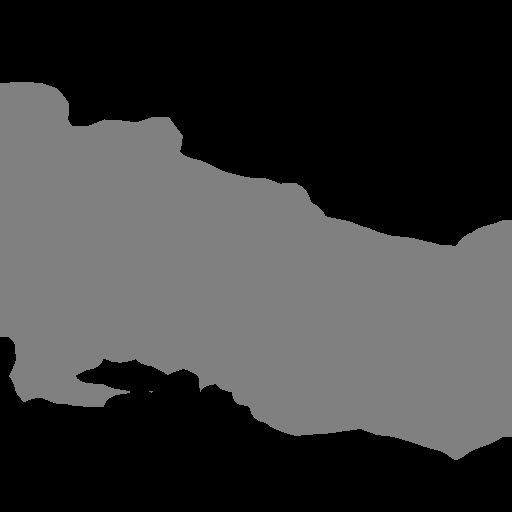} \\

   \rotatebox{90}{\hspace{+3ex}100-50} &  
   \includegraphics[width=\imgsize]{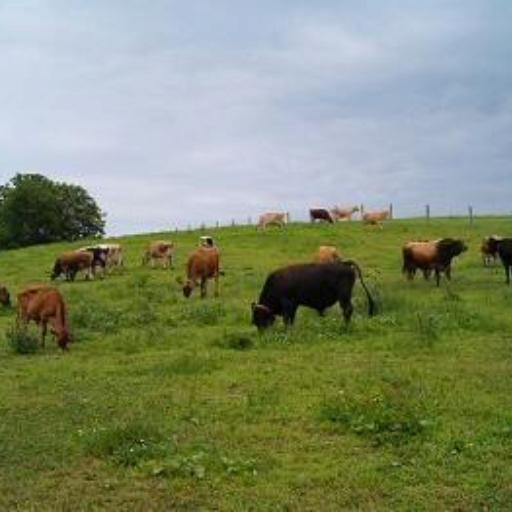} &
   \includegraphics[width=\imgsize]{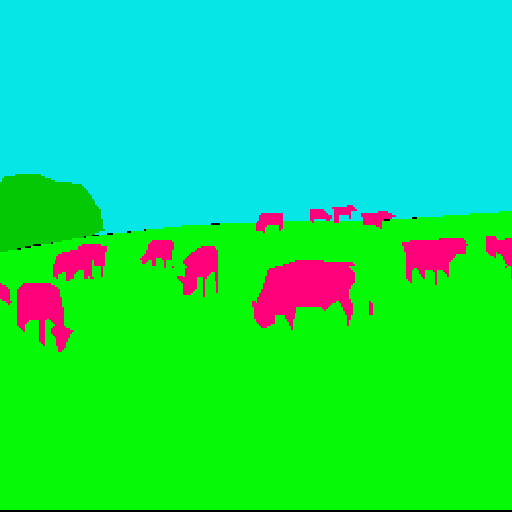} & 
   \includegraphics[width=\imgsize]{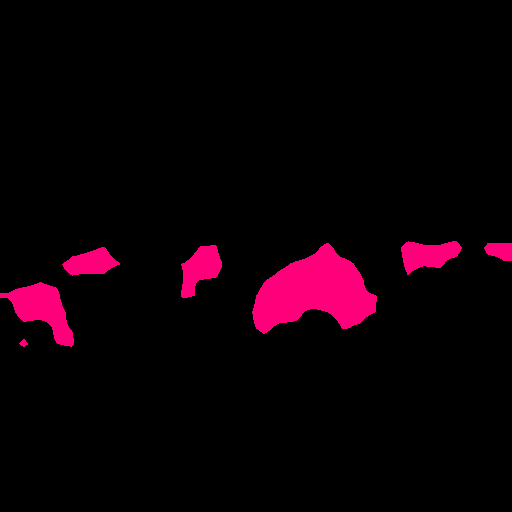} & 
   \includegraphics[width=\imgsize]{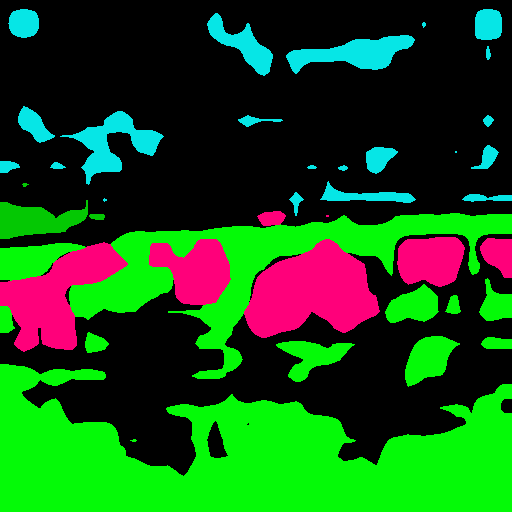} & 
   \includegraphics[width=\imgsize]{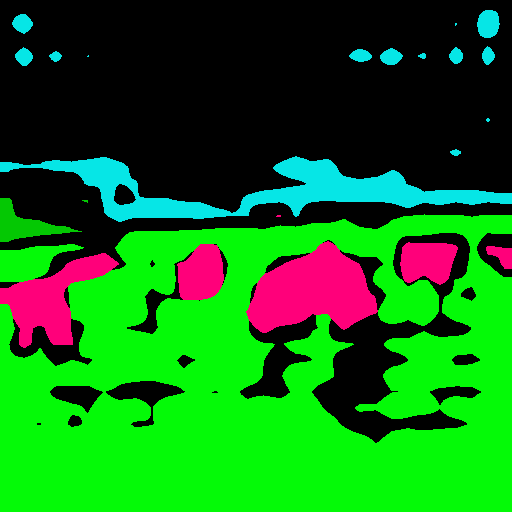} &
   \includegraphics[width=\imgsize]{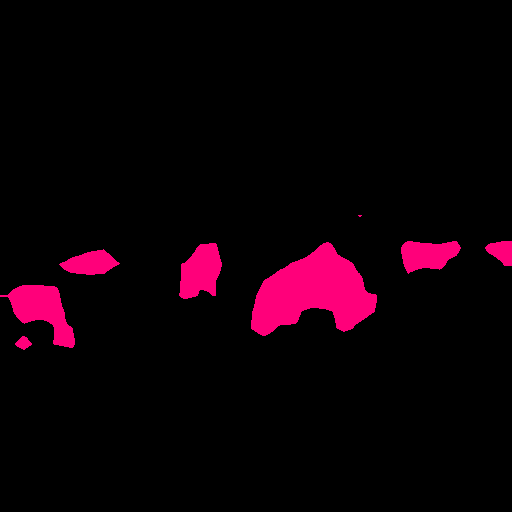} &
   \includegraphics[width=\imgsize]{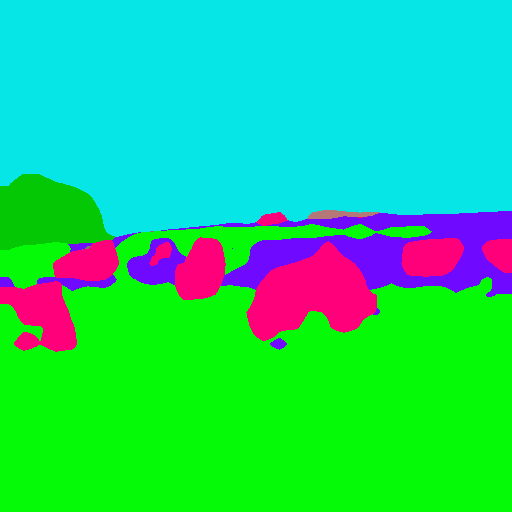} &
   \includegraphics[width=\imgsize]{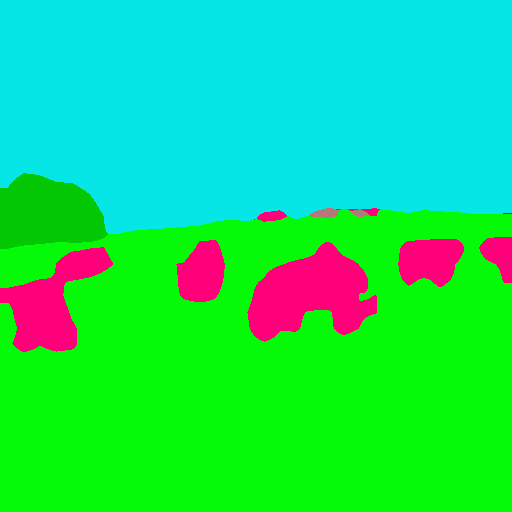} &
   \includegraphics[width=\imgsize]{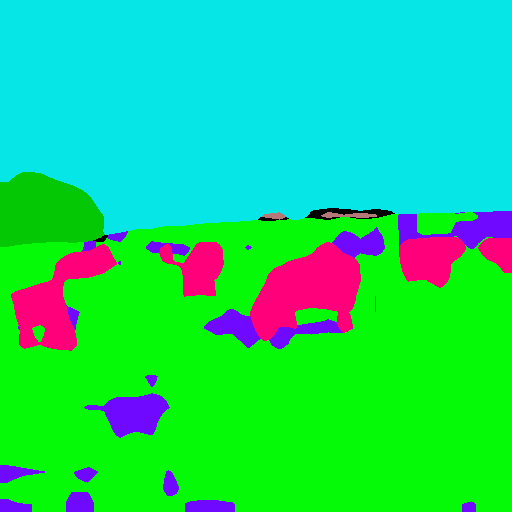} \\ 
   
   \rotatebox{90}{\hspace{+3ex}100-10} &  
   \includegraphics[width=\imgsize]{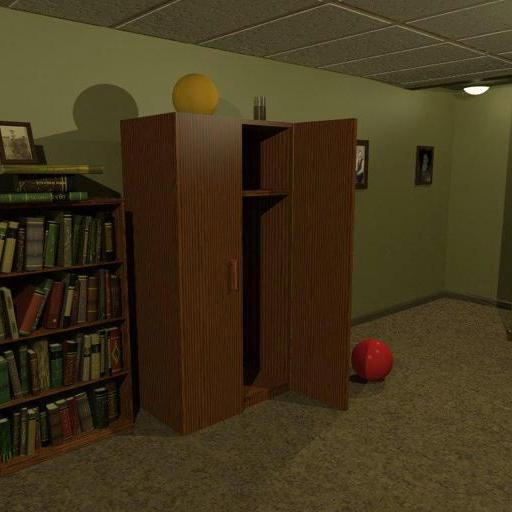} &
   \includegraphics[width=\imgsize]{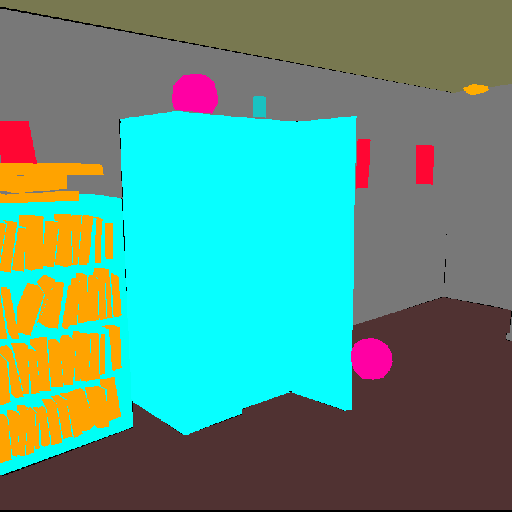} & 
   \includegraphics[width=\imgsize]{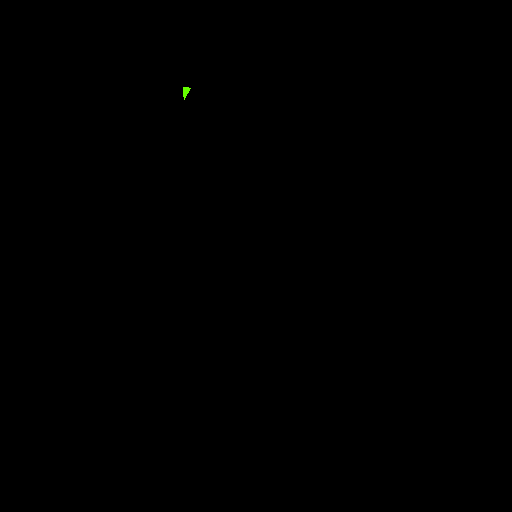} & 
   \includegraphics[width=\imgsize]{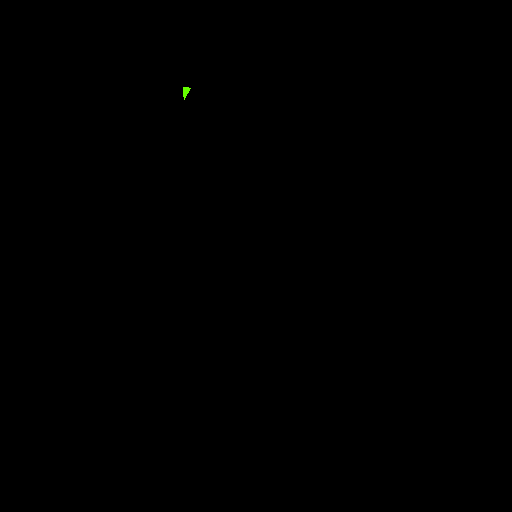} & 
   \includegraphics[width=\imgsize]{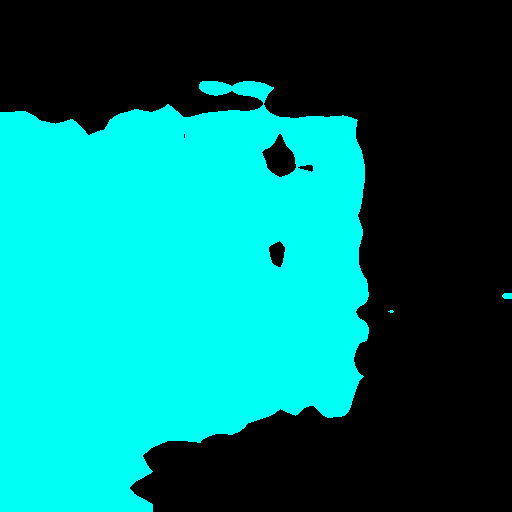} &
   \includegraphics[width=\imgsize]{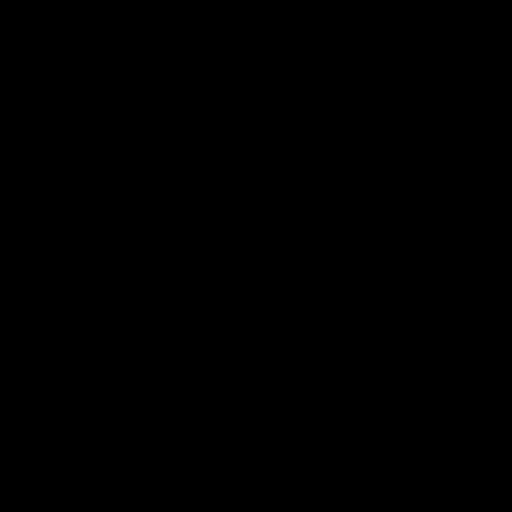} &
   \includegraphics[width=\imgsize]{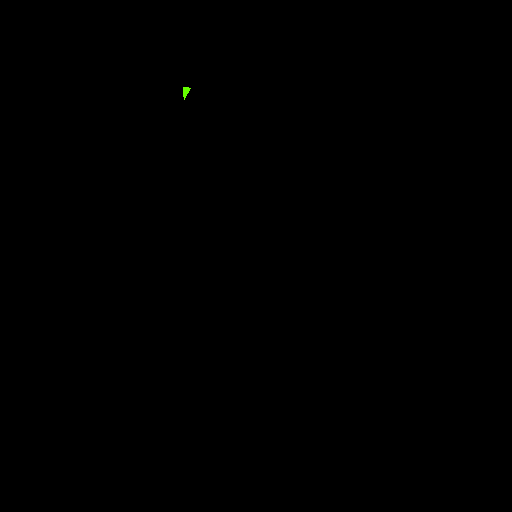} &
   \includegraphics[width=\imgsize]{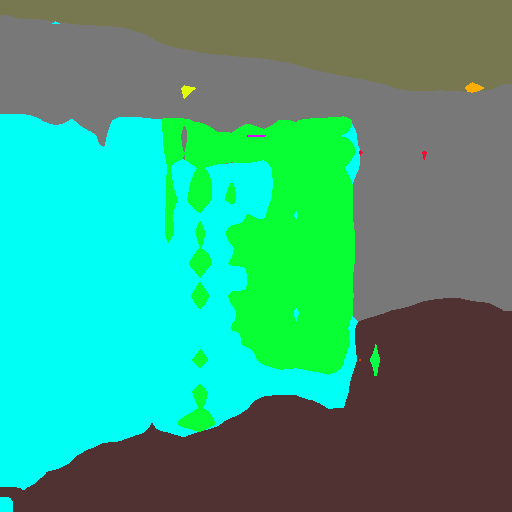} &
   \includegraphics[width=\imgsize]{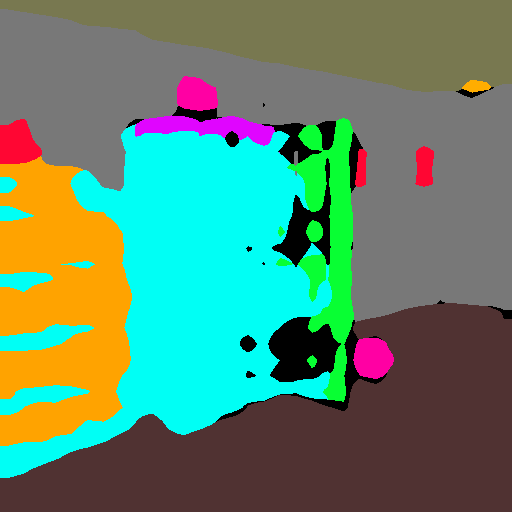} \\ 
   
   \rotatebox{90}{\hspace{+4ex}50-50} &  
   \includegraphics[width=\imgsize]{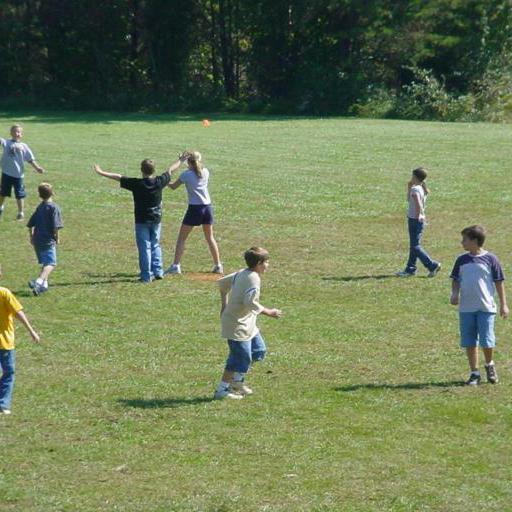} &
   \includegraphics[width=\imgsize]{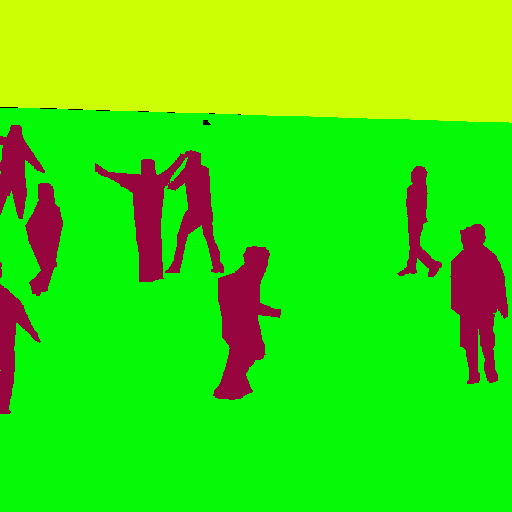} & 
   \includegraphics[width=\imgsize]{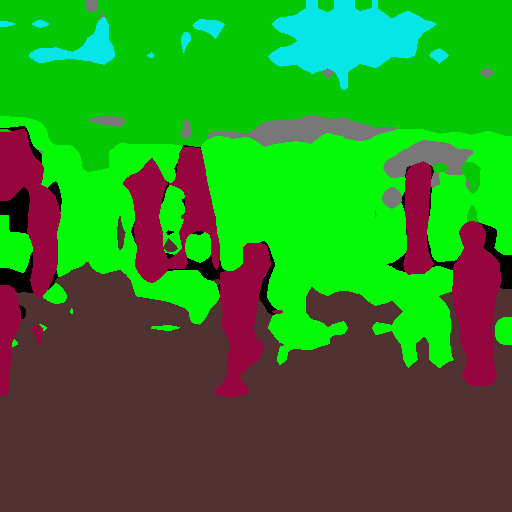} & 
   \includegraphics[width=\imgsize]{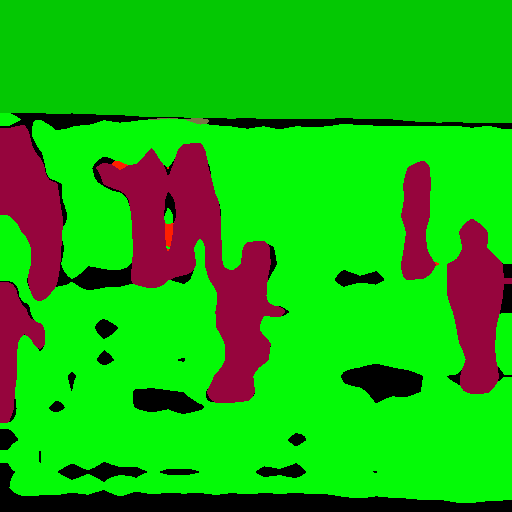} & 
   \includegraphics[width=\imgsize]{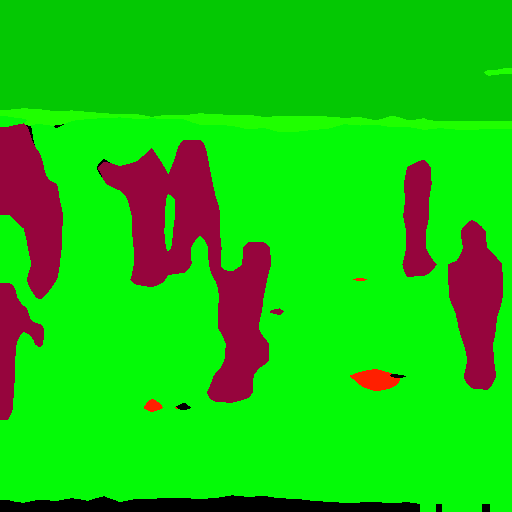} &
   \includegraphics[width=\imgsize]{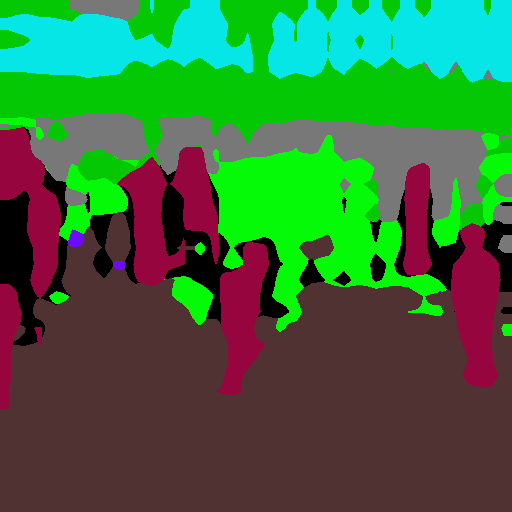} &
   \includegraphics[width=\imgsize]{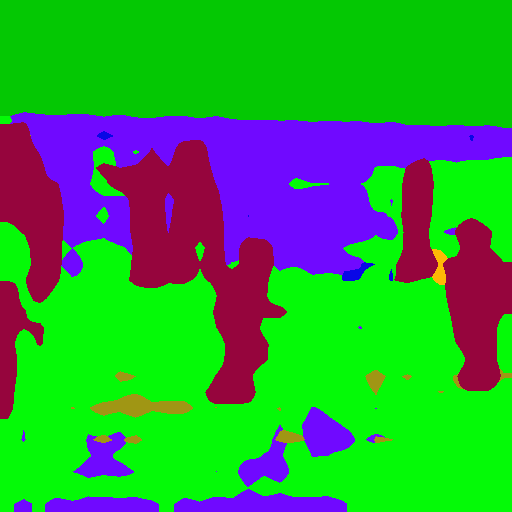} &
   \includegraphics[width=\imgsize]{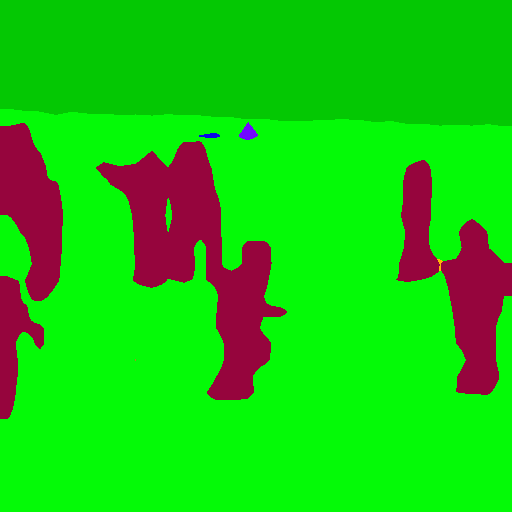} &
   \includegraphics[width=\imgsize]{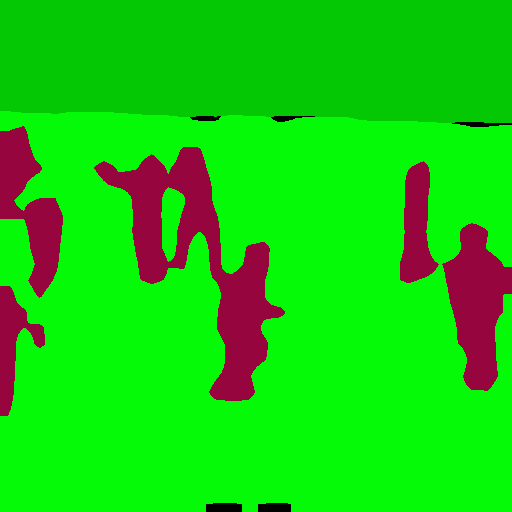} \\ 
   
  &  RGB  &  label  &  FT  &  LwF \cite{kirkpatrick2017overcoming}  &  ILT \cite{michieli2019incremental}  &  CIL \cite{klingner2020class} & MiB \cite{cermelli2020modeling} & SDR (ours) & offline \\
  
\end{tabular}
\caption{Qualitative results of competing approaches in different scenarios on disjoint VOC 2012 and ADE20K (\textit{best viewed in colors}).}
\label{fig:voc_qualitative}
\end{figure*}
%
%
%

\subsection{Pascal VOC2012}
\label{sec:pascal}

Following previous works \cite{shmelkov2017incremental,michieli2019incremental,michieli2020knowledge,cermelli2020modeling}, we design three main experiments adding one class (19-1), five classes at once (15-5) and five classes sequentially (15-1) added in alphabetical order. In Table~\ref{tab:voc_quantitative} we report comprehensive results on the three experimental protocols defined in Section \ref{subsec:protocols}. \revision{Results are averaged for mIoU of classes in the base step (\textit{old}), for classes in the incremental steps (\textit{new}) and for \textit{all} classes, and are reported at the end of all the incremental steps.} 
\revised{For \cite{cermelli2020modeling} we also report the original results in their paper (denoted with MiB$\dag$), that uses a different backbone (thus different pre-trained model) and batch size.}

We can appreciate forgetting of previous classes and intransigence in learning new ones even when adding as little as one class (the \textit{tv/monitor} class is added) in the scenario 19-1. FT always leads to the worst mIoU in terms of \textit{old}, \textit{new} and \textit{all} classes. 
Incremental methods designed for semantic segmentation allow for a stable improvement across the experimental protocols, in particular MiB, that is specifically targeted to solve the disjoint and the overlapped scenarios, while CIL and ILT encounter difficulties in the overlapped scenario. Also LwF allows for a good improvement while its single-headed version has lower performance in this scenario.
 Our method (SDR) significantly outperforms all the competitors in the disjoint and overlapped scenarios (with a gap of more than $3\%$ against the best competing approach in the disjoint setup), while in the sequential setup the gap is smaller. Further adding on top of our method the MiB framework \revised{(\ie, unbiased cross entropy, knowledge distillation and classifier initialization)}, which we regard as the current state-of-the-art approach for class incremental semantic segmentation, the results increase on all the scenarios, showing that proposed techniques are complementary with respect to previous schemes.

When moving to the addition of 5 classes at once (\ie, \textit{potted plant}, \textit{sheep}, \textit{sofa}, \textit{train}, \textit{tv/monitor}) we immediately notice an overall increased drop of performance of all compared methods, especially in disjoint and overlapped protocols, due to the increased domain shift occurring when adding more classes at once with very variable content. In this and in the following scenario, indeed, we are adding to the model classes belonging to different macroscopic groups, according to \cite{pascalvoc2012}, which are responsible for a variegate distribution: three indoor classes (\textit{potted plant}, \textit{sofa} and \textit{tv/monitor}), one animal class (\textit{sheep}) and one vehicle class (\textit{train}). All compared methods obtain a relevant improvement with respect to FT 
but are always surpassed by SDR, which in particular outrun the best competing method (MiB) by more than $20\%$ in the disjoint scenario.

In the final scenario we add the last 5 classes sequentially in 5 consecutive learning steps. This approach leads to the largest accuracy drop being the model exposed to a reiterated addition of single classes, which are also coming from different semantic contexts.
In the sequential scenario LwF and MiB (which is designed for background shift) show poor final accuracy. ILT and CIL, instead, show results comparable to our approach.
In the disjoint and in the overlapped scenarios all the methods heavily suffer from the semantic shift undergone by the background class: LwF (both versions) and ILT have poor performance in these scenarios, while CIL is able to achieve some improvement only in the disjoint scenario. The best competitor is again MiB that is able to obtain a mIoU of $33\%$ and $36.7\%$ in the disjoint and overlapped scenarios respectively.
Our approach (SDR) is able to significantly increase the final mIoU in both scenarios to $48.1\%$ and $39.2\%$; it achieves a remarkable result especially in the disjoint scenario thanks to the novel features-level constraints which help the model to maintain accuracy on old classes while learning new ones. 

Visual results for each scenario in the disjoint protocol are shown in the first three rows of Fig.~\ref{fig:voc_qualitative}, where our method is compared against all the competitors consistently obtaining better segmentation maps.
For example, our method does not mislead the \textit{bus} windows with \textit{tv/monitor} instances in row 1 differently from several competitors (which are more biased toward predicting the novel class), and it is the only one able to distinguish the \textit{bus} in row 2 and the \textit{car} in row 3 from the similar-looking \textit{train} class. Here, \textit{train} is added in the incremental step causing catastrophic forgetting of similar classes in competing approaches.

\begin{table}[t]
  \caption{mIoU over multiple incremental scenarios on disjoint setup of ADE20K. Best in \textbf{bold}, runner-up \underline{underlined}.}
  \setlength{\tabcolsep}{1.3pt}
  \small
  \centering
    \begin{tabular}{|l|ccc|ccc|ccc|}
    \hline
          & \multicolumn{3}{|c|}{100-50} & \multicolumn{3}{c|}{100-10} & \multicolumn{3}{c|}{50-50} \\
          Method & old   & new   & all   & old   & new   & all   & old   & new   & all \\\hline
    FT    & 0.0   & 22.5  & 7.5   & 0.0   & 2.5   & 0.8   & 13.9  & 12.0  & 12.6 \\\hdashline
    LwF \cite{li2018learning}   & 25.0  & 23.9  & 24.6  & 5.4   & 5.6   & 5.5   & 32.2  & 22.9  & 26.0 \\
    LwF-MC \cite{rebuffi2017icarl} & 8.6   & 0.0   & 5.8   & 0.0   & 0.9   & 0.3   & 2.8   & 0.5   & 1.2 \\ 
    ILT \cite{michieli2019incremental}  & 27.2  & 21.7  & 25.4  & 0.0   & 0.2   & 0.8   & \underline{41.9}  & 21.1  & 28.0 \\
    CIL \cite{klingner2020class}   & 0.0   & 22.5  & 7.5   & 0.0   & 2.0   & 0.6   & 14.0  & 11.9  & 12.6 \\
    MiB \cite{cermelli2020modeling}  & \textbf{37.6} & 24.7  & \underline{33.3}  & \underline{21.0} & 5.3   & 15.8  & 39.1  & 22.6  & 28.1 \\\hdashline
    SDR (ours) & 37.4  & \underline{24.8}  & 33.2  & \textbf{28.9} & \underline{7.4}   & \underline{21.7}  & 40.9  & \underline{23.8}  & \underline{29.5} \\
    SDR+MiB & \underline{37.5}  & \textbf{25.5} & \textbf{33.5} & \textbf{28.9} & \textbf{11.7} & \textbf{23.2} & \textbf{42.9} & \textbf{25.4} & \textbf{31.3} \\\hdashline
    offline & 43.9  & 27.2  & 38.3  & 43.9  & 27.2  & 38.3  & 50.9  & 32.1  & 38.3 \\\hline
    \end{tabular}%
  \label{tab:ade_quantitative}%
\end{table}%

\subsection{ADE20K}
\label{sec:ade}

Following  \cite{cermelli2020modeling} we split the dataset into disjoint image sets with the only constraint that a minimum number of images (\ie, 50) have labeled pixels on $\mathcal{C}_k$. Classes are ordered according to \cite{zhou2017scene}. In this comparison we report the same competing methods of Section \ref{sec:pascal}. The scenarios we consider are the addition of the last 50 classes at once (100-50), of the last 50 classes 10 at a time (100-10) and of the last 100 classes in 2 steps of 50 classes each (50-50). The results are summarized in Table~\ref{tab:ade_quantitative}, where we can appreciate that the proposed approach outperforms competitors in every scenario, in particular with a larger gain when multiple incremental steps are performed. When adding 50 classes at a time LwF-MC and CIL achieve low results and are outperformed by the other competitors (\ie, LwF, ILT and MiB), which in turn are always consistently surpassed by our framework. In the scenario 100-10, instead, all competing approaches (except for MiB) are unable to provide useful outputs leading to extremely low results, while our method stands out from competitors outperforming also MiB by a good margin. Visual results for each scenario are reported as last rows of Fig.~\ref{fig:voc_qualitative}, which confirm our considerations showing how SDR produces less noisy predictions and does not overestimate the background as some competitors.

\subsection{Ablation Study}

\begin{table}[t]
\caption{\revision{Ablation on disjoint VOC2012 15-1 in terms of mIoU.}}
\vspace{0.1cm}
\label{tab:ablation}
\setlength{\tabcolsep}{4pt}
\centering
\begin{tabular}{cccccc|ccc}
$\mathcal{L}_{ce}$ & $\mathcal{L}_{pm}$  & $\mathcal{L}_{sp}$ & $\mathcal{L}_{cl}$  & $\mathcal{L}'_{kd}$ &$\mathcal{L}_{kd}$ & old & new & all \\\hline
\checkmark & & & & & & 5.8 & 4.9 & 5.6 \\
\checkmark & & & & \checkmark & & 30.0 & 11.0 & 25.4 \\
\checkmark & \checkmark & & & & & 18.7 & 9.0 & 16.4 \\
\checkmark & \checkmark & \checkmark & & & & 40.4 & 12.9 & 33.9 \\
\checkmark & \checkmark & \checkmark & \checkmark & & & 41.0 & 13.2 & 34.4 \\\hline
\checkmark & \checkmark & \checkmark & \checkmark & \checkmark & & 50.0 & \textbf{15.9} & 41.9 \\
\checkmark & \checkmark & \checkmark & \checkmark &  & \checkmark & \textbf{59.2} & 12.9 & \textbf{48.1} \\
\end{tabular}
\end{table}

To evaluate the effect of each component, we report an ablation analysis in Table~\ref{tab:ablation} on the Pascal dataset in the  challenging 15-1 scenario. As already noticed, FT leads to a great degradation of mIoU. Early continual semantic segmentation approaches use a classical output-level knowledge distillation \cite{michieli2019incremental,michieli2020knowledge,klingner2020class} which  show discrete benefits boosting the mIoU by almost $20\%$. 
\revised{Each component of the  approach} significantly contributes to the final mIoU providing non-overlapping and mutual benefits. Matching prototypes, sparsifying features vectors and constraining them via the contrastive objective regularize the latent space bringing large improvements on both old and new classes. \revision{We observe that also the contrastive loss brings a significant contribution if applied alone improving the mIoU of $13.5\%$.} Introducing standard output-level knowledge distillation on top increases the accuracy on old classes mainly, and its unbiased version prevents forgetting even more accounting for the background shift across the incremental learning steps.

\begin{table}[t]
\caption{\revision{Standard (non-incremental) semantic segmentation.}}
\vspace{0.1cm}
\label{tab:standard_segmentation}
\setlength{\tabcolsep}{6pt}
\centering
\begin{tabular}{ccc|cc}
$\mathcal{L}_{ce}$ & $\mathcal{L}_{sp}$ & $\mathcal{L}_{cl}$  & mIoU$_\mathrm{VOC2012}$ & mIoU$_\mathrm{ADE20K}$ \\\hline
\checkmark & & & 75.4  & 38.3 \\
\checkmark & \checkmark  & & 75.8 & 38.7 \\
\checkmark & & \checkmark  & 75.8 & 38.8 \\
\checkmark & \checkmark & \checkmark  & \textbf{76.3} & \textbf{39.3} \\
\end{tabular}
\end{table}

Finally, we show that two of the proposed approaches (namely, sparsity and contrastive learning) may be beneficial also for the more general case of standard (\ie, non incremental) semantic segmentation. Hence, we conduct some additional experiments on Pascal VOC2012 and ADE20K, reported in Table~\ref{tab:standard_segmentation}, showing the clear benefit of the two components in this setup. On both datasets the outcome is consistent, gaining $0.9\%$ and $1\%$ respectively, even starting from an architecture (\ie, Deeplab-v3+) which is already state of the art.

\section{Conclusion}
\label{sec:conclusion}

In this paper we presented some latent representation shaping techniques to prevent forgetting in continual semantic segmentation. In particular, the proposed constraints on the latent space regularize the learning process reducing forgetting whilst simultaneously improving the recognition of novel classes.
A prototypes matching constraint enforces latent space consistency on old classes, a features sparsification objective reduces the number of active channels limiting cross-talk between features of different classes, and contrastive learning clusters features according to their semantic while tearing apart those of different classes. Our evaluation shows the effectiveness of the proposed techniques, which can also be seamlessly applied in combination of previous methods (\eg, knowledge distillation). Future research will exploit the proposed techniques in different tasks, such as standard semantic segmentation and class-incremental open-set domain adaptation, and explore the combination of our approach with  output-level techniques.

{\small
\bibliographystyle{ieee_fullname}
\bibliography{biblio}

\begin{thebibliography}{10}\itemsep=-1pt

\bibitem{achille2018life}
Alessandro Achille, Tom Eccles, Loic Matthey, Chris Burgess, Nicholas Watters,
  Alexander Lerchner, and Irina Higgins.
\newblock Life-long disentangled representation learning with cross-domain
  latent homologies.
\newblock In {\em Neural Information Processing Systems (NeurIPS)}, pages
  9873--9883, 2018.

\bibitem{aljundi2018selfless}
Rahaf Aljundi, Marcus Rohrbach, and Tinne Tuytelaars.
\newblock Selfless sequential learning.
\newblock {\em Proceedings of the International Conference on Learning
  Representations (ICLR)}, 2018.

\bibitem{castro2018end}
Francisco~M Castro, Manuel~J Mar{\'\i}n-Jim{\'e}nez, Nicol{\'a}s Guil, Cordelia
  Schmid, and Karteek Alahari.
\newblock End-to-end incremental learning.
\newblock In {\em Proceedings of the European Conference on Computer Vision
  (ECCV)}, pages 233--248, 2018.

\bibitem{cermelli2020modeling}
Fabio Cermelli, Massimiliano Mancini, Samuel~Rota Bul{\`o}, Elisa Ricci, and
  Barbara Caputo.
\newblock Modeling the background for incremental learning in semantic
  segmentation.
\newblock In {\em Proceedings of the IEEE Conference on Computer Vision and
  Pattern Recognition (CVPR)}, 2020.

\bibitem{chaudhry2018riemannian}
Arslan Chaudhry, Puneet~K Dokania, Thalaiyasingam Ajanthan, and Philip~HS Torr.
\newblock Riemannian walk for incremental learning: Understanding forgetting
  and intransigence.
\newblock In {\em Proceedings of the European Conference on Computer Vision
  (ECCV)}, pages 532--547, 2018.

\bibitem{chen2018deeplab}
Liang-Chieh Chen, George Papandreou, Iasonas Kokkinos, Kevin Murphy, and Alan~L
  Yuille.
\newblock Deeplab: Semantic image segmentation with deep convolutional nets,
  atrous convolution, and fully connected crfs.
\newblock {\em IEEE Transactions on Pattern Analysis and Machine Intelligence
  (PAMI)}, 40(4):834--848, 2018.

\bibitem{chen2017rethinking}
Liang-Chieh Chen, George Papandreou, Florian Schroff, and Hartwig Adam.
\newblock Rethinking atrous convolution for semantic image segmentation.
\newblock {\em arXiv preprint arXiv:1706.05587}, 2017.

\bibitem{deeplabv3plus2018}
Liang-Chieh Chen, Yukun Zhu, George Papandreou, Florian Schroff, and Hartwig
  Adam.
\newblock Encoder-decoder with atrous separable convolution for semantic image
  segmentation.
\newblock In {\em Proceedings of the European Conference on Computer Vision
  (ECCV)}, 2018.

\bibitem{chen2020simple}
Ting Chen, Simon Kornblith, Mohammad Norouzi, and Geoffrey Hinton.
\newblock A simple framework for contrastive learning of visual
  representations.
\newblock {\em Proceedings of the International Conference on Machine Learning
  (ICML)}, 2020.

\bibitem{delange2019continual}
Matthias De~Lange, Rahaf Aljundi, Marc Masana, Sarah Parisot, Xu Jia, Ales
  Leonardis, Gregory Slabaugh, and Tinne Tuytelaars.
\newblock Continual learning: A comparative study on how to defy forgetting in
  classification tasks.
\newblock {\em arXiv preprint arXiv:1909.08383}, 2019.

\bibitem{deng2009imagenet}
Jia Deng, Wei Dong, Richard Socher, Li-Jia Li, Kai Li, and Li Fei-Fei.
\newblock Imagenet: A large-scale hierarchical image database.
\newblock In {\em Proceedings of the IEEE Conference on Computer Vision and
  Pattern Recognition (CVPR)}, pages 248--255. Ieee, 2009.

\bibitem{deng2019cluster}
Zhijie Deng, Yucen Luo, and Jun Zhu.
\newblock Cluster alignment with a teacher for unsupervised domain adaptation.
\newblock In {\em Proceedings of the International Conference on Computer
  Vision (ICCV)}, pages 9944--9953, 2019.

\bibitem{dhar2018learning}
Prithviraj Dhar, Rajat~Vikram Singh, Kuan-Chuan Peng, Ziyan Wu, and Rama
  Chellappa.
\newblock Learning without memorizing.
\newblock {\em arXiv preprint arXiv:1811.08051}, 2018.

\bibitem{doersch2017multi}
Carl Doersch and Andrew Zisserman.
\newblock Multi-task self-supervised visual learning.
\newblock In {\em Proceedings of the International Conference on Computer
  Vision (ICCV)}, pages 2051--2060, 2017.

\bibitem{dong2018few}
Nanqing Dong and Eric~P Xing.
\newblock Few-shot semantic segmentation with prototype learning.
\newblock In {\em Proceedings of the British Machine Vision Conference (BMVC)},
  volume~3, 2018.

\bibitem{pascalvoc2012}
M. Everingham, L. Van~Gool, C.~K.~I. Williams, J. Winn, and A. Zisserman.
\newblock {The {PASCAL} {V}isual {O}bject {C}lasses {C}hallenge 2012
  {R}esults}, 2012.

\bibitem{fernando2017pathnet}
Chrisantha Fernando, Dylan Banarse, Charles Blundell, Yori Zwols, David Ha,
  Andrei~A Rusu, Alexander Pritzel, and Daan Wierstra.
\newblock Pathnet: Evolution channels gradient descent in super neural
  networks.
\newblock {\em arXiv preprint arXiv:1701.08734}, 2017.

\bibitem{french1999catastrophic}
Robert~M French.
\newblock Catastrophic forgetting in connectionist networks.
\newblock {\em Trends in Cognitive Sciences}, 3(4):128--135, 1999.

\bibitem{garcia2018survey}
Alberto Garcia-Garcia, Sergio Orts-Escolano, Sergiu Oprea, Victor
  Villena-Martinez, Pablo Martinez-Gonzalez, and Jose Garcia-Rodriguez.
\newblock A survey on deep learning techniques for image and video semantic
  segmentation.
\newblock {\em Applied Soft Computing}, 70:41--65, 2018.

\bibitem{goodfellow2013empirical}
Ian~J Goodfellow, Mehdi Mirza, Da Xiao, Aaron Courville, and Yoshua Bengio.
\newblock An empirical investigation of catastrophic forgetting in
  gradient-based neural networks.
\newblock {\em Proceedings of the International Conference on Learning
  Representations (ICLR)}, 2014.

\bibitem{guo2018review}
Yanming Guo, Yu Liu, Theodoros Georgiou, and Michael~S Lew.
\newblock A review of semantic segmentation using deep neural networks.
\newblock {\em {International Journal of Multimedia Information Retrieval}},
  7(2):87--93, 2018.

\bibitem{hadsell2006dimensionality}
Raia Hadsell, Sumit Chopra, and Yann LeCun.
\newblock Dimensionality reduction by learning an invariant mapping.
\newblock In {\em Proceedings of the IEEE Conference on Computer Vision and
  Pattern Recognition (CVPR)}, volume~2, pages 1735--1742. IEEE, 2006.

\bibitem{he2020momentum}
Kaiming He, Haoqi Fan, Yuxin Wu, Saining Xie, and Ross Girshick.
\newblock Momentum contrast for unsupervised visual representation learning.
\newblock In {\em Proceedings of the IEEE Conference on Computer Vision and
  Pattern Recognition (CVPR)}, pages 9729--9738, 2020.

\bibitem{he2016deep}
Kaiming He, Xiangyu Zhang, Shaoqing Ren, and Jian Sun.
\newblock Deep residual learning for image recognition.
\newblock In {\em Proceedings of the IEEE Conference on Computer Vision and
  Pattern Recognition (CVPR)}, pages 770--778, 2016.

\bibitem{hinton2015distilling}
Geoffrey Hinton, Oriol Vinyals, and Jeff Dean.
\newblock Distilling the knowledge in a neural network.
\newblock {\em Neural Information Processing Systems, Deep Learning and
  Representation Learning Workshop}, 2015.

\bibitem{hou2019learning}
Saihui Hou, Xinyu Pan, Chen~Change Loy, Zilei Wang, and Dahua Lin.
\newblock Learning a unified classifier incrementally via rebalancing.
\newblock In {\em Proceedings of the IEEE Conference on Computer Vision and
  Pattern Recognition (CVPR)}, pages 831--839, 2019.

\bibitem{javed2019meta}
Khurram Javed and Martha White.
\newblock Meta-learning representations for continual learning.
\newblock In {\em Neural Information Processing Systems (NeurIPS)}, pages
  1820--1830, 2019.

\bibitem{ji2019invariant}
Xu Ji, Jo{\~a}o~F Henriques, and Andrea Vedaldi.
\newblock Invariant information clustering for unsupervised image
  classification and segmentation.
\newblock In {\em Proceedings of the International Conference on Computer
  Vision (ICCV)}, pages 9865--9874, 2019.

\bibitem{kang2019contrastive}
Guoliang Kang, Lu Jiang, Yi Yang, and Alexander~G Hauptmann.
\newblock Contrastive adaptation network for unsupervised domain adaptation.
\newblock In {\em Proceedings of the IEEE Conference on Computer Vision and
  Pattern Recognition (CVPR)}, pages 4893--4902, 2019.

\bibitem{kemker20178measuring}
Ronald Kemker, Marc McClure, Angelina Abitino, Tyler Hayes, and Christopher
  Kanan.
\newblock Measuring catastrophic forgetting in neural networks.
\newblock In {\em Proceedings of the AAAI Conference on Artificial
  Intelligence}, 2018.

\bibitem{khosla2020supervised}
Prannay Khosla, Piotr Teterwak, Chen Wang, Aaron Sarna, Yonglong Tian, Phillip
  Isola, Aaron Maschinot, Ce Liu, and Dilip Krishnan.
\newblock Supervised contrastive learning.
\newblock {\em arXiv preprint arXiv:2004.11362}, 2020.

\bibitem{kirkpatrick2017overcoming}
James Kirkpatrick, Razvan Pascanu, Neil Rabinowitz, Joel Veness, Guillaume
  Desjardins, Andrei~A Rusu, Kieran Milan, John Quan, Tiago Ramalho, Agnieszka
  Grabska-Barwinska, Demis Hassabis, Claudia Clopath, Dharshan Kumaran, and
  Raia Hadsell.
\newblock Overcoming catastrophic forgetting in neural networks.
\newblock {\em Proceedings of the National Academy of Sciences (PNAS)},
  114(13):3521--3526, 2017.

\bibitem{klingner2020class}
Marvin Klingner, Andreas B{\"a}r, Philipp Donn, and Tim Fingscheidt.
\newblock Class-incremental learning for semantic segmentation re-using neither
  old data nor old labels.
\newblock {\em International Conference on Intelligent Transportation Systems},
  2020.

\bibitem{li2019rilod}
Dawei Li, Serafettin Tasci, Shalini Ghosh, Jingwen Zhu, Junting Zhang, and
  Larry Heck.
\newblock Rilod: near real-time incremental learning for object detection at
  the edge.
\newblock In {\em Proceedings of the 4th ACM/IEEE Symposium on Edge Computing},
  pages 113--126, 2019.

\bibitem{li2019learn}
Xilai Li, Yingbo Zhou, Tianfu Wu, Richard Socher, and Caiming Xiong.
\newblock Learn to grow: A continual structure learning framework for
  overcoming catastrophic forgetting.
\newblock In {\em Proceedings of the International Conference on Machine
  Learning (ICML)}, 2019.

\bibitem{li2018learning}
Zhizhong Li and Derek Hoiem.
\newblock Learning without forgetting.
\newblock {\em IEEE Transactions on Pattern Analysis and Machine Intelligence
  (PAMI)}, 40(12):2935--2947, 2018.

\bibitem{liang2019distant}
Jian Liang, Ran He, Zhenan Sun, and Tieniu Tan.
\newblock Distant supervised centroid shift: A simple and efficient approach to
  visual domain adaptation.
\newblock In {\em Proceedings of the IEEE Conference on Computer Vision and
  Pattern Recognition (CVPR)}, pages 2975--2984, 2019.

\bibitem{long2015fully}
Jonathan Long, Evan Shelhamer, and Trevor Darrell.
\newblock Fully convolutional networks for semantic segmentation.
\newblock In {\em Proceedings of the IEEE Conference on Computer Vision and
  Pattern Recognition (CVPR)}, pages 3431--3440, 2015.

\bibitem{mallya2018piggyback}
Arun Mallya, Dillon Davis, and Svetlana Lazebnik.
\newblock Piggyback: Adapting a single network to multiple tasks by learning to
  mask weights.
\newblock In {\em Proceedings of the European Conference on Computer Vision
  (ECCV)}, pages 67--82, 2018.

\bibitem{mallya2018packnet}
Arun Mallya and Svetlana Lazebnik.
\newblock Packnet: Adding multiple tasks to a single network by iterative
  pruning.
\newblock In {\em Proceedings of the IEEE Conference on Computer Vision and
  Pattern Recognition}, pages 7765--7773, 2018.

\bibitem{mccloskey1989catastrophic}
Michael McCloskey and Neal~J Cohen.
\newblock Catastrophic interference in connectionist networks: The sequential
  learning problem.
\newblock In {\em Psychology of learning and motivation}, volume~24, pages
  109--165. Elsevier, 1989.

\bibitem{michieli2019incremental}
Umberto Michieli and Pietro Zanuttigh.
\newblock {Incremental Learning Techniques for Semantic Segmentation}.
\newblock In {\em Proceedings of the International Conference on Computer
  Vision Workshops (ICCVW)}, 2019.

\bibitem{michieli2020knowledge}
Umberto Michieli and Pietro Zanuttigh.
\newblock Knowledge distillation for incremental learning in semantic
  segmentation.
\newblock {\em Computer Vision and Image Understanding}, 205:103167, 2021.

\bibitem{misra2020self}
Ishan Misra and Laurens van~der Maaten.
\newblock Self-supervised learning of pretext-invariant representations.
\newblock In {\em Proceedings of the IEEE Conference on Computer Vision and
  Pattern Recognition (CVPR)}, pages 6707--6717, 2020.

\bibitem{oreshkin2018tadam}
Boris Oreshkin, Pau Rodr{\'\i}guez~L{\'o}pez, and Alexandre Lacoste.
\newblock Tadam: Task dependent adaptive metric for improved few-shot learning.
\newblock {\em Neural Information Processing Systems (NeurIPS)}, 31:721--731,
  2018.

\bibitem{ostapenko2019learning}
Oleksiy Ostapenko, Mihai Puscas, Tassilo Klein, Patrick Jahnichen, and Moin
  Nabi.
\newblock Learning to remember: A synaptic plasticity driven framework for
  continual learning.
\newblock In {\em Proceedings of the IEEE Conference on Computer Vision and
  Pattern Recognition (CVPR)}, pages 11321--11329, 2019.

\bibitem{ozdemir2019extending}
Firat Ozdemir and Orcun Goksel.
\newblock Extending pretrained segmentation networks with additional anatomical
  structures.
\newblock {\em International journal of computer assisted radiology and
  surgery}, 14(7):1187--1195, 2019.

\bibitem{parisi2019continual}
German~I Parisi, Ronald Kemker, Jose~L Part, Christopher Kanan, and Stefan
  Wermter.
\newblock Continual lifelong learning with neural networks: A review.
\newblock {\em Neural Networks}, 2019.

\bibitem{peng2019domain}
Xingchao Peng, Zijun Huang, Ximeng Sun, and Kate Saenko.
\newblock Domain agnostic learning with disentangled representations.
\newblock {\em Proceedings of the International Conference on Machine Learning
  (ICML)}, 2019.

\bibitem{pinheiro2018unsupervised}
Pedro~O Pinheiro.
\newblock Unsupervised domain adaptation with similarity learning.
\newblock In {\em Proceedings of the IEEE Conference on Computer Vision and
  Pattern Recognition (CVPR)}, pages 8004--8013, 2018.

\bibitem{rebuffi2017icarl}
Sylvestre-Alvise Rebuffi, Alexander Kolesnikov, Georg Sperl, and Christoph~H
  Lampert.
\newblock icarl: Incremental classifier and representation learning.
\newblock In {\em Proceedings of the IEEE Conference on Computer Vision and
  Pattern Recognition (CVPR)}, pages 2001--2010, 2017.

\bibitem{schwarz2018progress}
Jonathan Schwarz, Jelena Luketina, Wojciech~M Czarnecki, Agnieszka
  Grabska-Barwinska, Yee~Whye Teh, Razvan Pascanu, and Raia Hadsell.
\newblock Progress \& compress: A scalable framework for continual learning.
\newblock In {\em Proceedings of the International Conference on Machine
  Learning (ICML)}, 2018.

\bibitem{serra2018overcoming}
Joan Serra, Didac Suris, Marius Miron, and Alexandros Karatzoglou.
\newblock Overcoming catastrophic forgetting with hard attention to the task.
\newblock In {\em Proceedings of the International Conference on Machine
  Learning (ICML)}, 2018.

\bibitem{shekhar2013generalized}
Sumit Shekhar, Vishal~M Patel, Hien~V Nguyen, and Rama Chellappa.
\newblock Generalized domain-adaptive dictionaries.
\newblock In {\em Proceedings of the IEEE Conference on Computer Vision and
  Pattern Recognition (CVPR)}, pages 361--368, 2013.

\bibitem{shin2017continual}
Hanul Shin, Jung~Kwon Lee, Jaehong Kim, and Jiwon Kim.
\newblock Continual learning with deep generative replay.
\newblock In {\em Neural Information Processing Systems (NeurIPS)}, pages
  2990--2999, 2017.

\bibitem{shmelkov2017incremental}
Konstantin Shmelkov, Cordelia Schmid, and Karteek Alahari.
\newblock Incremental learning of object detectors without catastrophic
  forgetting.
\newblock In {\em Proceedings of the International Conference on Computer
  Vision (ICCV)}, pages 3400--3409, 2017.

\bibitem{snell2017prototypical}
Jake Snell, Kevin Swersky, and Richard Zemel.
\newblock Prototypical networks for few-shot learning.
\newblock In {\em Neural Information Processing Systems (NeurIPS)}, pages
  4077--4087, 2017.

\bibitem{tasar2018incremental}
Onur Tasar, Yuliya Tarabalka, and Pierre Alliez.
\newblock {Incremental Learning for Semantic Segmentation of Large-Scale Remote
  Sensing Data}.
\newblock {\em arXiv preprint arXiv:1810.12448}, 2018.

\bibitem{tian2019contrastive}
Yonglong Tian, Dilip Krishnan, and Phillip Isola.
\newblock Contrastive multiview coding.
\newblock {\em arXiv preprint arXiv:1906.05849}, 2019.

\bibitem{tian2020generalized}
Zhuotao Tian, Xin Lai, Li Jiang, Michelle Shu, Hengshuang Zhao, and Jiaya Jia.
\newblock Generalized few-shot semantic segmentation.
\newblock {\em arXiv preprint arXiv:2010.05210}, 2020.

\bibitem{toldo2021unsupervised}
Marco Toldo, Umberto Michieli, and Pietro Zanuttigh.
\newblock Unsupervised domain adaptation in semantic segmentation via
  orthogonal and clustered embeddings.
\newblock In {\em Proceedings of the IEEE/CVF Winter Conference on Applications
  of Computer Vision (WACV)}, pages 1358--1368, 2021.

\bibitem{vu2019advent}
Tuan-Hung Vu, Himalaya Jain, Maxime Bucher, Matthieu Cord, and Patrick
  P{\'e}rez.
\newblock Advent: Adversarial entropy minimization for domain adaptation in
  semantic segmentation.
\newblock In {\em Proceedings of the IEEE Conference on Computer Vision and
  Pattern Recognition (CVPR)}, pages 2517--2526, 2019.

\bibitem{wang2019panet}
Kaixin Wang, Jun~Hao Liew, Yingtian Zou, Daquan Zhou, and Jiashi Feng.
\newblock Panet: Few-shot image semantic segmentation with prototype alignment.
\newblock In {\em Proceedings of the International Conference on Computer
  Vision (ICCV)}, pages 9197--9206, 2019.

\bibitem{wang2017growing}
Yu-Xiong Wang, Deva Ramanan, and Martial Hebert.
\newblock Growing a brain: Fine-tuning by increasing model capacity.
\newblock In {\em Proceedings of the IEEE Conference on Computer Vision and
  Pattern Recognition (CVPR)}, pages 2471--2480, 2017.

\bibitem{wu2019improving}
Si Wu, Jian Zhong, Wenming Cao, Rui Li, Zhiwen Yu, and Hau-San Wong.
\newblock Improving domain-specific classification by collaborative learning
  with adaptation networks.
\newblock In {\em Proceedings of the AAAI Conference on Artificial
  Intelligence}, volume~33, pages 5450--5457, 2019.

\bibitem{wu2018incremental}
Yue Wu, Yinpeng Chen, Lijuan Wang, Yuancheng Ye, Zicheng Liu, Yandong Guo,
  Zhengyou Zhang, and Yun Fu.
\newblock Incremental classifier learning with generative adversarial networks.
\newblock {\em arXiv preprint arXiv:1802.00853}, 2018.

\bibitem{wu2018unsupervised}
Zhirong Wu, Yuanjun Xiong, Stella~X Yu, and Dahua Lin.
\newblock Unsupervised feature learning via non-parametric instance
  discrimination.
\newblock In {\em Proceedings of the IEEE Conference on Computer Vision and
  Pattern Recognition (CVPR)}, pages 3733--3742, 2018.

\bibitem{xian2016latent}
Yongqin Xian, Zeynep Akata, Gaurav Sharma, Quynh Nguyen, Matthias Hein, and
  Bernt Schiele.
\newblock Latent embeddings for zero-shot classification.
\newblock In {\em Proceedings of the IEEE Conference on Computer Vision and
  Pattern Recognition (CVPR)}, pages 69--77, 2016.

\bibitem{xiao2014error}
Tianjun Xiao, Jiaxing Zhang, Kuiyuan Yang, Yuxin Peng, and Zheng Zhang.
\newblock Error-driven incremental learning in deep convolutional neural
  network for large-scale image classification.
\newblock In {\em Proceedings of the ACM International Conference on
  Multimedia}, pages 177--186. ACM, 2014.

\bibitem{xie2018learning}
Shaoan Xie, Zibin Zheng, Liang Chen, and Chuan Chen.
\newblock Learning semantic representations for unsupervised domain adaptation.
\newblock In {\em Proceedings of the International Conference on Machine
  Learning (ICML)}, pages 5423--5432, 2018.

\bibitem{ye2019unsupervised}
Mang Ye, Xu Zhang, Pong~C Yuen, and Shih-Fu Chang.
\newblock Unsupervised embedding learning via invariant and spreading instance
  feature.
\newblock In {\em Proceedings of the IEEE Conference on Computer Vision and
  Pattern Recognition (CVPR)}, pages 6210--6219, 2019.

\bibitem{yu2017dilated}
Fisher Yu, Vladlen Koltun, and Thomas Funkhouser.
\newblock Dilated residual networks.
\newblock In {\em Proceedings of the IEEE Conference on Computer Vision and
  Pattern Recognition (CVPR)}, pages 472--480, 2017.

\bibitem{zenke2017continual}
Friedemann Zenke, Ben Poole, and Surya Ganguli.
\newblock Continual learning through synaptic intelligence.
\newblock In {\em Proceedings of the International Conference on Machine
  Learning (ICML)}, pages 3987--3995, 2017.

\bibitem{zhang2015domain}
Heng Zhang, Vishal~M Patel, Sumit Shekhar, and Rama Chellappa.
\newblock Domain adaptive sparse representation-based classification.
\newblock In {\em {IEEE International Conference and Workshops on Automatic
  Face and Gesture Recognition (FG)}}, volume~1, pages 1--8. IEEE, 2015.

\bibitem{zhao2017pyramid}
Hengshuang Zhao, Jianping Shi, Xiaojuan Qi, Xiaogang Wang, and Jiaya Jia.
\newblock Pyramid scene parsing network.
\newblock In {\em Proceedings of the IEEE Conference on Computer Vision and
  Pattern Recognition (CVPR)}, pages 2881--2890, 2017.

\bibitem{zhou2017scene}
Bolei Zhou, Hang Zhao, Xavier Puig, Sanja Fidler, Adela Barriuso, and Antonio
  Torralba.
\newblock Scene parsing through ade20k dataset.
\newblock In {\em Proceedings of the IEEE Conference on Computer Vision and
  Pattern Recognition (CVPR)}, pages 633--641, 2017.

\bibitem{zhuang2019local}
Chengxu Zhuang, Alex~Lin Zhai, and Daniel Yamins.
\newblock Local aggregation for unsupervised learning of visual embeddings.
\newblock In {\em Proceedings of the International Conference on Computer
  Vision (ICCV)}, pages 6002--6012, 2019.

\end{thebibliography}
}


\pagebreak
\pagebreak

\begin{center}
  \textbf{\large Continual Semantic Segmentation via Repulsion-Attraction of Sparse and Disentangled Latent Representations \\ \vspace{0.1cm} \textit{Supplementary Material}}\\[.2cm]
  Umberto Michieli and Pietro Zanuttigh\\[.1cm]
  {\itshape University of Padova,\\ Department of Information Engineering
  }
\end{center} 

\setcounter{equation}{0}
\setcounter{figure}{0}
\setcounter{table}{0}
\setcounter{page}{1}
\renewcommand{\theequation}{S\arabic{equation}}
\renewcommand{\thefigure}{S\arabic{figure}}
\renewcommand{\thetable}{S\arabic{table}}

In this document we present some additional material to better motivate our method and we conduct some supplementary experiments.
More in detail, we start by discussing some of the design choices that led to the models of the losses and constraints presented in the main paper in Section \ref{sec:design}.  Then, Section \ref{sec:ablation} shows some additional ablation studies. Finally, many additional qualitative and quantitative results 
for both the Pascal VOC2012 and the ADE20K datasets are presented in Sections \ref{sec:qualitative}, \ref{sec:steps} and \ref{sec:quantitative}.

\section{Design Choices}
\label{sec:design}

In this section we present some additional discussion and results motivating the design choices behind the various modules exploited in our work.

\noindent \textbf{Prototypes Matching} enforces latent space consistency on old classes, forcing the encoder to produce similar latent representation for previously seen classes in the subsequent steps. The target is achieved by considering the  Euclidean distance in the latent space (see Section 4.1 of the paper). Although different distance metrics could have been used in principle (\eg, cosine distance \cite{wang2019panet,snell2017prototypical,oreshkin2018tadam}) we found that a simple Euclidean distance was easier to understand and very computationally efficient results similar to more complex schemes.

\noindent \textbf{Contrastive Learning} aims at clustering features according to their semantics while tearing apart those of different classes (see Section 4.2 of the paper): we implement it as an attractive force between latent representations with their prototypical representation, against a repulsive one between prototypes of different semantic categories. This attraction-repulsion rule is enforced again using an Euclidean distance metric. 

\noindent \textbf{Knowledge Distillation} is employed to constraint the decoder to preserve previous knowledge at the output-level and it is implemented as a standard cross-entropy on the output softmax probabilities between old model and current model predictions \cite{michieli2019incremental,michieli2020knowledge,cermelli2020modeling,klingner2020class} (see Section 4.4 of the paper). 

\noindent \textbf{Sparsity}: We think that the most peculiar constraint is represented by the  sparsity objective. However, the underlying concept is simple: applying some latent-level sparsification we allow the model to retain enough discriminative power to accommodate the upcoming representations of novel classes without cross-talk with previous ones (see Section 4.3). Here, a wide range of possibilities could be considered to address the aforementioned task and one may wonder why the sparsity constraint was designed as it is presented in the main paper. {\rev In this document, we present an empirically-driven ablation study to understand the behavior of our constraints.}
Common sparsity losses are the L0 or L1 norms of feature vectors; however, we show that they achieve lower accuracy.
In this work, we define the sparsity objective as the ratio between a stretching function (\ie, the sum of exponentials) and a linear function (\ie, the sum) applied to the feature vectors which were previously normalized with respect to the maximum value that is assumed by any of the feature channels for that particular class. 
In some extreme cases, the model of Eq.~(9) could lead to degenerate solutions, however we argue that these do not happen in practice on a model learning compact representations. We checked to avoid the zero division in the practical implementation, while the \textit{all-ones} case is degenerate in the sense that energy cannot be re-distributed in any way since all channels are already onset to the maximum value and, furthermore, this configuration would not be informative for the decoder.

{\rev First, we observe that the normalization with respect to the class-wise maximum value is an important step of the algorithm, as it gives the same importance to each class. We believe that such normalization is fair (sometimes, features of few particular classes may just be on average more active than features of other classes); however, we can think of getting rid of it and  normalizing with other strategies, \eg, with respect to:}
\begin{itemize} 
\item the maximum value for each feature (\textit{norm max});
\item the overall maximum value (\textit{norm max overall});
\item the \textit{L2} norm of each feature (\textit{norm L2}). 
\end{itemize}
In such cases, Eq. (8) would become respectively:
\begin{equation}
\mathbf{\bar{f}}_i =  
\frac{\mathbf{f}_i}
{    \max_{f_{i,j} \in \mathbf{f}_i } f_{i,j} }
\ \ \ \ \ \ \ \ \ \
\mathbf{f}_i \in \mathbf{F}_n 
\end{equation}
\begin{equation}
\mathbf{\bar{f}}_i =  
\frac{\mathbf{f}_i}
{    \max_{g_{j,l} \in \mathbf{g}_j } g_{j,l} }
\ \ \ \ \ \ \ \ \ \
\mathbf{f}_i , \mathbf{g}_j \in \mathbf{F}_n 
\end{equation}
\begin{equation}
\mathbf{\bar{f}}_i =  
\frac{\mathbf{f}_i}
{    || \mathbf{f}_i ||_2 }
\ \ \ \ \ \ \ \ \ \
\mathbf{f}_i \in \mathbf{F}_n 
\end{equation}

Furthermore, in principle any stretching function could be applied in spite of the sum of exponentials over the linear sum. For instance, the sum of squares (\textit{power 2}) or sum of the cubic powers (\textit{power 3}) could be used as stretching functions: \ie, formulating Eq. (9) respectively as:

\begin{equation}
\mathcal{L}_{sp} =  \frac{1}{|       \mathbf{f}_{i} \in \mathbf{F}_n     |}
\sum_{\mathbf{f}_{i} \in \mathbf{F}_n}
\frac{\sum_j \bar{f}_{i,j}^2}
{ \sum_j \bar{f}_{i,j}}
\end{equation}
\begin{equation}
\mathcal{L}_{sp} =  \frac{1}{|       \mathbf{f}_{i} \in \mathbf{F}_n     |}
\sum_{\mathbf{f}_{i} \in \mathbf{F}_n}
\frac{\sum_j \bar{f}_{i,j}^3}
{ \sum_j \bar{f}_{i,j}}.
\end{equation}

Finally, following the success of recent works exploiting entropy minimization \cite{vu2019advent}  techniques, an alternative strategy could be to minimize the entropy of the latent representations  opportunely preceded by \textit{L1} or \textit{softmax} normalization of each feature vector in order to obtain a probability distribution over the channels. More formally:
\begin{equation}
\mathbf{\bar{f}}_i =  
\frac{\mathbf{f}_i}
{    || \mathbf{f}_i ||_1 }
\ \ \ \ \ \ \ \ \ \
\mathbf{f}_i \in \mathbf{F}_n 
\end{equation}
\begin{equation}
\mathbf{\bar{f}}_i =  
\frac{\exp\left(\mathbf{f}_i \right)}
{    \sum_j \exp \left( f_{i,j}  \right) }
\ \ \ \ \ \ \ \ \ \
\mathbf{f}_i \in \mathbf{F}_n 
\end{equation}

\begin{equation}
\mathcal{L}_{sp} =  \frac{1}{|       \mathbf{f}_{i} \in \mathbf{F}_n     |}
\sum_{\mathbf{f}_{i} \in \mathbf{F}_n}
\sum_j - \bar{f}_{i,j} \cdot \log \left(   \bar{f}_{i,j}   \right)
\end{equation}

Table~\ref{tab:sparsity} shows the performance of the aforementioned approaches in the 19-1 and 15-1 disjoint scenarios  on Pascal VOC2012. Different normalization rules bring to consistently lower results, proving the efficacy of using class guidance during normalization. Also, different stretching functions are found to be less adequate for our purpose reducing the final mIoU of about $2\%$ to $4\%$. Finally, entropy minimization techniques obtain competitive and comparable results in the $15-1$ scenario, while they experience a drop of about $2-3\%$ of mIoU when only one class is added.

\begin{table}[h]
\caption{Comparison of different $\mathcal{L}_{sp}$ in terms of mIoU in the disjoint scenarios 19-1 and 15-1 on Pascal VOC2012 dataset.}
\vspace{0.1cm}
\label{tab:sparsity}
\setlength{\tabcolsep}{6pt}
\centering
\begin{tabular}{|l|c|c|}
\hline
Method & mIoU$_{19-1}$ & mIoU$_{15-1}$ \\\hline
\textit{L0} & 66.7 & 46.3 \\
\textit{L1} & 65.9 & 45.4 \\\hdashline
\textit{norm max} & 67.4 & 47.8 \\
\textit{norm max overall} & 67.5 & 45.6 \\
\textit{norm L2} & 64.8 & 44.3 \\\hdashline
\textit{power 2} & 66.3 & 44.2 \\
\textit{power 3} & 66.6 & 45.3 \\\hdashline
\textit{entropy (L1)} & 65.3 & 48.0 \\
\textit{entropy (softmax)} & 66.0 & 48.0\\\hline
\textit{ours} & \textbf{68.4} & \textbf{48.1} \\\hline
\end{tabular}
\end{table}

\begin{figure}[ht]
\centering
\includegraphics[trim={0.7cm 0.15cm 1.4cm 0.5cm}, clip, width=\linewidth]{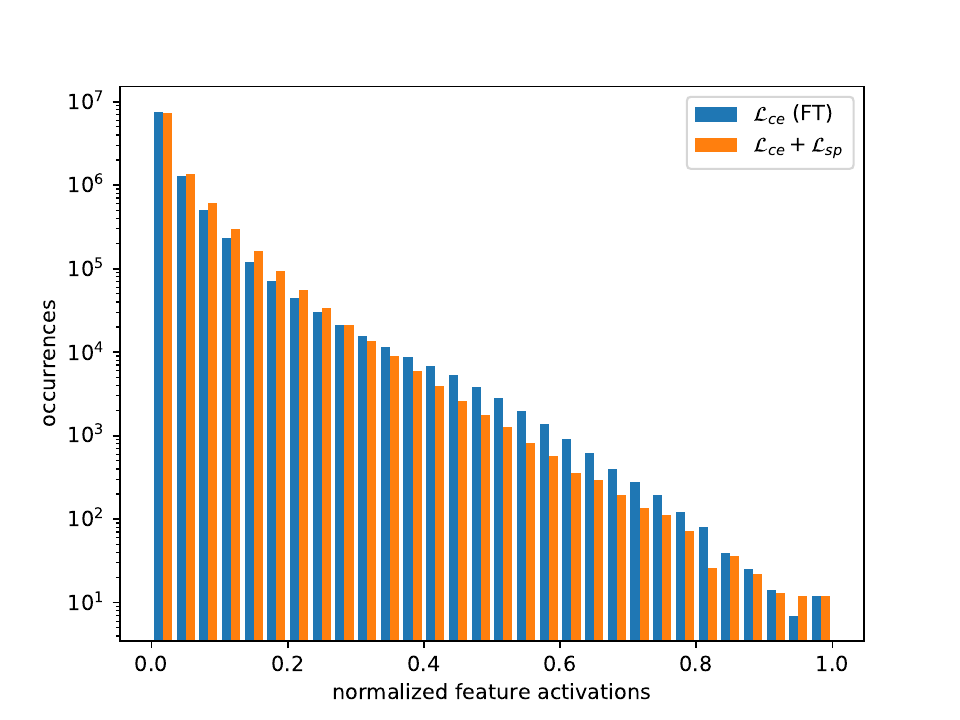}
\caption{\rev{Histogram of occurrences of feature activations. Features are normalized with respect to the class-wise maximum value over all feature channels (Eq.~\eqref{eq:features_normalization}).}}
\label{fig:histogram_spa}
\end{figure}

\rev{Finally, in Figure~\ref{fig:histogram_spa} we show the effect of the proposed module on the feature vectors. The plot shows the feature distribution of the fine-tuning (FT) approach against the addition of the sparsity regularizing constraint. We can appreciate how adding $\mathcal{L}_{sp}$ reduced spurious activations in the range $0.3-0.85$, that correspond to uncertain and unreliable predictions.}

\section{Additional Ablation Studies}
\label{sec:ablation}

In this section, we report a couple of additional ablation studies concerning the dataset size and the pre-training. 

\textbf{Random Split.} 
Looking at Table~1 of the main paper, we see that in some cases, especially on the 15-1 setup, the proposed method is still far from the offline reference. An interesting question is whether this is due to the difficulty of handling new classes or if, more fundamentally, it is due to an inherent difficulty to train a network using only a small subset of the data at each step. To answer this, we split the dataset equally in $5$ parts (each part containing all classes, thus removing the complexity of learning new classes) and then we trained the network sequentially on each of this parts. We obtained $69.9\%$ of mIoU against $75.4\%$ of the joint training, $5.6\%$ of the FT (disjoint) and $48.1\%$ of SDR (disjoint). The difference with respect to joint training is relatively small, and it could be due to sub-optimal network weights estimation (samples are taken from the $5$ parts accessed subsequently, instead of the full dataset); on the other side, the difference with respect to FT is very large proving that handling unseen classes is the key issue and the proposed latent constraints aim at addressing it.

\textbf{Considerations on Pre-Training.}
The results reported in the main paper have been obtained initializing the weights of the backbone ResNet-101 approach on the ImageNet dataset. This is the standard setup in continual semantic segmentation approaches \cite{michieli2019incremental,cermelli2020modeling,michieli2020knowledge,klingner2020class}. Additional considerations have been already addressed in \cite{michieli2020knowledge}, where it has been shown that pre-training on a segmentation benchmark could boost the accuracy; nonetheless, the ranking of the proposed strategies is mainly maintained.

On the other hand, even ImageNet contains visual samples for many of the elements present in the Pascal dataset (for classification task instead of segmentation), potentially  limiting the magnitude of decay on \textit{old} tasks, and likely raising accuracies for \textit{new} concepts that are not necessarily completely new to the encoder. 
Here, we show how the network performs without such a strong prior on the latent representations. The results are strongly affected by the fact that datasets for in-the-wild segmentation are often too small to reliably train complex deep networks from scratch. We trained on VOC2012 without pre-training and we achieved a low mIoU of $24.4\%$ when training for $30$ epochs, as we do in the main paper, and $40.9\%$, when training for $120$ epochs (about $30$ hours of computation). In continual learning, the final mIoU are also lower (as the starting point is much lower), but the improvements achieved by our approach and the ranking of the various methods are preserved, for instance in VOC2012 15-1 disjoint the accuracy of SDR ($13.5\%$) is still significantly above FT ($4.1\%$) and MiB ($10.9\%$).


\section{Additional Qualitative Results}
\label{sec:qualitative}

%
%
%
%

Many qualitative experimental results are reported for all the different scenarios, experimental protocols (\ie, sequential, disjoint and overlapped) and datasets.

\textbf{Pascal VOC2012.} The results for this dataset are reported in Figures~\ref{fig:voc_qualitative_sequential}, \ref{fig:voc_qualitative_disjoint} and \ref{fig:voc_qualitative_overlapped} respectively for sequential, disjoint and overlapped protocols. In each figure, 3 images for each scenario (\ie, 19-1, 15-5 and 15-1) are depicted. We compare our method with na\"ive fine tuning and the competitors, \ie, LwF 
 \cite{li2018learning}, 
  ILT \cite{michieli2019incremental}, CIL \cite{klingner2020class} and MiB \cite{cermelli2020modeling}.
   The images  show how our approach alleviates forgetting and at the same time accommodates new classes to learn. On the other side, the fine-tuning and the compared approaches often deviate (\ie, are biased) in predicting novel classes being added or the special background class.

\textbf{ADE20K.} We report several visual results in Figure~\ref{fig:ade_qualitative} also for this dataset. In particular, we show 3 images for each scenario (\ie, 100-50, 100-10, 50-50). Again, we can appreciate how our method largely outperforms compared approaches in all scenarios better capturing the details of the shapes of the objects (e.g, in rows 1-4) and not degenerating into an overestimation of the background  (\eg, in the 100-10 scenario). In particular, we notice how compared approaches have big difficulties in  handling multiple additions of multiple classes (they struggle in tackling catastrophic forgetting in the 100-10 scenario), while our method can achieve reasonably good output segmentation maps also in the most challenging scenarios.

\section{Qualitative Results Across Incremental Steps}
\label{sec:steps}

In this section we analyze the  performance across the various incremental steps, comparing our method with the top performing competitor (\ie, MiB \cite{cermelli2020modeling}). 

\textbf{Pascal VOC2012.} The results on two sample scenes from this dataset are reported in Figure~\ref{fig:voc_qualitative_varying} for the disjoint 15-1 experimental protocol, where an initial training stage over 15 classes is followed by 5 incremental learning steps each carrying one class to be learned. In the first row our method shows quite robust results across the different learning steps, being able to preserve content semantics. MiB, instead, is able to avoid catastrophic forgetting for one incremental step but it degenerates after introducing the \textit{sheep} class (step 2), which is predicted in spite of \textit{person} probably due to the confusion of the arms and legs (caused also by their similar color). 
Latent representations got even more damaged across subsequent steps, while our approach (SDR) can reduce the interference on latent representations of old classes. Similar considerations also holds for the second set of images, although forgetting is less evident in this scenario: SDR achieves superior performance thanks to correct spatial localization and latent disentanglement.

\textbf{ADE20K.} For this dataset we consider two distinct scenarios: \ie, 5 incremental steps each adding 10 categories to the model (100-10) in Figure~\ref{fig:ade_100_10_qualitative_varying}, and 2 incremental steps each adding 50 classes to the model (50-50) in Figure~\ref{fig:ade_50_50_qualitative_varying}. 

 The first scenario is definitely the most challenging one as the model need to adapt 5 times to discover new (and possibly unrelated) classes. Nevertheless, we can appreciate that our model obtain quite robust results across the various steps in the 2 sample scenes shown in Figure~\ref{fig:ade_100_10_qualitative_varying}, while MiB suffers more from catastrophic forgetting previous knowledge. 
In the first sample scene our approach shows a small gradual degradation across the multiple steps, while MiB firstly completely looses the wall on the background in step 2, then the curtain in step 3 and finally also the hand basin in step 4. Similarly, in the second scene our approach maintains very good results across all the steps, while MiB at the third step misleads the sky on the background. 

In Figure~\ref{fig:ade_50_50_qualitative_varying} we consider the case in which only two incremental steps with 50 classes each are performed. It can be appreciated how in the first step the predicted segmentation maps are quite precise according to both our approach and MiB, but, in both examples, MiB produces a less precise map after the second incremental step. 
More in detail, we remark some differences: our model can identify the tree (green) in the first image, that MiB only partially captures in the first step and completely misses it in the second. Similarly, SDR preserves the walls (gray) in the second image that MiB misleads in the second step.
Again, the latent space regularization helps in preserving previous classes representations and in accommodating new classes.

\section{Quantitative Results: per-Class Accuracy}
\label{sec:quantitative}

We also analyze per-class accuracy for all compared methods in some scenarios. We report the results of per-class IoU and per-class pixel accuracy (PA) on the disjoint 19-1 (Tables~\ref{tab:voc_19_1} and \ref{tab:voc_19_1_PA}), 15-5 (Tables~\ref{tab:voc_15_5} and \ref{tab:voc_15_5_PA}) and 15-1 (Tables~\ref{tab:voc_15_1} and \ref{tab:voc_15_1_PA}) scenarios on the VOC2012 dataset.

Even when adding as little as 1 class (scenario 19-1 in Tables \ref{tab:voc_19_1} and \ref{tab:voc_19_1_PA}) we appreciate how FT and LwF-MC are generally able to learn the new class to some extent but they catastrophically forget previous classes resulting in a poor final mIoU. This performance drop is typically due to a biased prediction toward the new class (high per-class PA for that class but low IoU).
The other competing approaches and our proposal, instead, are more balanced across the various classes and can greatly alleviate forgetting (with performance gains distributed across the classes) when learning the new class, thus resulting in higher mIoUs.

Analyzing the per-class IoUs on the 15-5 case in Tables~\ref{tab:voc_15_5} and \ref{tab:voc_15_5_PA} we can appreciate how FT is completely unable to preserve knowledge about previous classes which are heavily forgotten. The competitors can better preserve knowledge related to previous classes while learning new classes but our approach shows superior results in both retaining old classes knowledge and in learning new ones.

The last 15-1 scenario is shown in Tables~\ref{tab:voc_15_1} and \ref{tab:voc_15_1_PA}. Here we can confirm most of the previous considerations; our method outperforms all the competitors proving its scalability when multiple incremental steps are made. From the per-class pixel accuracy we can observe that most of competing approaches are biased toward the prediction of the very few last classes added to the model, thus reducing the IoU for the other classes.

\newcommand{\imgsizee}{18.7mm}
\begin{figure*}[htbp]

\subfigure{
\centering
\hspace{0.145cm}
\resizebox{17.22cm}{!}{%
\begin{tabular}{ccccccccccc}
\cellcolor[HTML]{000000}{\color[HTML]{FFFFFF} \textbf{background}} & 
\cellcolor[HTML]{800000}{\color[HTML]{FFFFFF}\textbf{aeroplane}} & 
\cellcolor[HTML]{008000}{\color[HTML]{FFFFFF} \textbf{bicycle}} & 
\cellcolor[HTML]{808000}{\color[HTML]{000000} \textbf{bird}} 
& \cellcolor[HTML]{000080}{\color[HTML]{FFFFFF}\textbf{boat}} 
& \cellcolor[HTML]{800080}{\color[HTML]{FFFFFF}\textbf{bottle}}
& \cellcolor[HTML]{008080}\textbf{bus} 
& \cellcolor[HTML]{808080}\textbf{car} 
&\cellcolor[HTML]{400000} {\color[HTML]{FFFFFF}\textbf{cat}} 
& \cellcolor[HTML]{C00000}\textbf{chair} 
& \cellcolor[HTML]{408000}{\color[HTML]{FFFFFF}\textbf{cow}} \\ \cline{11-11}

\cellcolor[HTML]{C08000}\textbf{din. table} 
& \cellcolor[HTML]{400080}{\color[HTML]{FFFFFF} \textbf{dog}} 
& \cellcolor[HTML]{C00080}{\color[HTML]{FFFFFF} \textbf{horse}} & 
\cellcolor[HTML]{408080}{\color[HTML]{FFFFFF} \textbf{mbike}} & 
\cellcolor[HTML]{C08080}{\color[HTML]{000000} \textbf{person}}
 & \cellcolor[HTML]{004000}{\color[HTML]{FFFFFF} \textbf{pplant}} 
 & \cellcolor[HTML]{804000}{\color[HTML]{FFFFFF} \textbf{sheep}} 
 & \cellcolor[HTML]{00C000}{\color[HTML]{000000} \textbf{sofa}} 
  & \cellcolor[HTML]{80C000} \textbf{train}
 & \cellcolor[HTML]{004080}{\color[HTML]{FFFFFF} \textbf{tv}}
 & \multicolumn{1}{|c|}{\textbf{unlabeled}} \\ \cline{11-11} 
\end{tabular}%
}
\vspace{0.1cm}
}

\setlength{\tabcolsep}{0.7pt} 
\renewcommand{\arraystretch}{0.6}
\centering
\begin{tabular}{ccccccccccc}
  
   \rotatebox{90}{\hspace{+4ex}19-1} &  
   \includegraphics[width=\imgsizee]{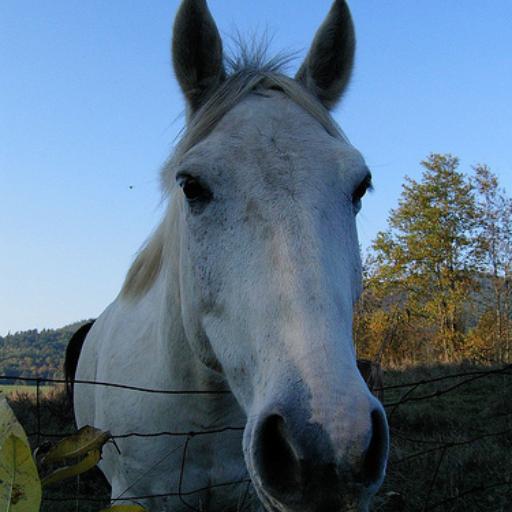} &
   \includegraphics[width=\imgsizee]{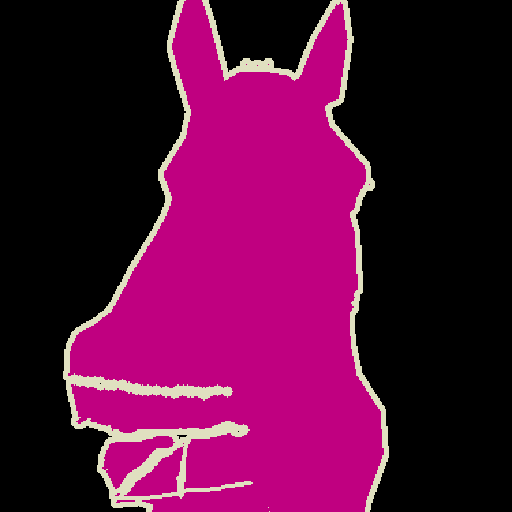} & 
   \includegraphics[width=\imgsizee]{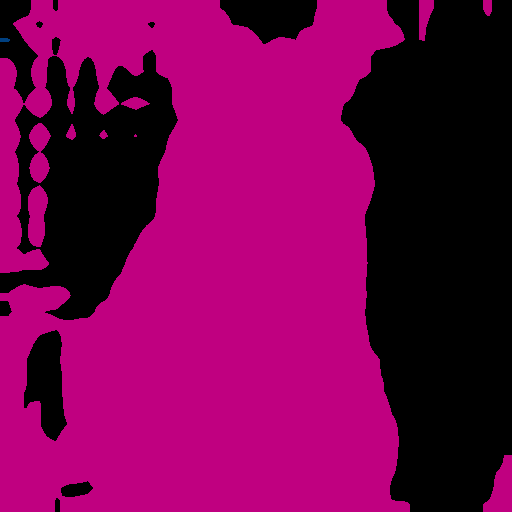} & 
   \includegraphics[width=\imgsizee]{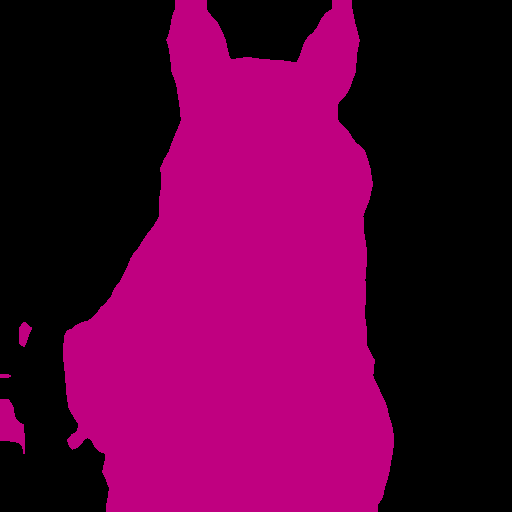} & 
   \includegraphics[width=\imgsizee]{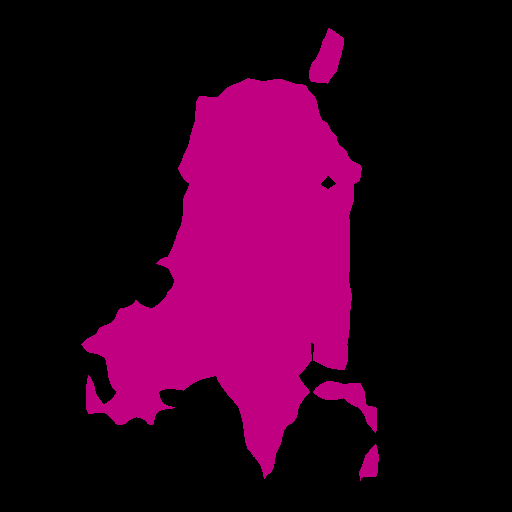} &
   \includegraphics[width=\imgsizee]{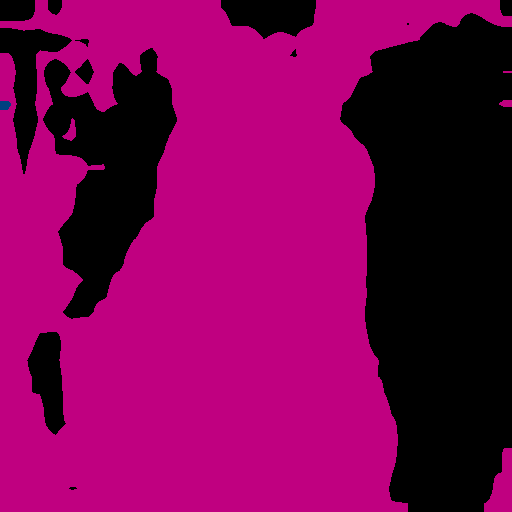} &
   \includegraphics[width=\imgsizee]{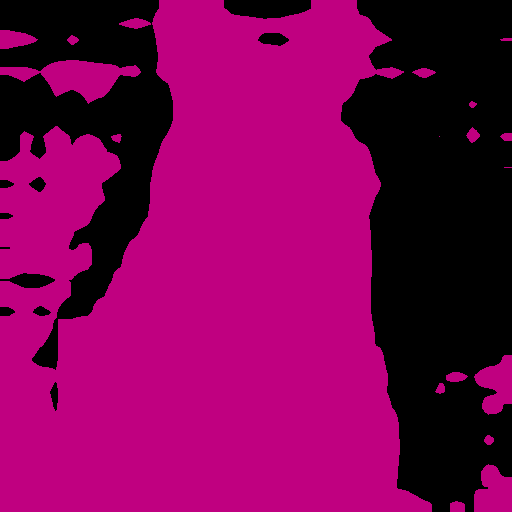} &
   \includegraphics[width=\imgsizee]{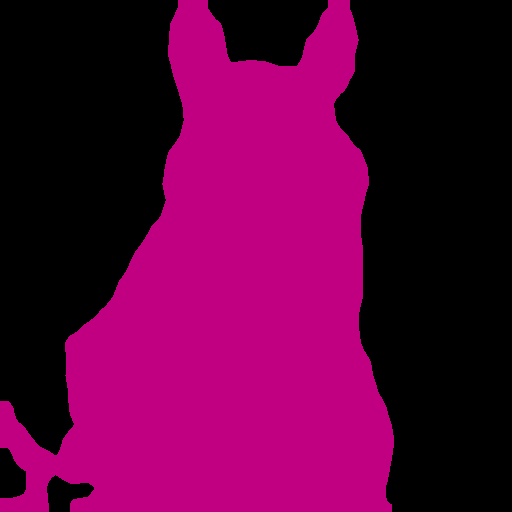} &
   \includegraphics[width=\imgsizee]{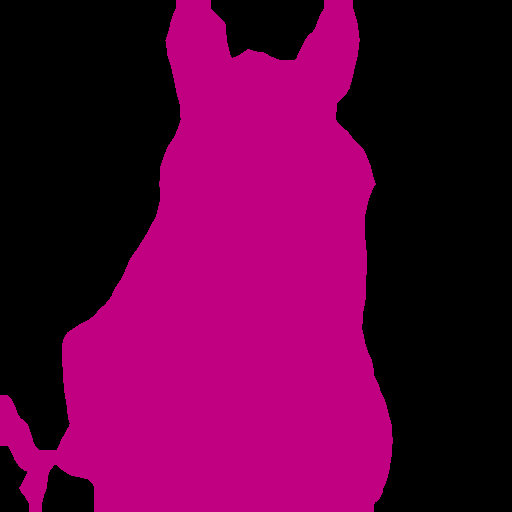} \\ 
   
      \rotatebox{90}{\hspace{+4ex}19-1} &  
   \includegraphics[width=\imgsizee]{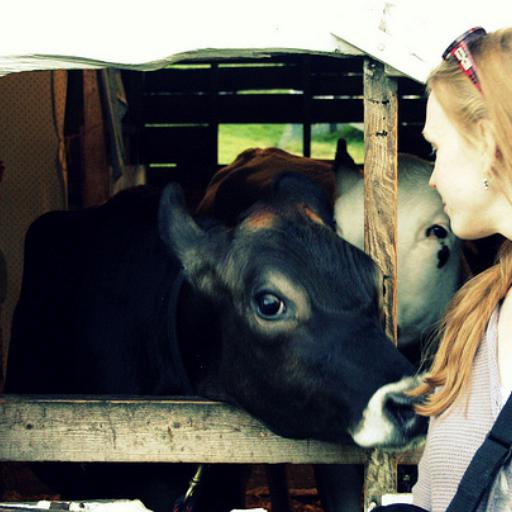} &
   \includegraphics[width=\imgsizee]{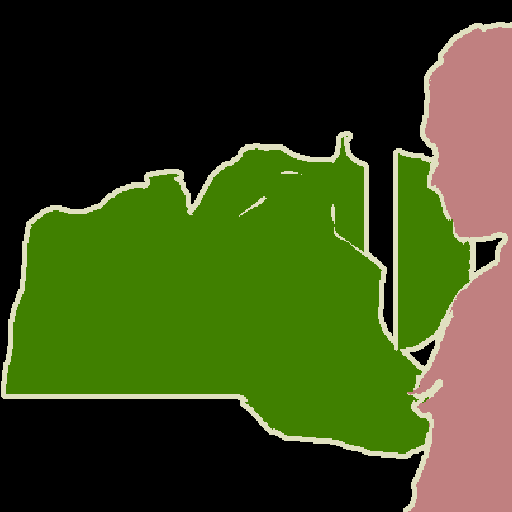} & 
   \includegraphics[width=\imgsizee]{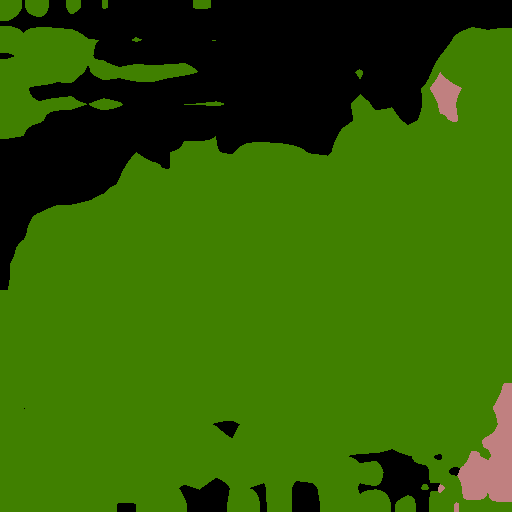} & 
   \includegraphics[width=\imgsizee]{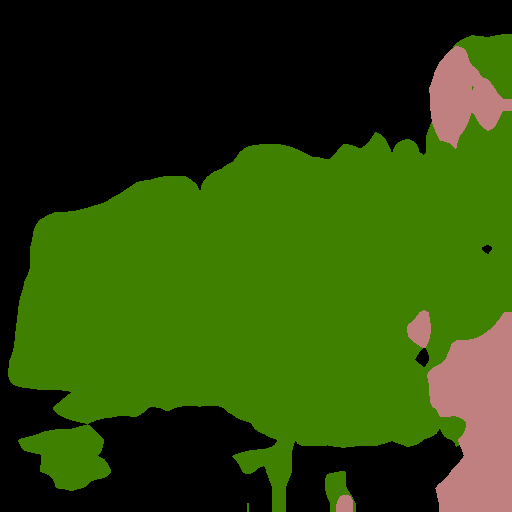} & 
   \includegraphics[width=\imgsizee]{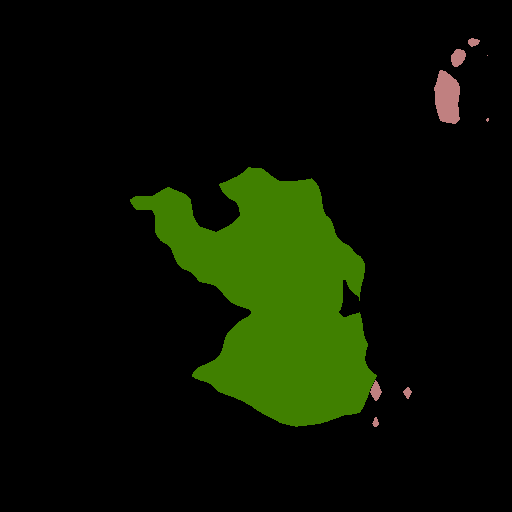} &
   \includegraphics[width=\imgsizee]{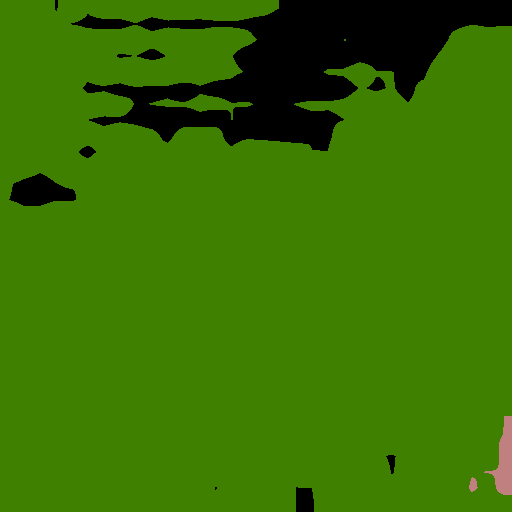} &
   \includegraphics[width=\imgsizee]{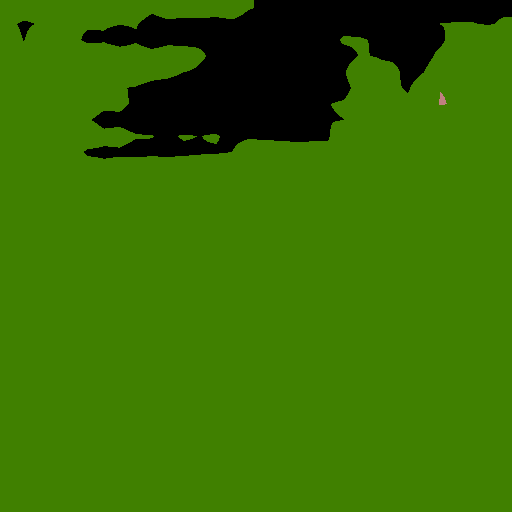} &
   \includegraphics[width=\imgsizee]{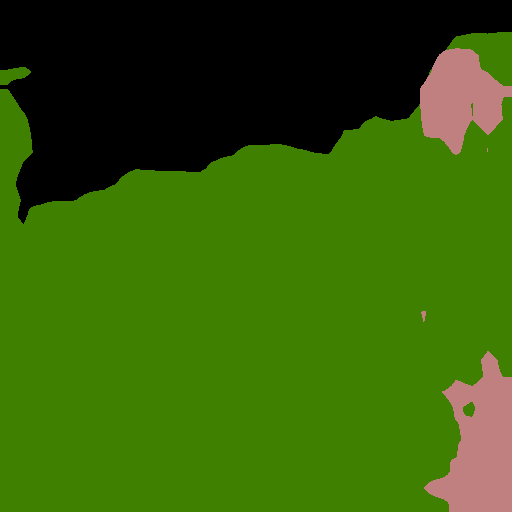} &
   \includegraphics[width=\imgsizee]{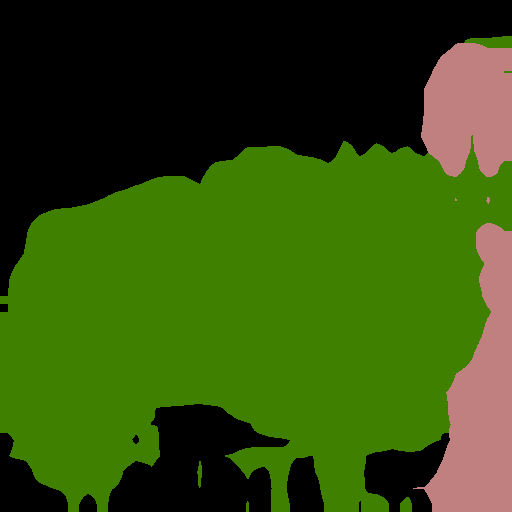} \\
\end{tabular}
\caption{Qualitative results on sample scenes in different scenarios (19-1, 15-5 and 15-1) on Pascal VOC 2012 of the proposed method and of competing approaches in the sequential setup (\textit{best viewed in colors}).}
\label{fig:voc_qualitative_sequential}
\end{figure*}

\begin{figure*}[htbp]

\subfigure{
\centering
\hspace{0.145cm}
\resizebox{17.22cm}{!}{%
\begin{tabular}{ccccccccccc}
\cellcolor[HTML]{000000}{\color[HTML]{FFFFFF} \textbf{background}} & 
\cellcolor[HTML]{800000}{\color[HTML]{FFFFFF}\textbf{aeroplane}} & 
\cellcolor[HTML]{008000}{\color[HTML]{FFFFFF} \textbf{bicycle}} & 
\cellcolor[HTML]{808000}{\color[HTML]{000000} \textbf{bird}} 
& \cellcolor[HTML]{000080}{\color[HTML]{FFFFFF}\textbf{boat}} 
& \cellcolor[HTML]{800080}{\color[HTML]{FFFFFF}\textbf{bottle}}
& \cellcolor[HTML]{008080}\textbf{bus} 
& \cellcolor[HTML]{808080}\textbf{car} 
&\cellcolor[HTML]{400000} {\color[HTML]{FFFFFF}\textbf{cat}} 
& \cellcolor[HTML]{C00000}\textbf{chair} 
& \cellcolor[HTML]{408000}{\color[HTML]{FFFFFF}\textbf{cow}} \\ \cline{11-11}

\cellcolor[HTML]{C08000}\textbf{din. table} 
& \cellcolor[HTML]{400080}{\color[HTML]{FFFFFF} \textbf{dog}} 
& \cellcolor[HTML]{C00080}{\color[HTML]{FFFFFF} \textbf{horse}} & 
\cellcolor[HTML]{408080}{\color[HTML]{FFFFFF} \textbf{mbike}} & 
\cellcolor[HTML]{C08080}{\color[HTML]{000000} \textbf{person}}
 & \cellcolor[HTML]{004000}{\color[HTML]{FFFFFF} \textbf{pplant}} 
 & \cellcolor[HTML]{804000}{\color[HTML]{FFFFFF} \textbf{sheep}} 
 & \cellcolor[HTML]{00C000}{\color[HTML]{000000} \textbf{sofa}} 
  & \cellcolor[HTML]{80C000} \textbf{train}
 & \cellcolor[HTML]{004080}{\color[HTML]{FFFFFF} \textbf{tv}}
 & \multicolumn{1}{|c|}{\textbf{unlabeled}} \\ \cline{11-11} 
\end{tabular}%
}
\vspace{0.1cm}
}

\setlength{\tabcolsep}{0.7pt} 
\renewcommand{\arraystretch}{0.6}
\centering
\begin{tabular}{ccccccccccc}
  
   \rotatebox{90}{\hspace{+4ex}19-1} &  
   \includegraphics[width=\imgsizee]{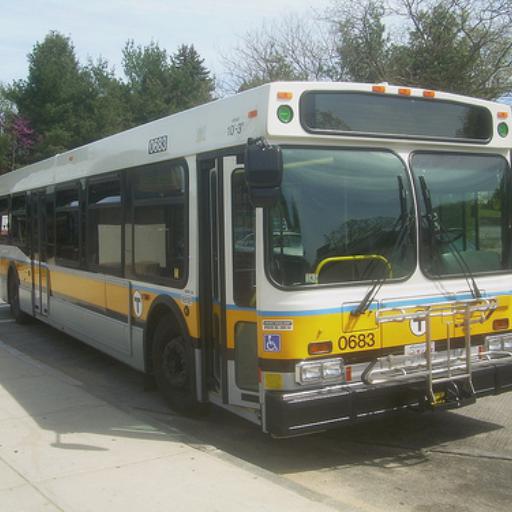} &
   \includegraphics[width=\imgsizee]{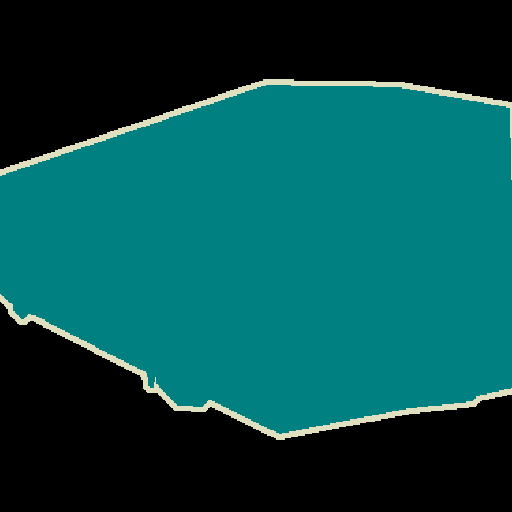} & 
   \includegraphics[width=\imgsizee]{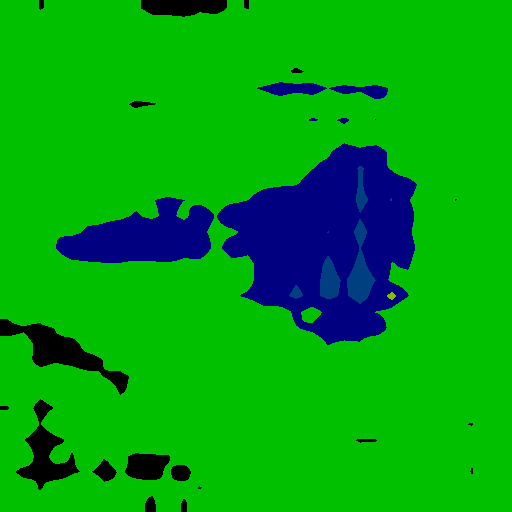} & 
   \includegraphics[width=\imgsizee]{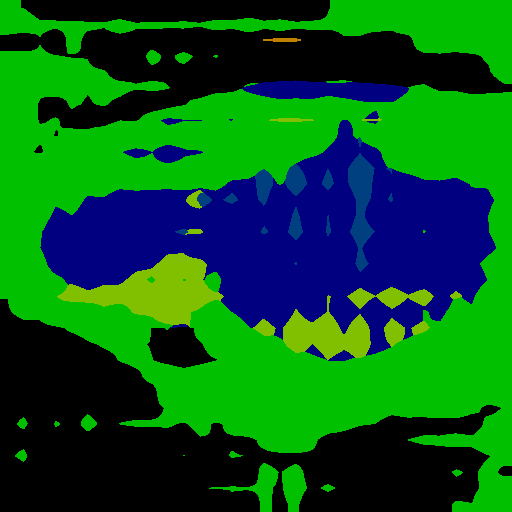} & 
   \includegraphics[width=\imgsizee]{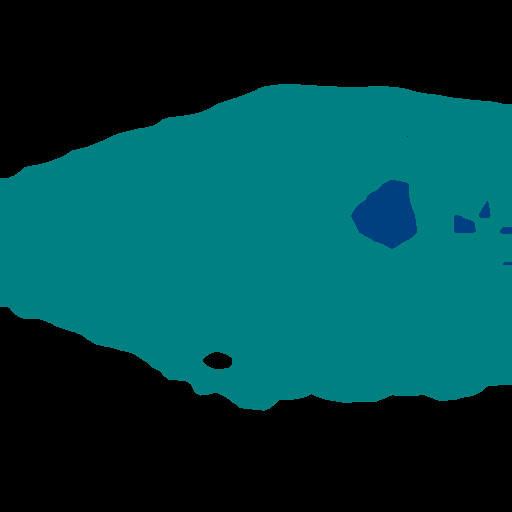} &
   \includegraphics[width=\imgsizee]{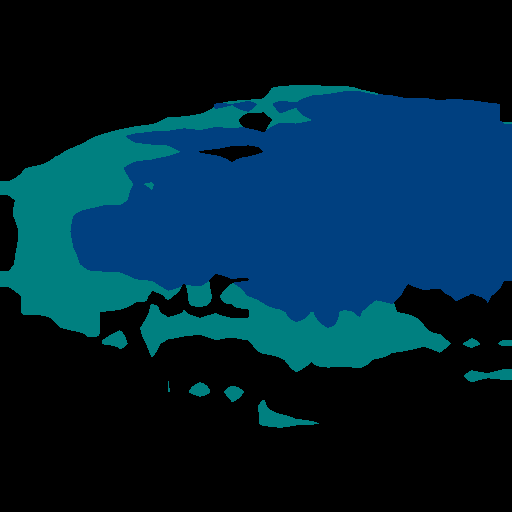} &
   \includegraphics[width=\imgsizee]{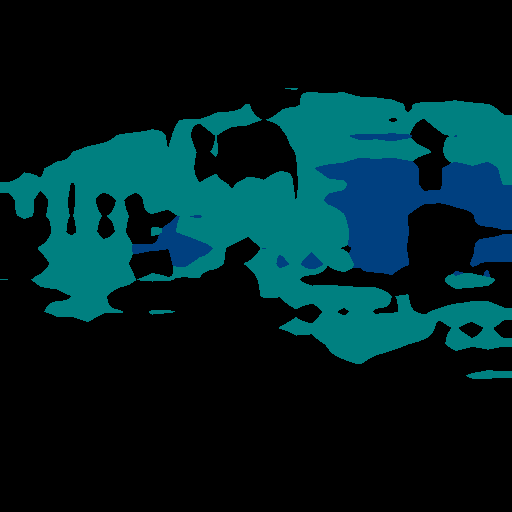} &
   \includegraphics[width=\imgsizee]{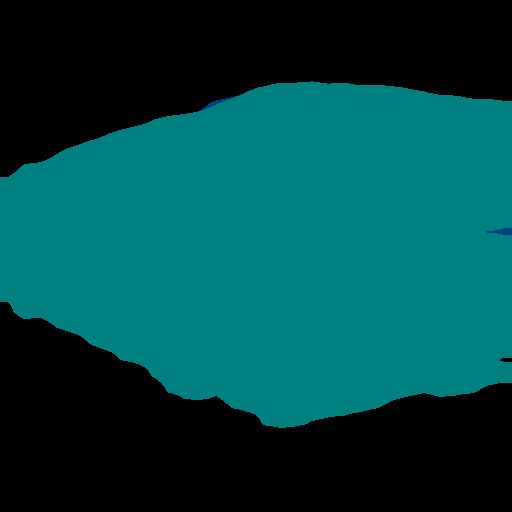} &
   \includegraphics[width=\imgsizee]{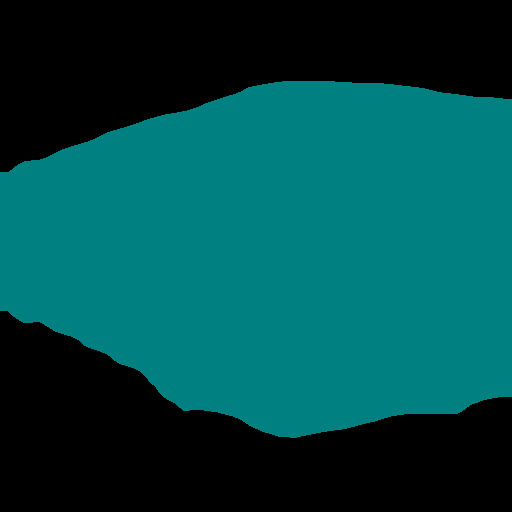} \\ 
   
      \rotatebox{90}{\hspace{+4ex}19-1} &  
   \includegraphics[width=\imgsizee]{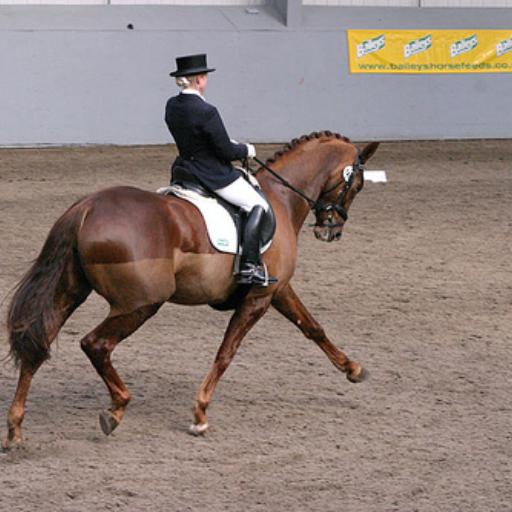} &
   \includegraphics[width=\imgsizee]{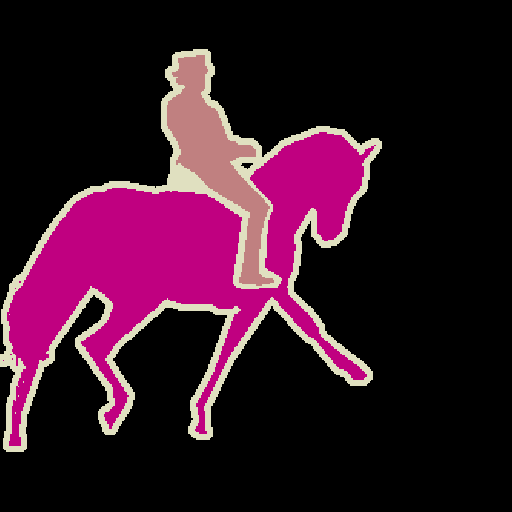} & 
   \includegraphics[width=\imgsizee]{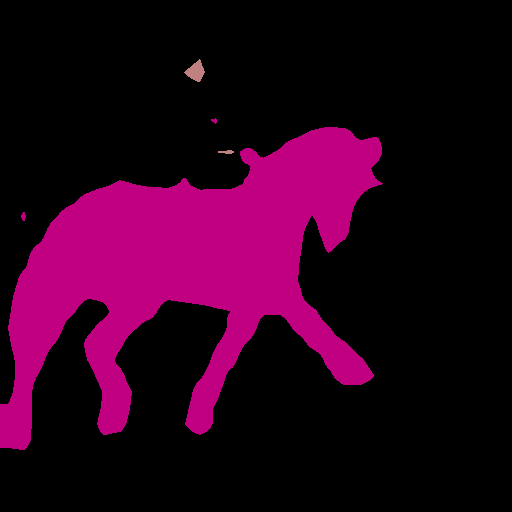} & 
   \includegraphics[width=\imgsizee]{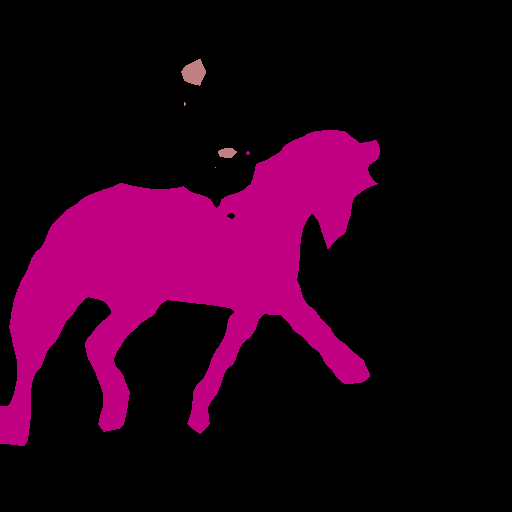} & 
   \includegraphics[width=\imgsizee]{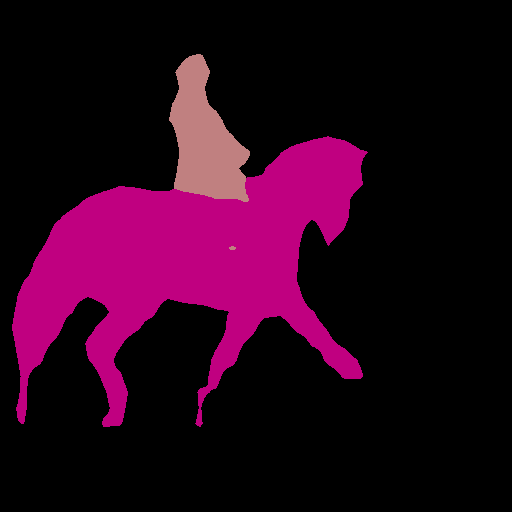} &
   \includegraphics[width=\imgsizee]{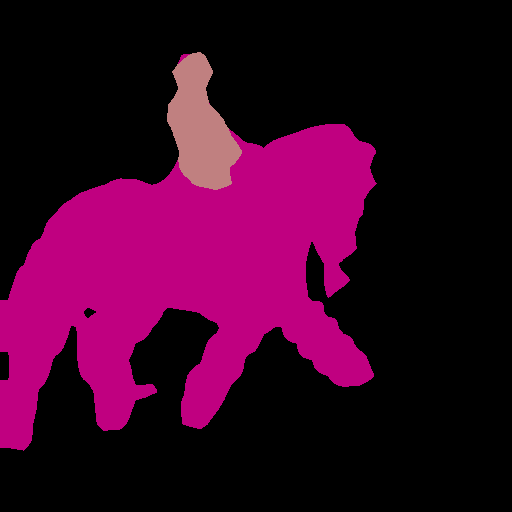} &
   \includegraphics[width=\imgsizee]{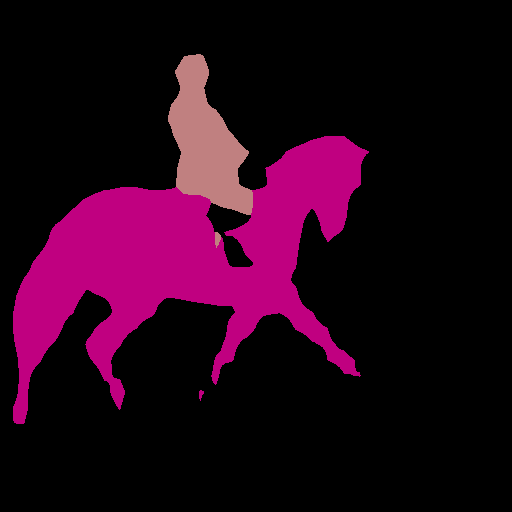} &
   \includegraphics[width=\imgsizee]{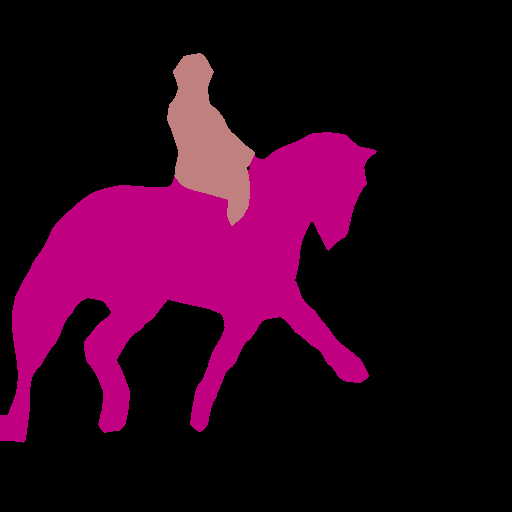} &
   \includegraphics[width=\imgsizee]{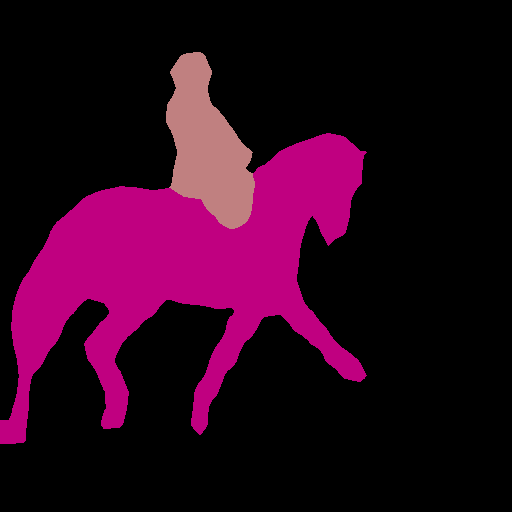} \\
\end{tabular}
\caption{Qualitative results on sample scenes in different scenarios (19-1, 15-5 and 15-1) on Pascal VOC 2012 of the proposed method and of competing approaches in the disjoint setup (\textit{best viewed in colors}).}
\label{fig:voc_qualitative_disjoint}
\end{figure*}

\begin{figure*}[htbp]

\subfigure{
\centering
\hspace{0.145cm}
\resizebox{17.22cm}{!}{%
\begin{tabular}{ccccccccccc}
\cellcolor[HTML]{000000}{\color[HTML]{FFFFFF} \textbf{background}} & 
\cellcolor[HTML]{800000}{\color[HTML]{FFFFFF}\textbf{aeroplane}} & 
\cellcolor[HTML]{008000}{\color[HTML]{FFFFFF} \textbf{bicycle}} & 
\cellcolor[HTML]{808000}{\color[HTML]{000000} \textbf{bird}} 
& \cellcolor[HTML]{000080}{\color[HTML]{FFFFFF}\textbf{boat}} 
& \cellcolor[HTML]{800080}{\color[HTML]{FFFFFF}\textbf{bottle}}
& \cellcolor[HTML]{008080}\textbf{bus} 
& \cellcolor[HTML]{808080}\textbf{car} 
&\cellcolor[HTML]{400000} {\color[HTML]{FFFFFF}\textbf{cat}} 
& \cellcolor[HTML]{C00000}\textbf{chair} 
& \cellcolor[HTML]{408000}{\color[HTML]{FFFFFF}\textbf{cow}} \\ \cline{11-11}

\cellcolor[HTML]{C08000}\textbf{din. table} 
& \cellcolor[HTML]{400080}{\color[HTML]{FFFFFF} \textbf{dog}} 
& \cellcolor[HTML]{C00080}{\color[HTML]{FFFFFF} \textbf{horse}} & 
\cellcolor[HTML]{408080}{\color[HTML]{FFFFFF} \textbf{mbike}} & 
\cellcolor[HTML]{C08080}{\color[HTML]{000000} \textbf{person}}
 & \cellcolor[HTML]{004000}{\color[HTML]{FFFFFF} \textbf{pplant}} 
 & \cellcolor[HTML]{804000}{\color[HTML]{FFFFFF} \textbf{sheep}} 
 & \cellcolor[HTML]{00C000}{\color[HTML]{000000} \textbf{sofa}} 
  & \cellcolor[HTML]{80C000} \textbf{train}
 & \cellcolor[HTML]{004080}{\color[HTML]{FFFFFF} \textbf{tv}}
 & \multicolumn{1}{|c|}{\textbf{unlabeled}} \\ \cline{11-11} 
\end{tabular}%
}
\vspace{0.1cm}
}

\setlength{\tabcolsep}{0.7pt} 
\renewcommand{\arraystretch}{0.6}
\centering
\begin{tabular}{ccccccccccc}
  
   \rotatebox{90}{\hspace{+4ex}19-1} &  
   \includegraphics[width=\imgsizee]{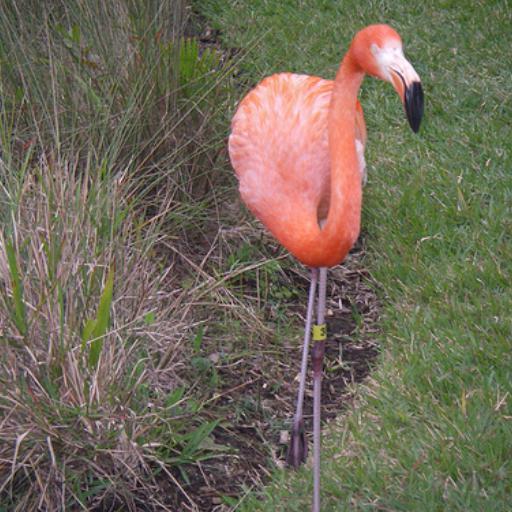} &
   \includegraphics[width=\imgsizee]{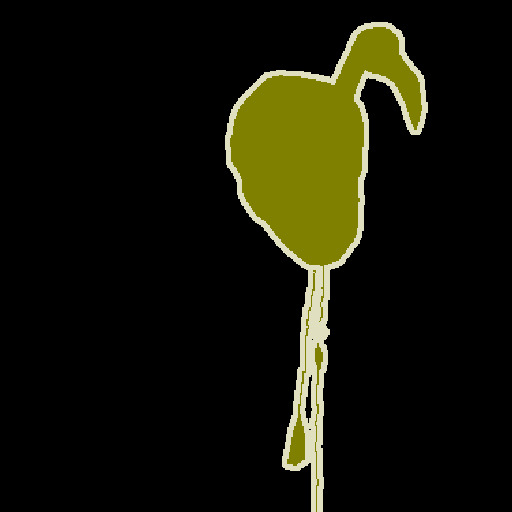} & 
   \includegraphics[width=\imgsizee]{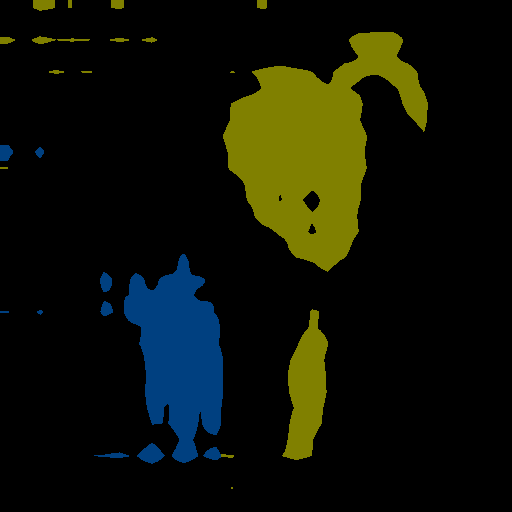} & 
   \includegraphics[width=\imgsizee]{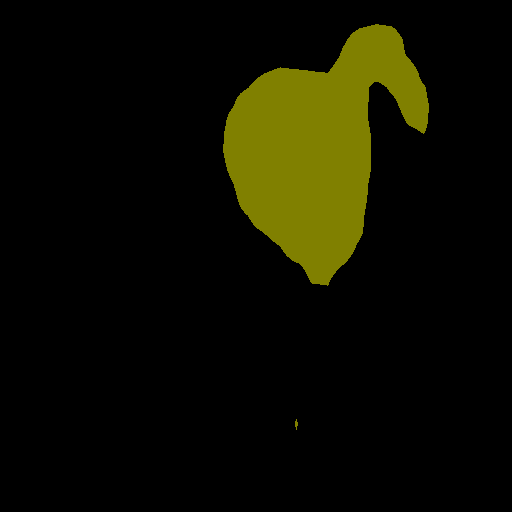} & 
   \includegraphics[width=\imgsizee]{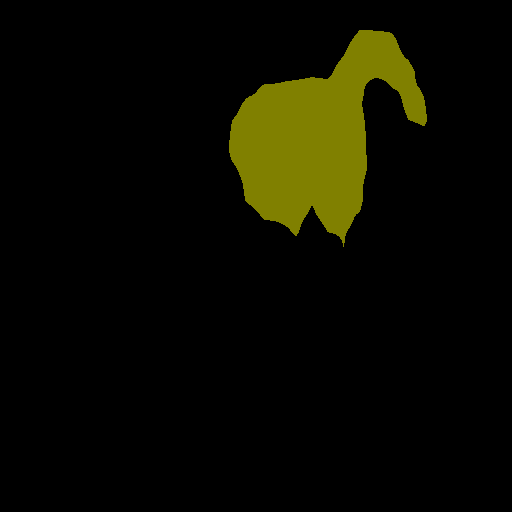} &
   \includegraphics[width=\imgsizee]{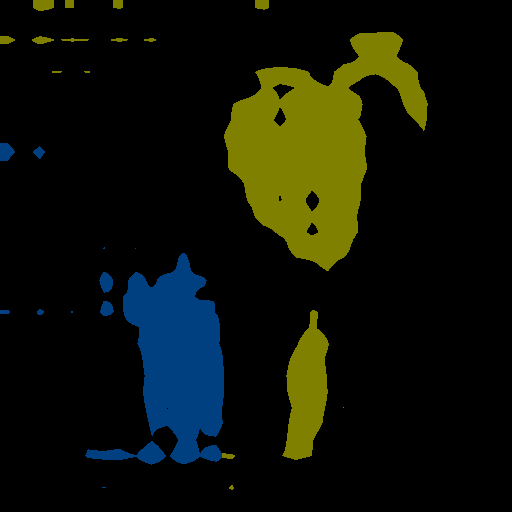} &
   \includegraphics[width=\imgsizee]{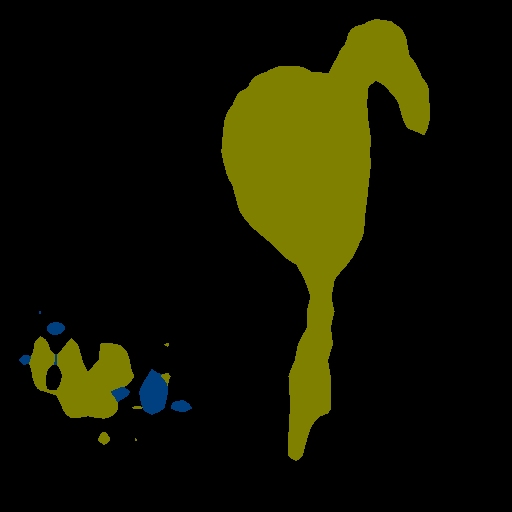} &
   \includegraphics[width=\imgsizee]{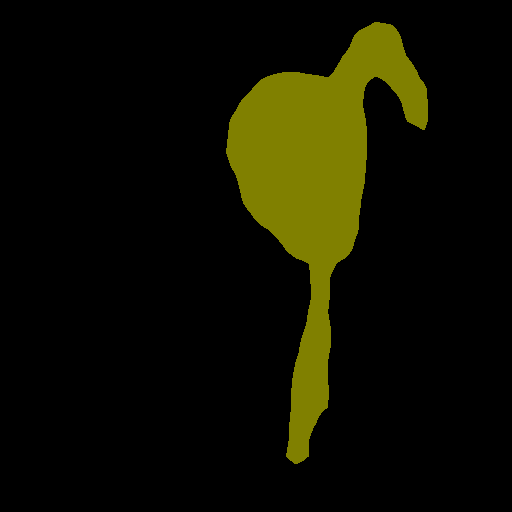} &
   \includegraphics[width=\imgsizee]{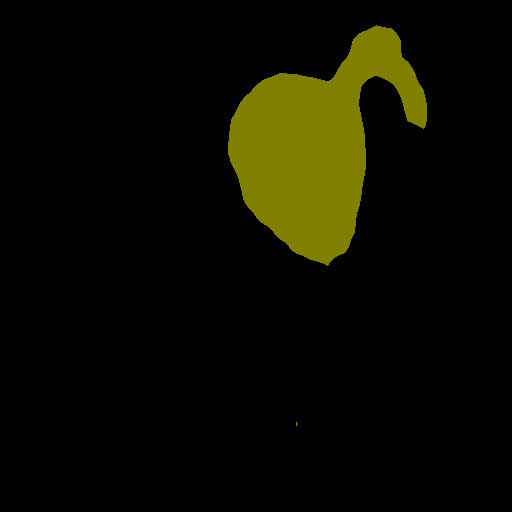} \\ 
   
      \rotatebox{90}{\hspace{+4ex}19-1} &  
   \includegraphics[width=\imgsizee]{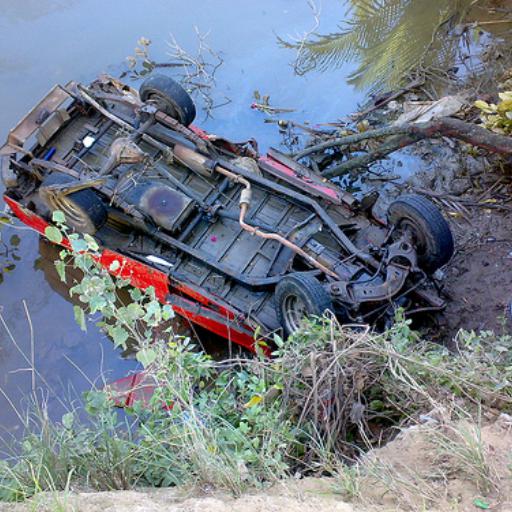} &
   \includegraphics[width=\imgsizee]{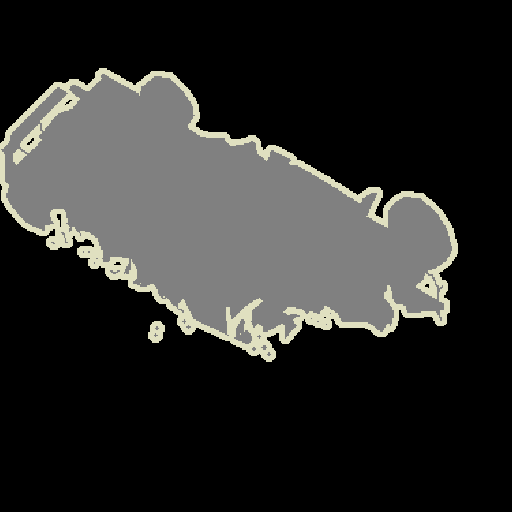} & 
   \includegraphics[width=\imgsizee]{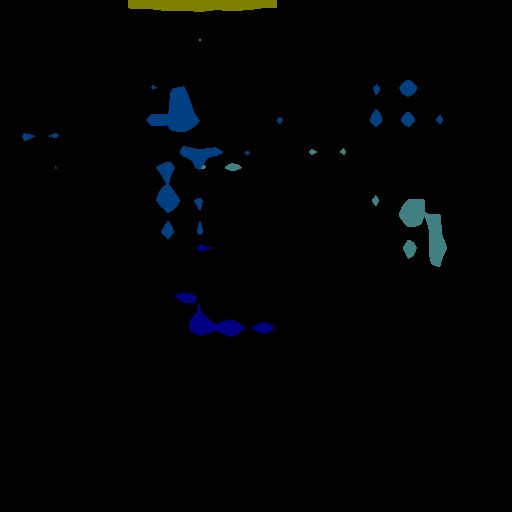} & 
   \includegraphics[width=\imgsizee]{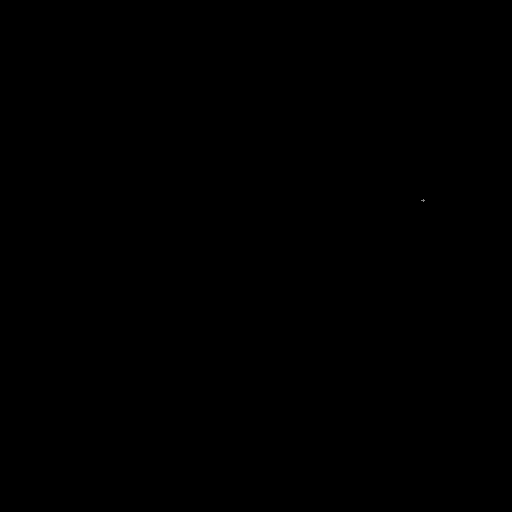} & 
   \includegraphics[width=\imgsizee]{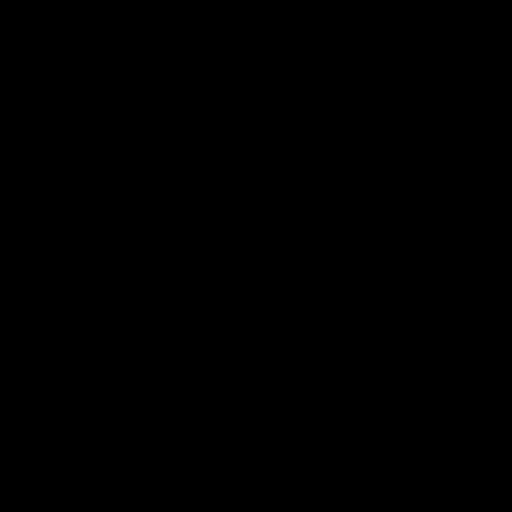} &
   \includegraphics[width=\imgsizee]{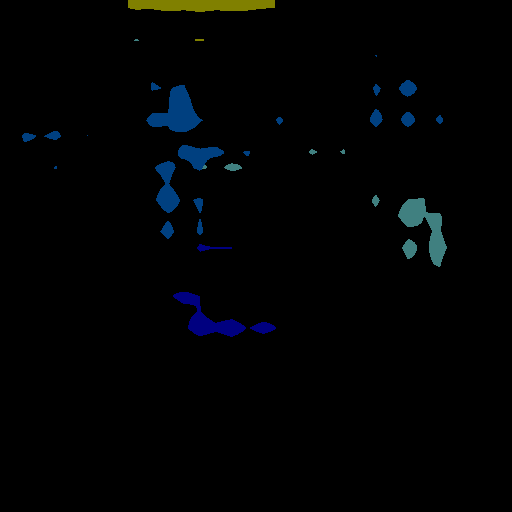} &
   \includegraphics[width=\imgsizee]{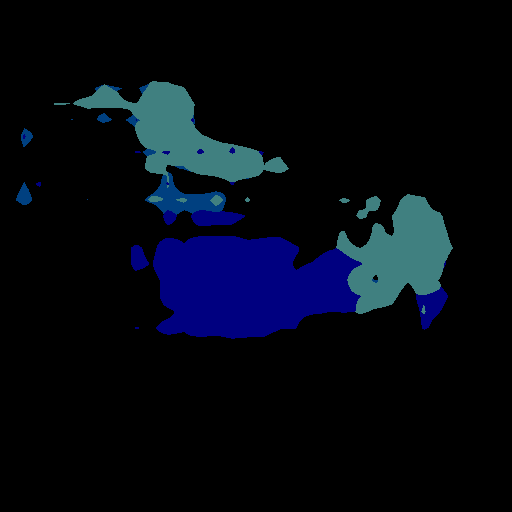} &
   \includegraphics[width=\imgsizee]{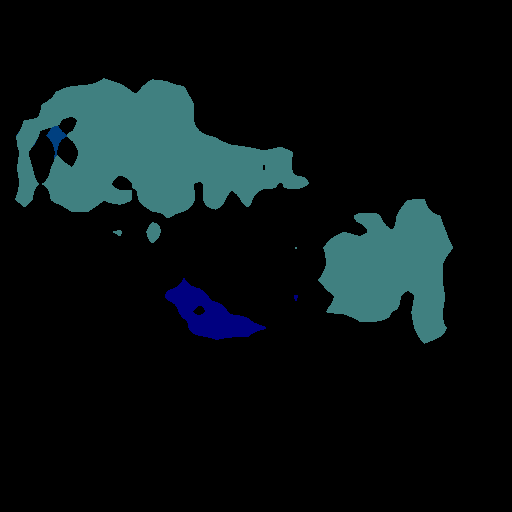} &
   \includegraphics[width=\imgsizee]{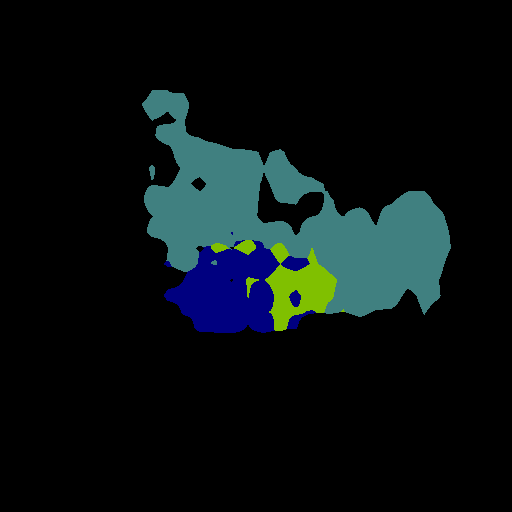} \\
\end{tabular}
\caption{Qualitative results on sample scenes in different scenarios (19-1, 15-5 and 15-1) on Pascal VOC 2012 of the proposed method and of competing approaches in the overlapped setup (\textit{best viewed in colors}).}
\label{fig:voc_qualitative_overlapped}
\end{figure*}

\begin{figure*}[htbp]
\setlength{\tabcolsep}{0.7pt} 
\renewcommand{\arraystretch}{0.6}
\centering
\begin{tabular}{ccccccccccc}
  
   \rotatebox{90}{\hspace{+3ex}100-50} &  
   \includegraphics[width=\imgsizee]{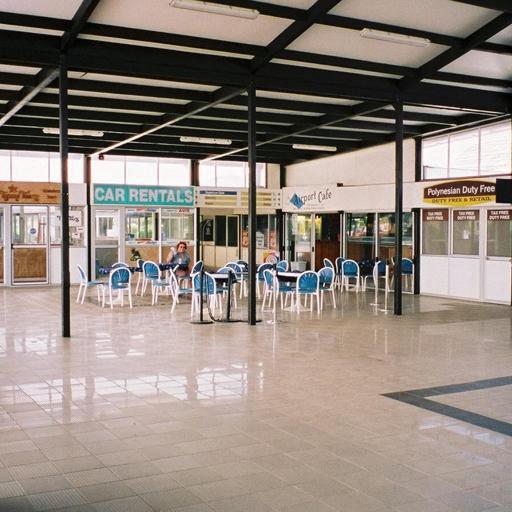} &
   \includegraphics[width=\imgsizee]{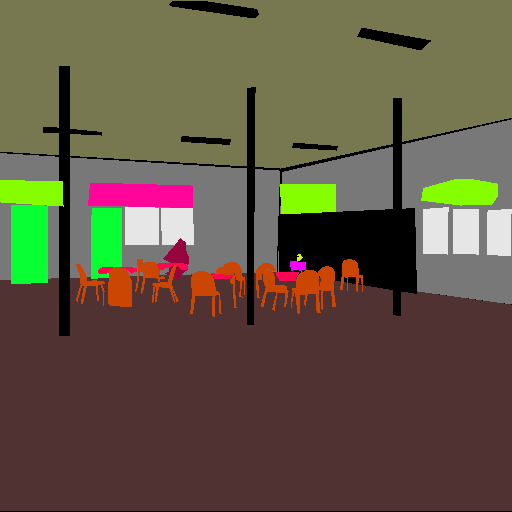} & 
   \includegraphics[width=\imgsizee]{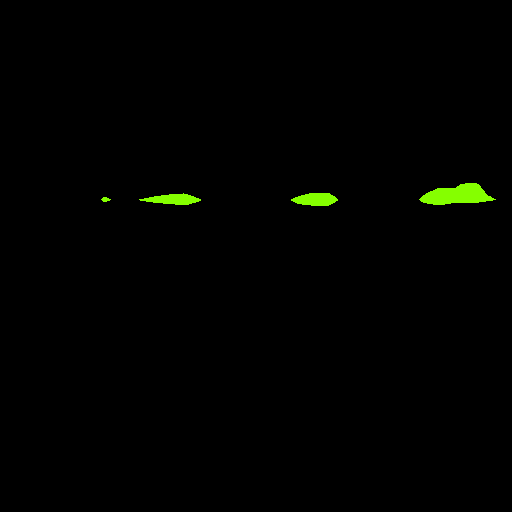} & 
   \includegraphics[width=\imgsizee]{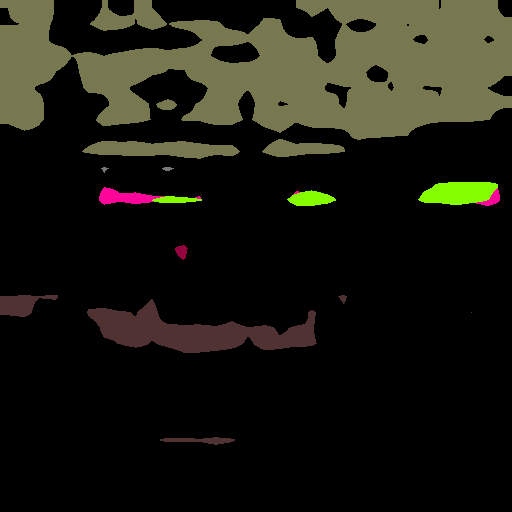} & 
   \includegraphics[width=\imgsizee]{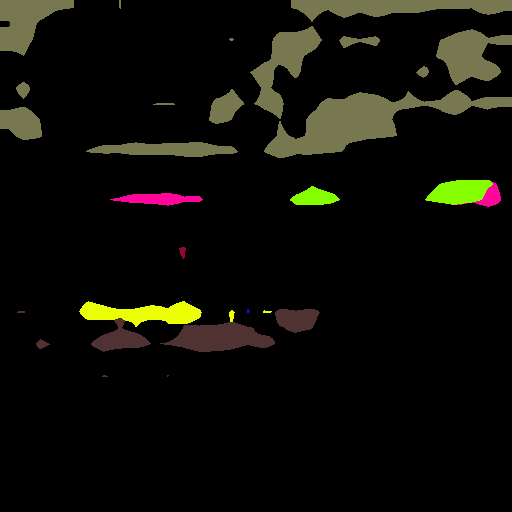} &
   \includegraphics[width=\imgsizee]{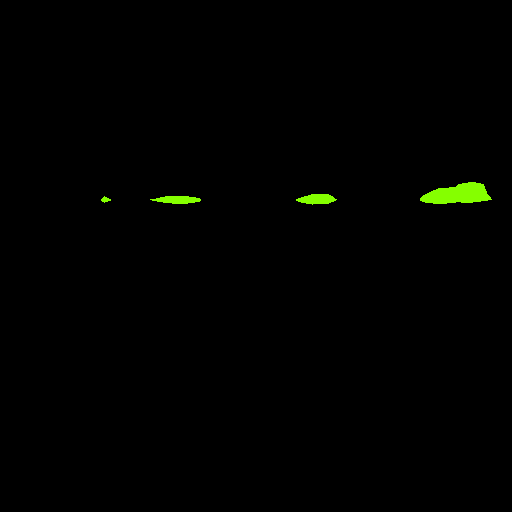} &
   \includegraphics[width=\imgsizee]{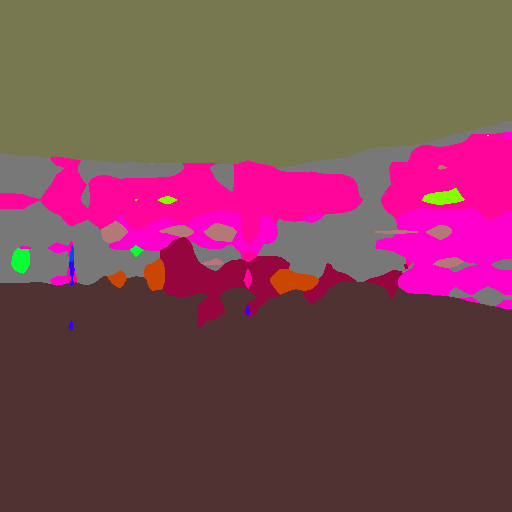} &
   \includegraphics[width=\imgsizee]{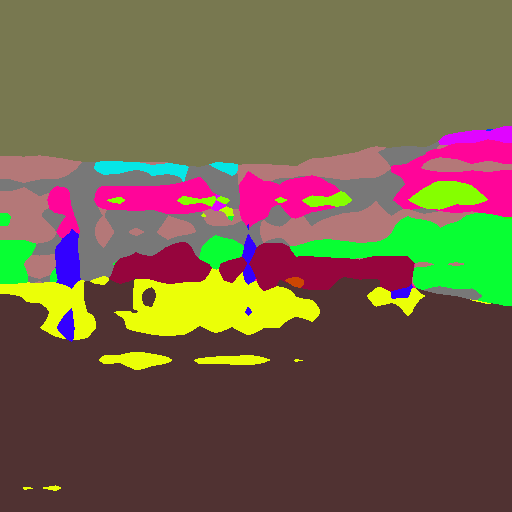} &
   \includegraphics[width=\imgsizee]{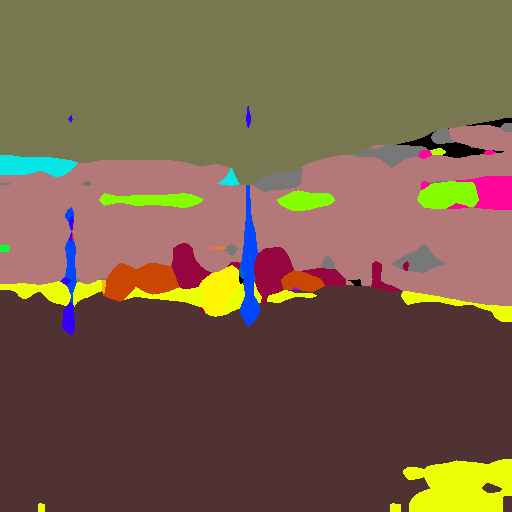} \\ 
   
      \rotatebox{90}{\hspace{+3ex}100-50} &  
   \includegraphics[width=\imgsizee]{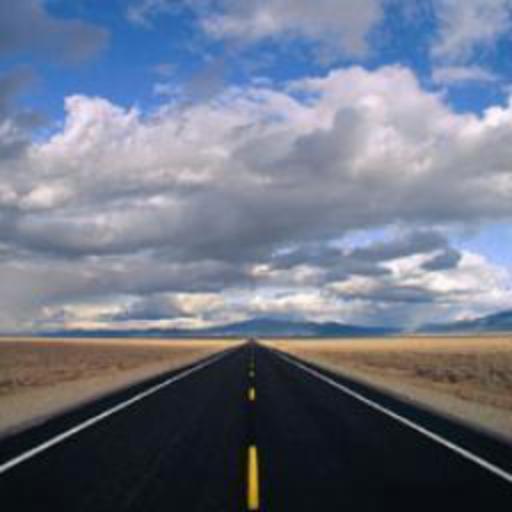} &
   \includegraphics[width=\imgsizee]{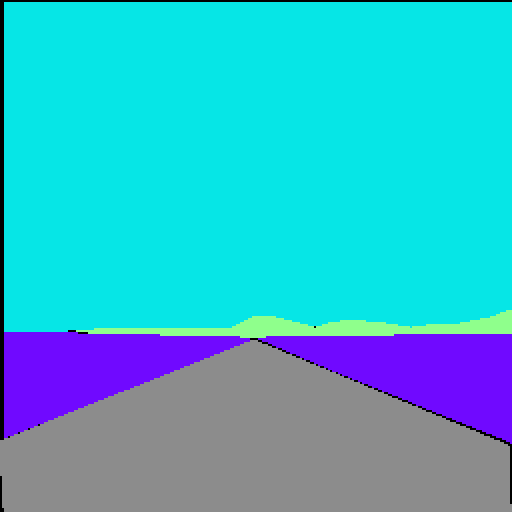} & 
   \includegraphics[width=\imgsizee]{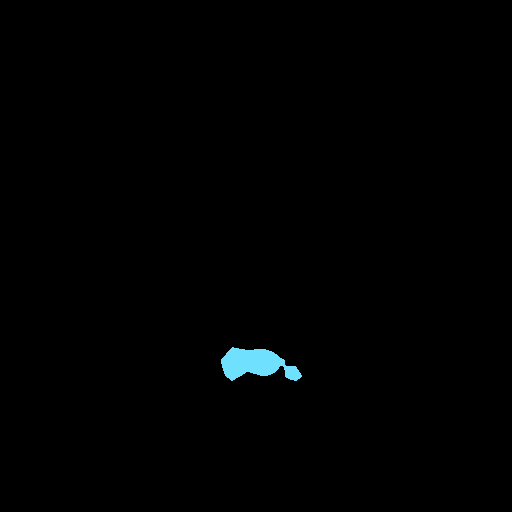} & 
   \includegraphics[width=\imgsizee]{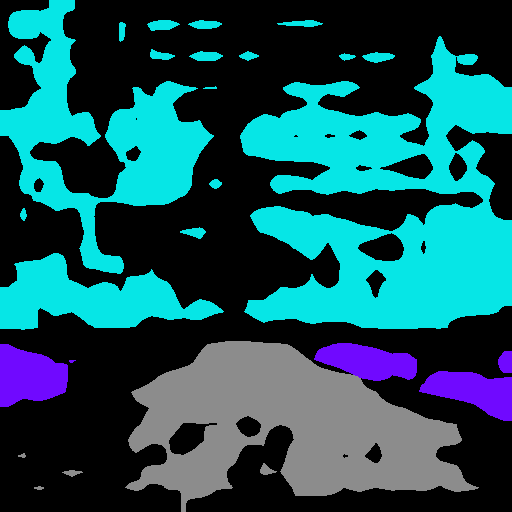} & 
   \includegraphics[width=\imgsizee]{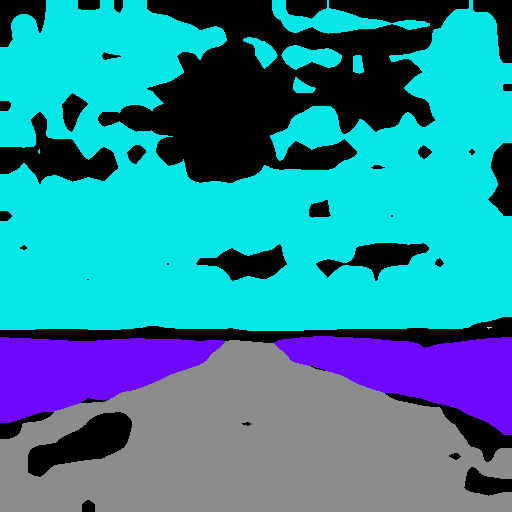} &
   \includegraphics[width=\imgsizee]{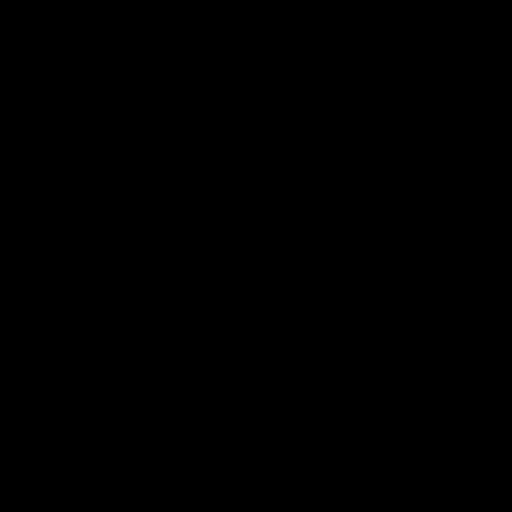} &
   \includegraphics[width=\imgsizee]{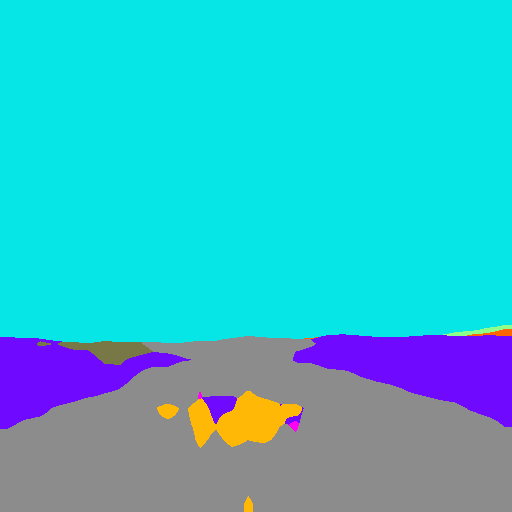} &
   \includegraphics[width=\imgsizee]{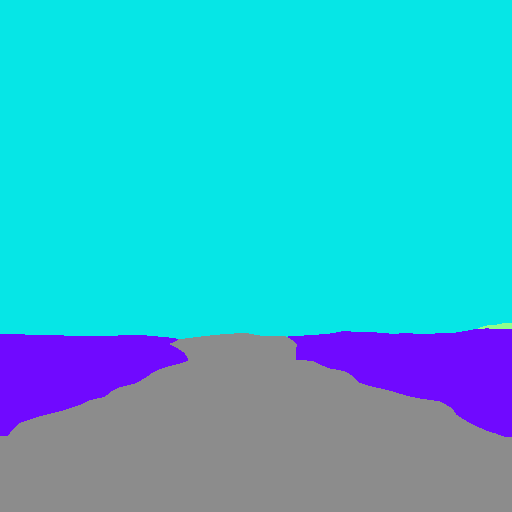} &
   \includegraphics[width=\imgsizee]{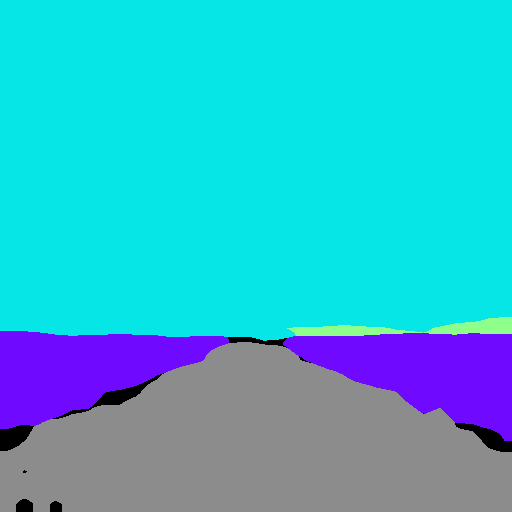} \\
\end{tabular}
\caption{Qualitative results on sample scenes in different scenarios (100-50, 100-10 and 50-50) on ADE20K of the proposed method and of competing approaches (\textit{best viewed in colors}).}
\label{fig:ade_qualitative}
\end{figure*}

\newcommand{\imgsizebis}{23.5mm}
\begin{figure*}[htbp]

\subfigure{
\centering
\hspace{0.145cm}
\resizebox{17.22cm}{!}{%
\begin{tabular}{ccccccccccc}
\cellcolor[HTML]{000000}{\color[HTML]{FFFFFF} \textbf{background}} & 
\cellcolor[HTML]{800000}{\color[HTML]{FFFFFF}\textbf{aeroplane}} & 
\cellcolor[HTML]{008000}{\color[HTML]{FFFFFF} \textbf{bicycle}} & 
\cellcolor[HTML]{808000}{\color[HTML]{000000} \textbf{bird}} 
& \cellcolor[HTML]{000080}{\color[HTML]{FFFFFF}\textbf{boat}} 
& \cellcolor[HTML]{800080}{\color[HTML]{FFFFFF}\textbf{bottle}}
& \cellcolor[HTML]{008080}\textbf{bus} 
& \cellcolor[HTML]{808080}\textbf{car} 
&\cellcolor[HTML]{400000} {\color[HTML]{FFFFFF}\textbf{cat}} 
& \cellcolor[HTML]{C00000}\textbf{chair} 
& \cellcolor[HTML]{408000}{\color[HTML]{FFFFFF}\textbf{cow}} \\ \cline{11-11}

\cellcolor[HTML]{C08000}\textbf{din. table} 
& \cellcolor[HTML]{400080}{\color[HTML]{FFFFFF} \textbf{dog}} 
& \cellcolor[HTML]{C00080}{\color[HTML]{FFFFFF} \textbf{horse}} & 
\cellcolor[HTML]{408080}{\color[HTML]{FFFFFF} \textbf{mbike}} & 
\cellcolor[HTML]{C08080}{\color[HTML]{000000} \textbf{person}}
 & \cellcolor[HTML]{004000}{\color[HTML]{FFFFFF} \textbf{pplant}} 
 & \cellcolor[HTML]{804000}{\color[HTML]{FFFFFF} \textbf{sheep}} 
 & \cellcolor[HTML]{00C000}{\color[HTML]{000000} \textbf{sofa}} 
  & \cellcolor[HTML]{80C000} \textbf{train}
 & \cellcolor[HTML]{004080}{\color[HTML]{FFFFFF} \textbf{tv}}
 & \multicolumn{1}{|c|}{\textbf{unlabeled}} \\ \cline{11-11} 
\end{tabular}%
}
\vspace{0.1cm}
}

\setlength{\tabcolsep}{0.7pt} 
\renewcommand{\arraystretch}{0.6}
\centering
\begin{tabular}{ccccccccccc}
  
   \rotatebox{90}{\hspace{+2ex}SDR (ours)} &  
   \includegraphics[width=\imgsizebis]{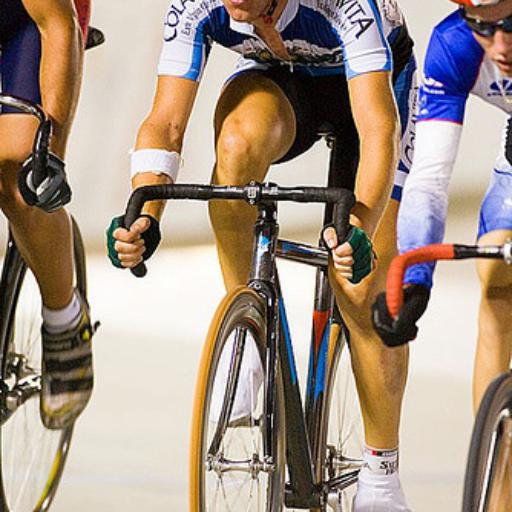} &
   \includegraphics[width=\imgsizebis]{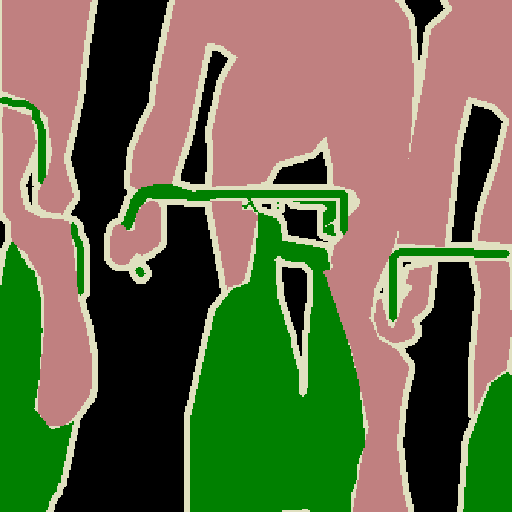} & 
   \includegraphics[width=\imgsizebis]{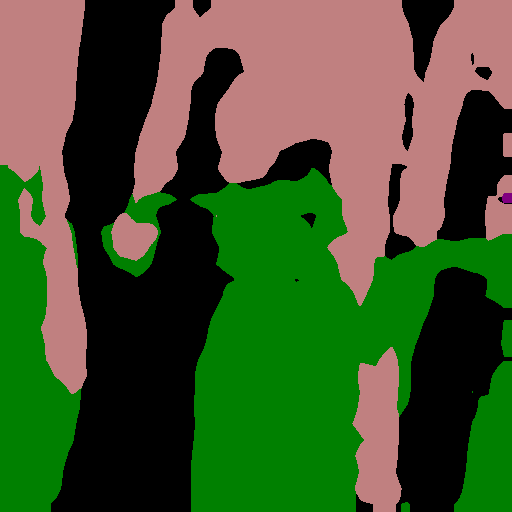} & 
   \includegraphics[width=\imgsizebis]{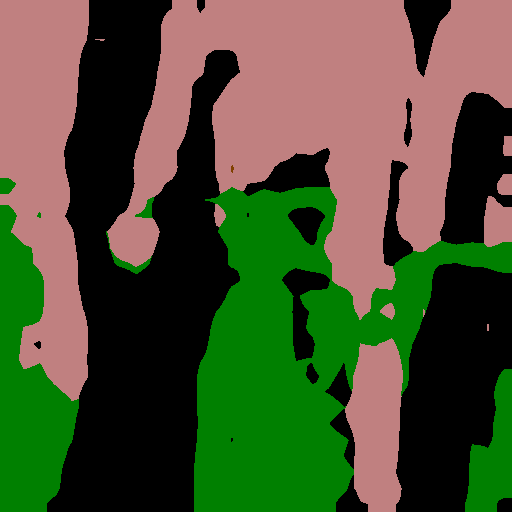} & 
   \includegraphics[width=\imgsizebis]{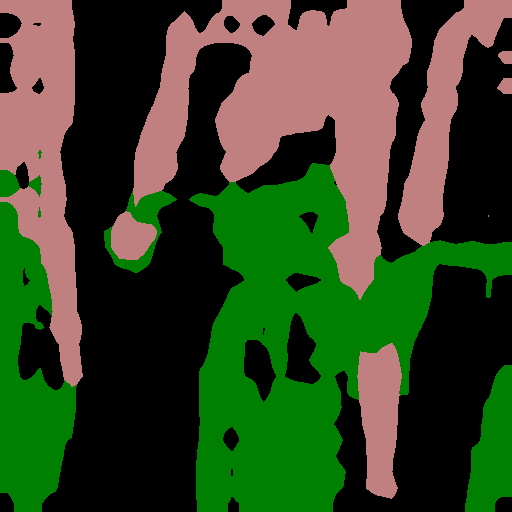} &
   \includegraphics[width=\imgsizebis]{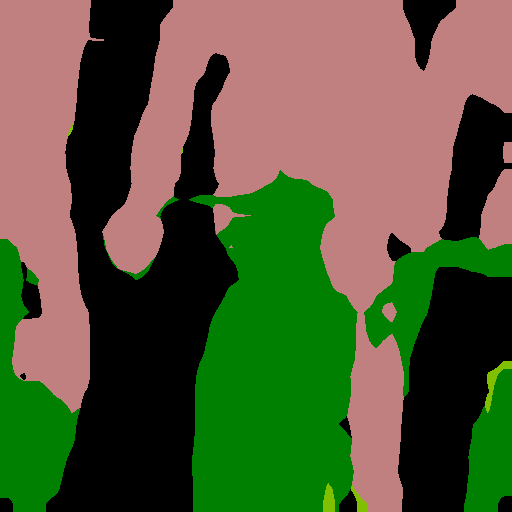} &
   \includegraphics[width=\imgsizebis]{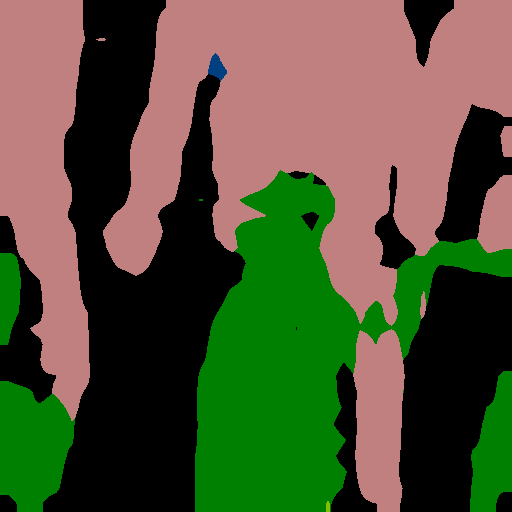} \\ 
   
      \rotatebox{90}{\hspace{+4ex}MiB \cite{cermelli2020modeling}} &  
   \includegraphics[width=\imgsizebis]{img/suppl/increasing_steps/disjoint/2007_001311_RGB.jpg} &
   \includegraphics[width=\imgsizebis]{img/suppl/increasing_steps/disjoint/2007_001311_label.png} & 
   \includegraphics[width=\imgsizebis]{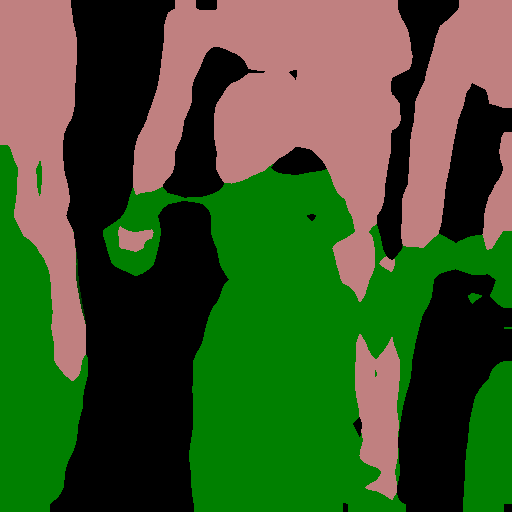} & 
   \includegraphics[width=\imgsizebis]{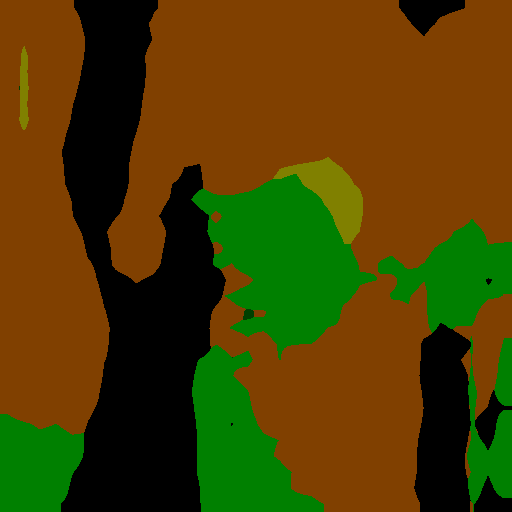} & 
   \includegraphics[width=\imgsizebis]{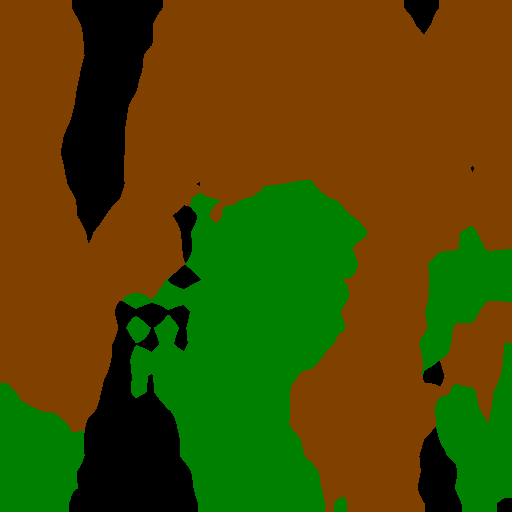} &
   \includegraphics[width=\imgsizebis]{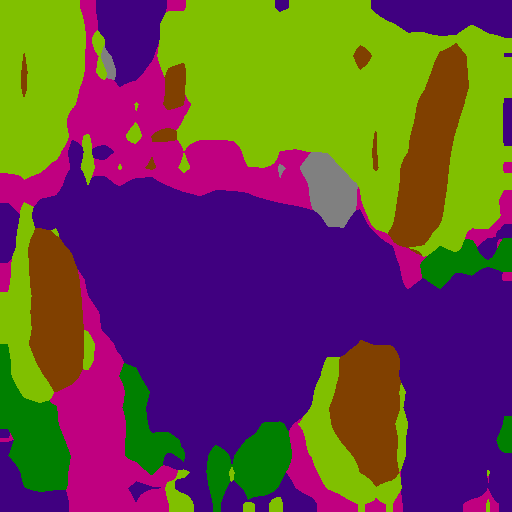} &
   \includegraphics[width=\imgsizebis]{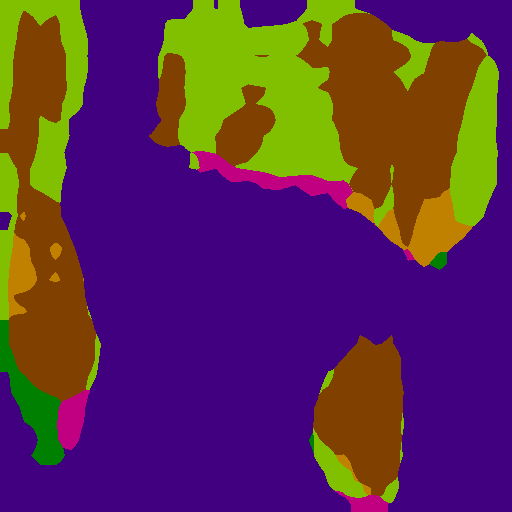} \\

     & & & & & & &  \\
\end{tabular}
\caption{Qualitative results on sample scenes in the disjoint experimental protocol 15-1 on Pascal VOC 2012 during  the various incremental steps (\textit{best viewed in colors}).}
\label{fig:voc_qualitative_varying}
\end{figure*}

\begin{figure*}[htbp]
\setlength{\tabcolsep}{0.7pt} 
\renewcommand{\arraystretch}{0.6}
\centering
\begin{tabular}{ccccccccccc}

         \rotatebox{90}{\hspace{+2ex}SDR (ours)} &  
   \includegraphics[width=\imgsizebis]{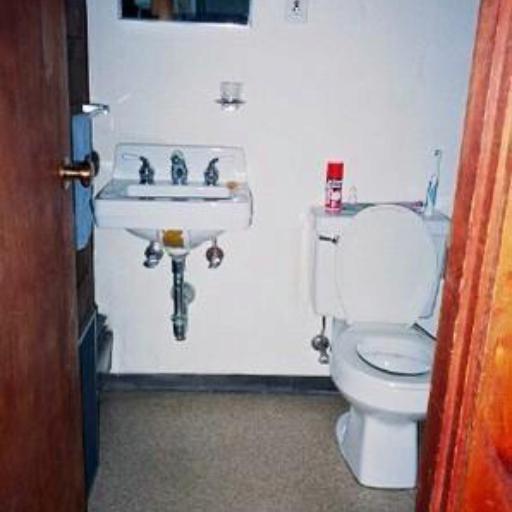} &
   \includegraphics[width=\imgsizebis]{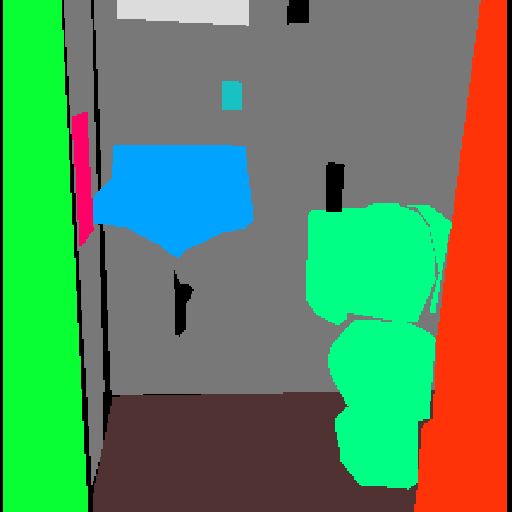} & 
   \includegraphics[width=\imgsizebis]{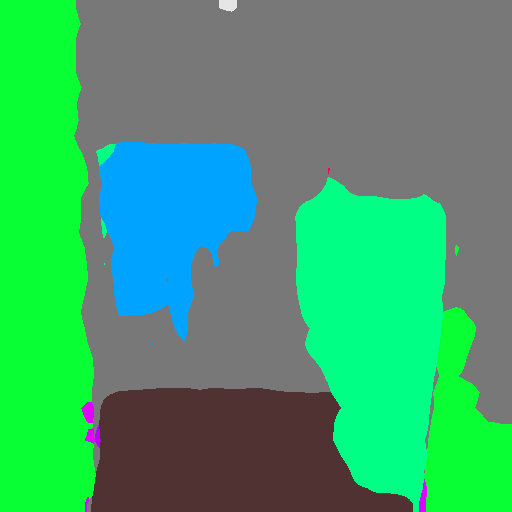} & 
   \includegraphics[width=\imgsizebis]{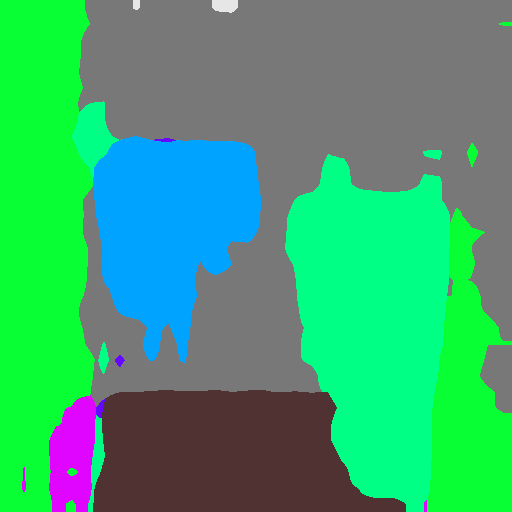} & 
   \includegraphics[width=\imgsizebis]{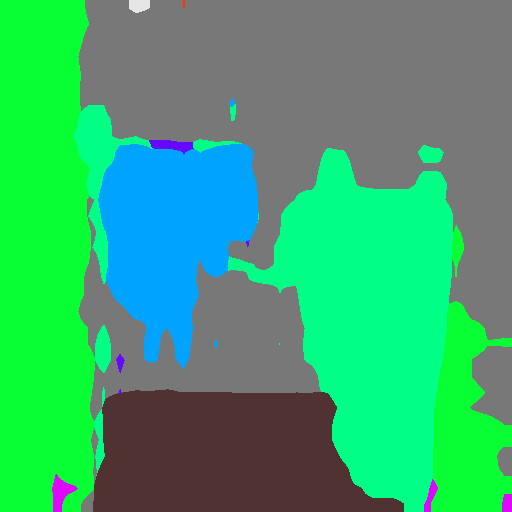} &
   \includegraphics[width=\imgsizebis]{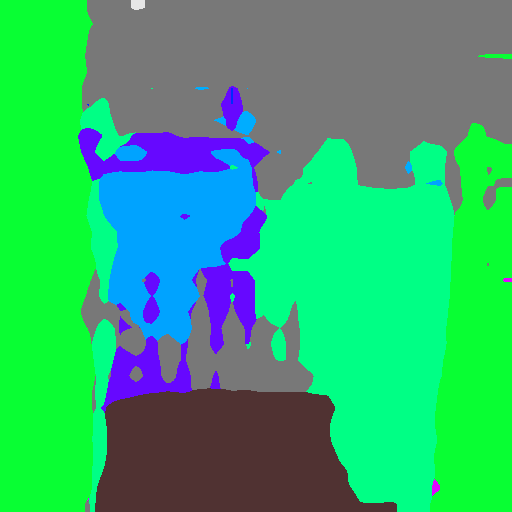} &
   \includegraphics[width=\imgsizebis]{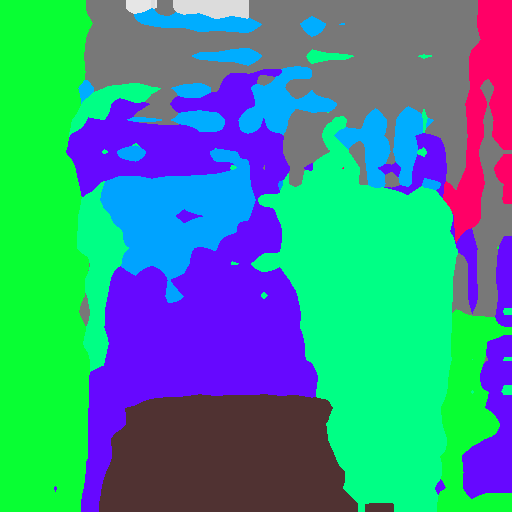} \\ 
   
      \rotatebox{90}{\hspace{+4ex}MiB \cite{cermelli2020modeling}} &  
   \includegraphics[width=\imgsizebis]{img/suppl/ade/increasing_steps/100-10/ADE_val_00000044_RGB.jpg} &
   \includegraphics[width=\imgsizebis]{img/suppl/ade/increasing_steps/100-10/ADE_val_00000044_label.png} & 
   \includegraphics[width=\imgsizebis]{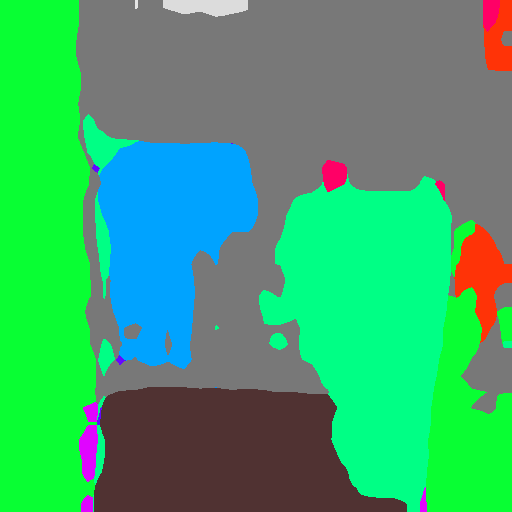} & 
   \includegraphics[width=\imgsizebis]{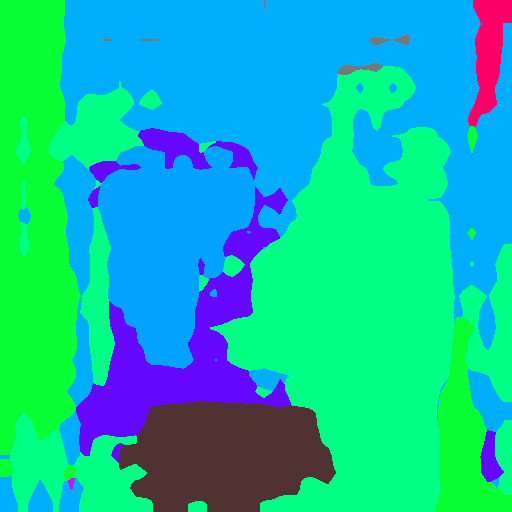} & 
   \includegraphics[width=\imgsizebis]{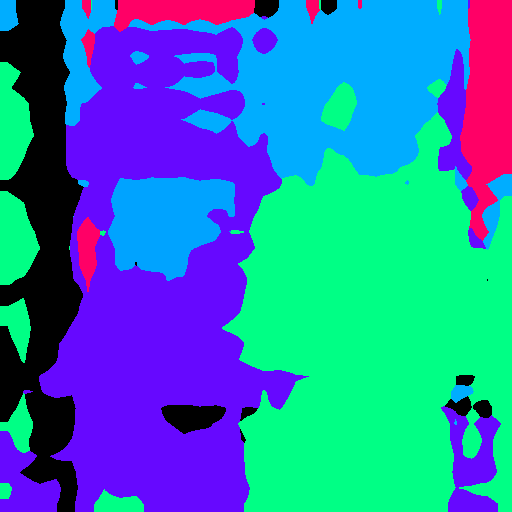} &
   \includegraphics[width=\imgsizebis]{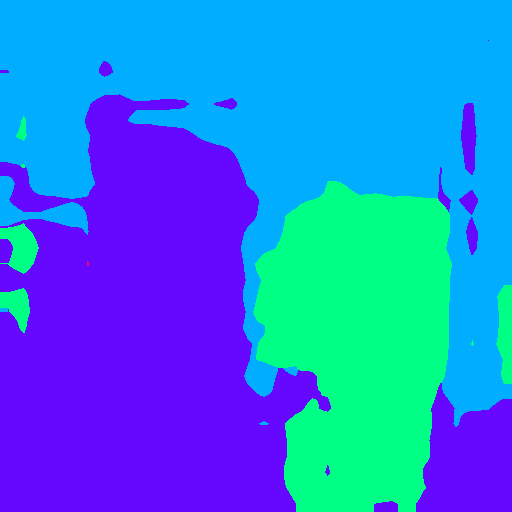} &
   \includegraphics[width=\imgsizebis]{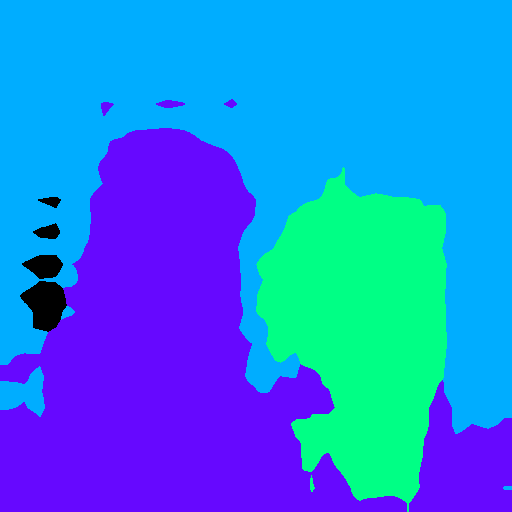} \\ 
   
     & & & & & & &  \\
\end{tabular}
\caption{Qualitative results on sample scenes in experimental protocol 100-10 on ADE20K during  the various incremental steps (\textit{best viewed in colors}).}
\label{fig:ade_100_10_qualitative_varying}
\end{figure*}

\newcommand{\imgsizetris}{23.5mm}
\begin{figure*}[htbp]
\setlength{\tabcolsep}{0.7pt} 
\renewcommand{\arraystretch}{0.6}
\centering
\begin{tabular}{cccccccc}
  
   \rotatebox{90}{\hspace{+2ex}SDR (ours)} &  
   \includegraphics[width=\imgsizetris]{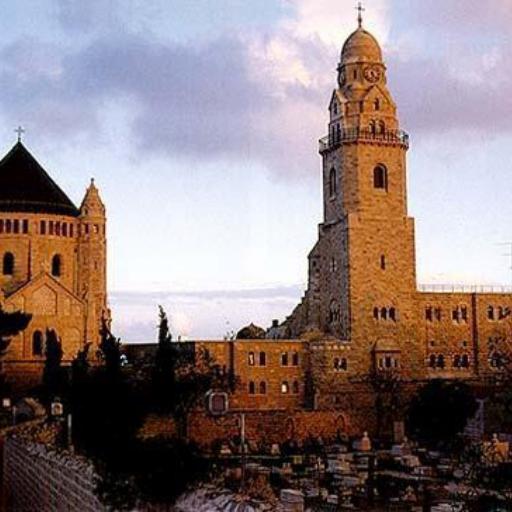} &
   \includegraphics[width=\imgsizetris]{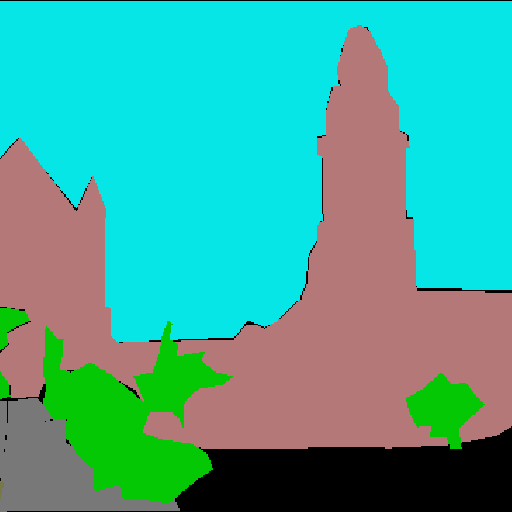} & 
   \includegraphics[width=\imgsizetris]{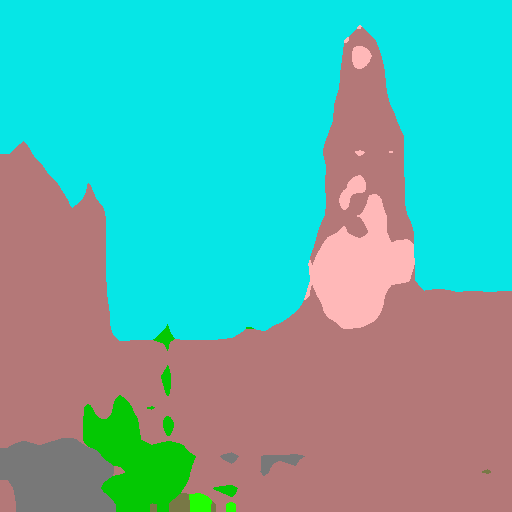} & 
   \includegraphics[width=\imgsizetris]{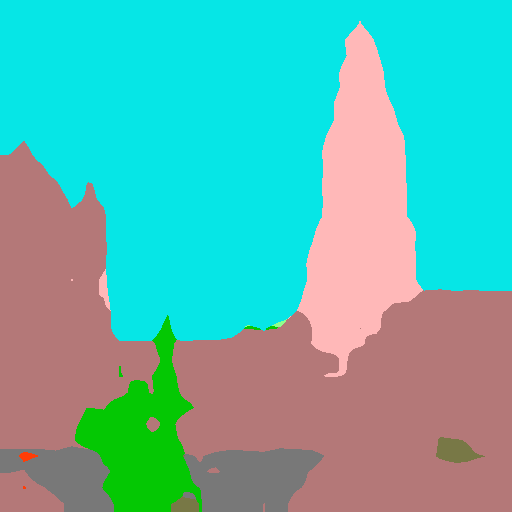} \\ 
   
      \rotatebox{90}{\hspace{+4ex}MiB \cite{cermelli2020modeling}} &   
      \includegraphics[width=\imgsizetris]{img/suppl/ade/increasing_steps/50/ADE_val_00000002_RGB.jpg} &
      \includegraphics[width=\imgsizetris]{img/suppl/ade/increasing_steps/50/ADE_val_00000002_label.png} & 
      \includegraphics[width=\imgsizetris]{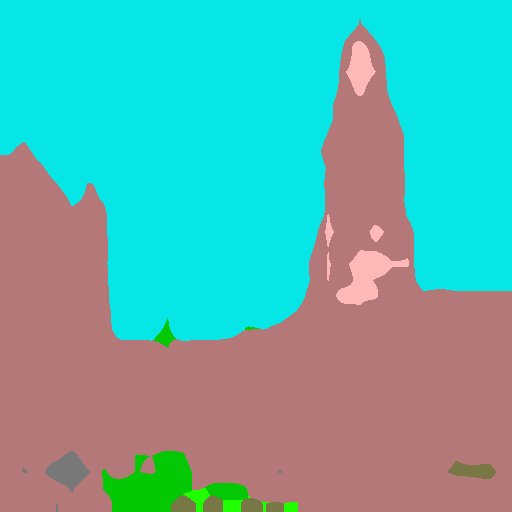} & 
      \includegraphics[width=\imgsizetris]{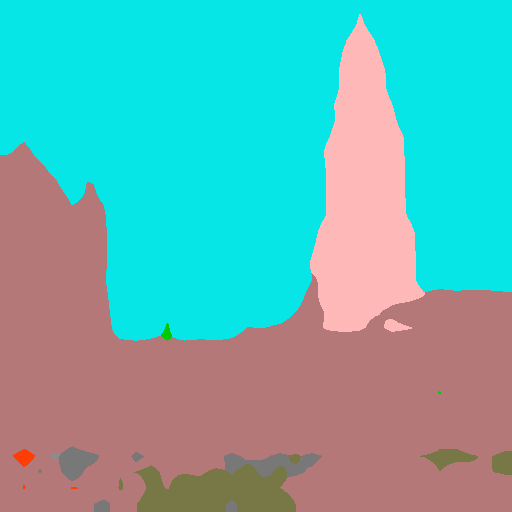} \\ 
      
        & & & & & & & \\
      
         \rotatebox{90}{\hspace{+2ex}SDR (ours)} &  
   \includegraphics[width=\imgsizetris]{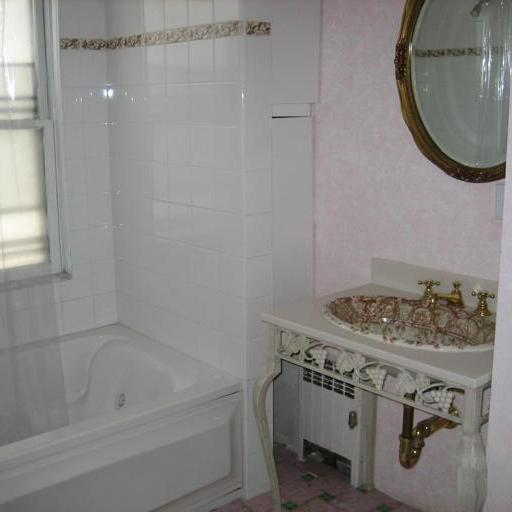} &
   \includegraphics[width=\imgsizetris]{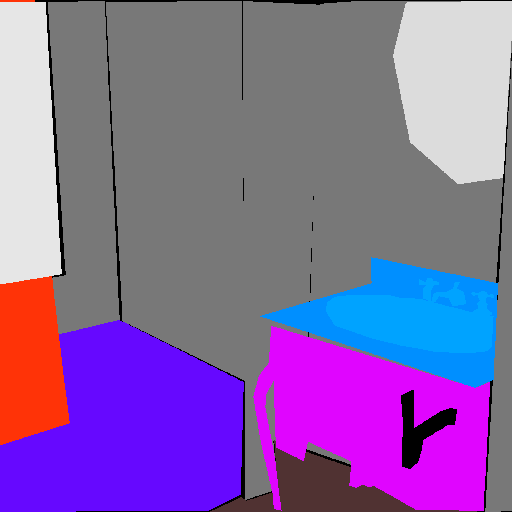} & 
   \includegraphics[width=\imgsizetris]{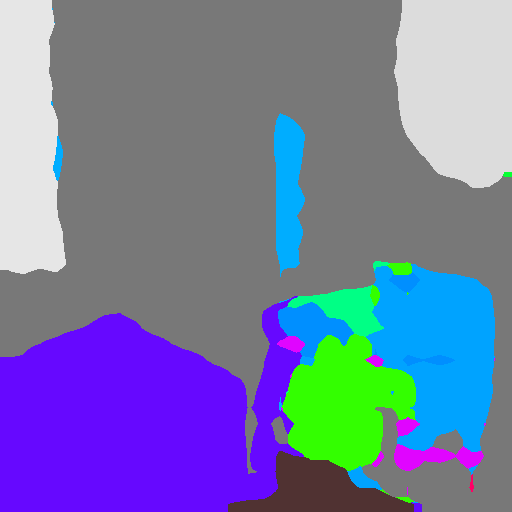} & 
   \includegraphics[width=\imgsizetris]{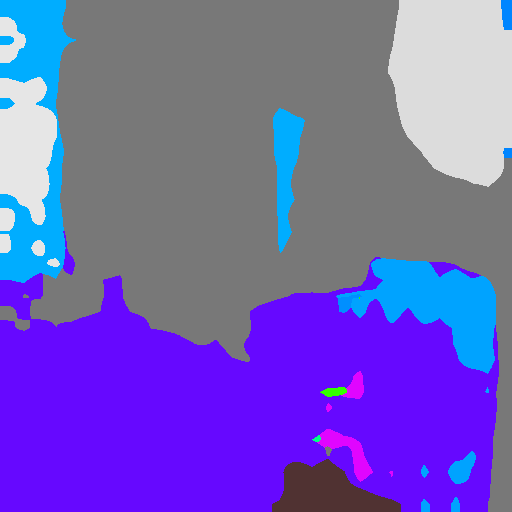} \\
\end{tabular}
\caption{Qualitative results on sample scenes in experimental protocol 50-50 on ADE20K during the various  incremental steps (\textit{best viewed in colors}).}
\label{fig:ade_50_50_qualitative_varying}
\end{figure*}

\clearpage

\begin{table*}[h]
  \centering
  \small
  \setlength{\tabcolsep}{1.1pt} 
  \caption{Per-class IoU of compared methods in disjoint experimental protocol on scenario 19-1 of Pascal VOC 2012.}
    \begin{tabular}{|l|cccccccccccccccccccc:c|ccc|}
    \hline
       Method  & \rotatebox{90}{backgr.} &  \rotatebox{90}{aero} &  \rotatebox{90}{bike} &  \rotatebox{90}{bird} &\rotatebox{90}{boat} & \rotatebox{90}{bottle} & \rotatebox{90}{bus}   &\rotatebox{90}{car} & \rotatebox{90}{cat} & \rotatebox{90}{chair} & \rotatebox{90}{cow} & \rotatebox{90}{din. table}& \rotatebox{90}{dog} & \rotatebox{90}{horse}   & \rotatebox{90}{mbike} & \rotatebox{90}{person} & \rotatebox{90}{plant} &  \rotatebox{90}{sheep} & \rotatebox{90}{sofa} & \rotatebox{90}{train} & \rotatebox{90}{tv} & old   & new   & all \\\hline
    FT    & 72.4 & 62.4 & 6.7 & 45.0 & 47.1 & 39.5 & 33.7 & 40.9 & 25.7 & 4.3 & 54.0 & 8.0 & 25.0 & 50.4 & 50.6 & 0.0 & 35.3 & 43.0 & 0.8 & 59.5 & 13.2 & 35.2 & 13.2 & 34.2 \\\hdashline
    LwF \cite{li2018learning}  &  87.6 & 75.4 & 31.1 & 71.7 & 50.8 & 66.0 & 81.6 & 79.0 & 87.9 & 32.1 & 66.9 & 49.9 & 84.1 & 66.2 & 77.3 & 79.4 & 51.8 & 68.5 & 42.1 & 65.8 & 28.3 & 65.8 & 28.3 & 64.0 \\
    LwF-MC \cite{rebuffi2017icarl} & 78.6 & 63.6 & 0.4 & 61.2 & 10.6 & 35.2 & 52.8 & 35.1 & 75.5 & 0.4 & 63.9 & 1.5 & 75.5 & 67.8 & 32.6 & 13.1 & 13.0 & 63.4 & 0.7 & 25.9 & 1.0 & 38.5 & 1.0 & 36.7 \\
\hline
    \end{tabular}%
  \label{tab:voc_19_1}%
\end{table*}%

\begin{table*}[htbp]
  \centering
  \small
  \setlength{\tabcolsep}{1.1pt} 
  \caption{Per-class pixel accuracy of compared methods in disjoint experimental protocol on scenario 19-1 of Pascal VOC 2012.}
    \begin{tabular}{|l|cccccccccccccccccccc:c|ccc|}
    \hline
       Method  & \rotatebox{90}{backgr.} &  \rotatebox{90}{aero} &  \rotatebox{90}{bike} &  \rotatebox{90}{bird} &\rotatebox{90}{boat} & \rotatebox{90}{bottle} & \rotatebox{90}{bus}   &\rotatebox{90}{car} & \rotatebox{90}{cat} & \rotatebox{90}{chair} & \rotatebox{90}{cow} & \rotatebox{90}{din. table}& \rotatebox{90}{dog} & \rotatebox{90}{horse}   & \rotatebox{90}{mbike} & \rotatebox{90}{person} & \rotatebox{90}{plant} &  \rotatebox{90}{sheep} & \rotatebox{90}{sofa} & \rotatebox{90}{train} & \rotatebox{90}{tv} & old   & new   & all \\\hline
    FT    & 91.5 & 79.9 & 7.2 & 74.9 & 71.1 & 44.0 & 34.3 & 46.4 & 26.1 & 4.5 & 72.6 & 8.1 & 25.4 & 78.0 & 53.9 & 0.0 & 40.6 & 58.5 & 0.8 & 64.3 & 82.0 & 35.2 & 13.2 & 34.2 \\\hdashline
    LwF \cite{li2018learning}  & 94.1 & 85.6 & 58.7 & 91.2 & 59.1 & 76.3 & 84.4 & 80.3 & 94.1 & 39.3 & 93.5 & 52.3 & 91.7 & 95.3 & 84.0 & 82.3 & 76.5 & 84.1 & 48.2 & 68.4 & 69.6 & 65.8 & 28.3 & 64.0 \\
    LwF-MC \cite{rebuffi2017icarl} & 99.8 & 65.5 & 0.4 & 63.1 & 10.7 & 39.6 & 53.1 & 35.3 & 78.4 & 0.5 & 66.5 & 1.5 & 77.8 & 72.0 & 34.0 & 13.1 & 14.4 & 65.9 & 0.7 & 25.9 & 1.0 & 38.5 & 1.0 & 36.7 \\
\hline
    \end{tabular}%
  \label{tab:voc_19_1_PA}%
\end{table*}%

\begin{table*}[htbp]
  \centering
  \small
  \setlength{\tabcolsep}{1.1pt} 
  \caption{Per-class IoU of compared methods in disjoint experimental protocol on scenario 15-5 of Pascal VOC 2012.}
    \begin{tabular}{|l|cccccccccccccccc:ccccc|ccc|}
    \hline
       Method  & \rotatebox{90}{backgr.} &  \rotatebox{90}{aero} &  \rotatebox{90}{bike} &  \rotatebox{90}{bird} &\rotatebox{90}{boat} & \rotatebox{90}{bottle} & \rotatebox{90}{bus}   &\rotatebox{90}{car} & \rotatebox{90}{cat} & \rotatebox{90}{chair} & \rotatebox{90}{cow} & \rotatebox{90}{din. table}& \rotatebox{90}{dog} & \rotatebox{90}{horse}   & \rotatebox{90}{mbike} & \rotatebox{90}{person} & \rotatebox{90}{plant} &  \rotatebox{90}{sheep} & \rotatebox{90}{sofa} & \rotatebox{90}{train} & \rotatebox{90}{tv} & old   & new   & all \\\hline
    FT    & 74.2 & 27.2 & 0.0 & 1.6 & 15.1 & 11.3 & 0.0 & 4.1 & 0.5 & 0.0 & 0.0 & 0.0 & 0.0 & 0.2 & 0.2 & 0.0 & 27.0 & 25.6 & 28.9 & 33.5 & 52.2 & 8.4 & 33.5 & 14.4 \\\hdashline
    LwF \cite{li2018learning}  & 83.4 & 59.1 & 21.7 & 16.7 & 36.8 & 47.0 & 18.7 & 62.5 & 52.3 & 6.6 & 4.8 & 37.7 & 35.9 & 44.9 & 55.5 & 51.6 & 22.6 & 27.8 & 25.3 & 39.6 & 51.1 & 39.7 & 33.3 & 38.2 \\
    LwF-MC \cite{rebuffi2017icarl} & 85.4 & 54.2 & 16.9 & 59.7 & 29.7 & 46.0 & 34.4 & 65.9 & 38.1 & 5.2 & 35.9 & 7.5 & 62.4 & 44.3 & 48.7 & 29.1 & 11.4 & 37.3 & 8.9 & 42.1 & 27.1 & 41.5 & 25.4 & 37.6 \\\hdashline
    ILT  \cite{michieli2019incremental} & 81.7 & 47.6 & 18.4 & 1.6 & 29.7 & 19.4 & 3.8 & 52.5 & 56.7 & 0.5 & 4.6 & 20.7 & 43.1 & 35.4 & 33.6 & 54.8 & 22.7 & 22.4 & 15.9 & 30.1 & 34.3 & 31.5 & 25.1 & 30.0 \\
\hline
    \end{tabular}%
  \label{tab:voc_15_5}%
\end{table*}%

\clearpage

\begin{table*}[htbp]
  \centering
  \small
  \setlength{\tabcolsep}{1.1pt} 
  \caption{Per-class pixel accuracy of compared methods in disjoint experimental protocol on scenario 15-5 of Pascal VOC 2012.}
    \begin{tabular}{|l|cccccccccccccccc:ccccc|ccc|}
    \hline
       Method  & \rotatebox{90}{backgr.} &  \rotatebox{90}{aero} &  \rotatebox{90}{bike} &  \rotatebox{90}{bird} &\rotatebox{90}{boat} & \rotatebox{90}{bottle} & \rotatebox{90}{bus}   &\rotatebox{90}{car} & \rotatebox{90}{cat} & \rotatebox{90}{chair} & \rotatebox{90}{cow} & \rotatebox{90}{din. table}& \rotatebox{90}{dog} & \rotatebox{90}{horse}   & \rotatebox{90}{mbike} & \rotatebox{90}{person} & \rotatebox{90}{plant} &  \rotatebox{90}{sheep} & \rotatebox{90}{sofa} & \rotatebox{90}{train} & \rotatebox{90}{tv} & old   & new   & all \\\hline
    FT    & 95.3 & 27.5 & 0.0 & 1.6 & 15.4 & 11.5 & 0.0 & 4.1 & 0.5 & 0.0 & 0.0 & 0.0 & 0.0 & 0.2 & 0.2 & 0.0 & 72.0 & 90.0 & 77.2 & 89.7 & 80.7 & 8.4 & 33.5 & 14.4 \\\hdashline
    LwF \cite{li2018learning}  & 91.9 & 79.4 & 35.4 & 16.9 & 50.9 & 49.0 & 19.4 & 71.0 & 78.8 & 8.0 & 5.2 & 39.7 & 36.3 & 78.5 & 59.2 & 53.3 & 67.1 & 91.2 & 74.2 & 81.6 & 76.5 & 39.7 & 33.3 & 38.2 \\
    LwF-MC \cite{rebuffi2017icarl} & 96.6 & 80.7 & 30.3 & 68.5 & 62.0 & 60.4 & 37.7 & 79.7 & 62.5 & 10.8 & 46.2 & 9.2 & 73.2 & 84.4 & 64.8 & 31.7 & 11.4 & 39.7 & 9.1 & 60.1 & 27.1 & 41.5 & 25.4 & 37.6 \\\hdashline
    ILT  \cite{michieli2019incremental} & 94.6 & 61.4 & 26.4 & 1.6 & 30.8 & 19.5 & 4.0 & 57.7 & 71.7 & 0.5 & 5.1 & 20.9 & 45.9 & 43.7 & 34.6 & 56.7 & 42.5 & 86.0 & 38.8 & 71.0 & 44.9 & 31.5 & 25.1 & 30.0 \\
\hline
    \end{tabular}%
  \label{tab:voc_15_5_PA}%
\end{table*}%

\begin{table*}[htbp]
  \centering
  \small
  \setlength{\tabcolsep}{1.1pt} 
  \caption{Per-class IoU of compared methods in disjoint experimental protocol on scenario 15-1 of Pascal VOC 2012.}
    \begin{tabular}{|l|cccccccccccccccc:c:c:c:c:c|ccc|}
    \hline
       Method  & \rotatebox{90}{backgr.} &  \rotatebox{90}{aero} &  \rotatebox{90}{bike} &  \rotatebox{90}{bird} &\rotatebox{90}{boat} & \rotatebox{90}{bottle} & \rotatebox{90}{bus}   &\rotatebox{90}{car} & \rotatebox{90}{cat} & \rotatebox{90}{chair} & \rotatebox{90}{cow} & \rotatebox{90}{din. table}& \rotatebox{90}{dog} & \rotatebox{90}{horse}   & \rotatebox{90}{mbike} & \rotatebox{90}{person} & \rotatebox{90}{plant} &  \rotatebox{90}{sheep} & \rotatebox{90}{sofa} & \rotatebox{90}{train} & \rotatebox{90}{tv} & old   & new   & all \\\hline
    FT    & 70.4& 5.5& 0.0& 5.9& 5.2& 0.5& 0.2& 1.6& 0.4& 0.0& 3.3& 0.0& 0.0& 0.2& 0.0& 0.0 & 0.1& 0.0& 0.0& 9.4& 14.8& 5.8& 4.9& 5.6\\\hdashline
    LwF \cite{li2018learning}  & 77.1& 12.0& 6.9& 52.6& 14.3& 23.1& 18.4& 27.3& 56.3& 20.5& 48.9& 8.3& 17.8& 12.6& 15.6& 8.3& 0.0& 17.0& 21.0& 18.6& 19.1& 26.2& 15.1& 23.6\\
    LwF-MC \cite{rebuffi2017icarl} & 69.5& 0.1& 0.0& 8.0& 0.1& 7.2& 0.0& 0.1& 8.1& 0.0& 6.6& 0.0& 8.0& 1.7& 0.3& 0.1& 0.0& 0.0& 0.0& 2.4& 8.1& 6.9& 2.1& 5.7\\\hdashline
    ILT \cite{michieli2019incremental}  & 69.4& 0.0& 2.1& 0.0& 0.0& 0.1& 0.0& 4.0& 0.0& 0.0& 0.0& 0.0& 1.4& 0.0& 0.0& 19.2& 0.0& 0.0& 0.0& 1.4& 4.6& 6.7& 1.2& 5.4\\
    CIL \cite{klingner2020class}  & 78.4& 2.4& 23.6& 47.9& 4.6& 32.9& 0.3& 29.9& 45.4& 15.4& 30.3& 2.4& 54.5& 13.0& 8.7& 59.7& 15.2& 17.5& 12.1& 20.9& 19.2& 33.3& 15.9& 29.1\\
\hline
    \end{tabular}%
  \label{tab:voc_15_1}%
\end{table*}%

\begin{table*}[htbp]
  \centering
  \small
  \setlength{\tabcolsep}{1.1pt} 
  \caption{Per-class pixel accuracy of compared methods in disjoint experimental protocol on scenario 15-1 of Pascal VOC 2012.}
    \begin{tabular}{|l|cccccccccccccccc:c:c:c:c:c|ccc|}
    \hline
       Method  & \rotatebox{90}{backgr.} &  \rotatebox{90}{aero} &  \rotatebox{90}{bike} &  \rotatebox{90}{bird} &\rotatebox{90}{boat} & \rotatebox{90}{bottle} & \rotatebox{90}{bus}   &\rotatebox{90}{car} & \rotatebox{90}{cat} & \rotatebox{90}{chair} & \rotatebox{90}{cow} & \rotatebox{90}{din. table}& \rotatebox{90}{dog} & \rotatebox{90}{horse}   & \rotatebox{90}{mbike} & \rotatebox{90}{person} & \rotatebox{90}{plant} &  \rotatebox{90}{sheep} & \rotatebox{90}{sofa} & \rotatebox{90}{train} & \rotatebox{90}{tv} & old   & new   & all \\\hline
    FT    & 98.5 & 5.6 & 0.0 & 5.9 & 5.4 & 0.5 & 0.2 & 1.6 & 0.4 & 0.0 & 3.4 & 0.0 & 0.0 & 0.2 & 0.0 & 0.0 & 0.1 & 0.0 & 0.0 & 9.8 & 80.1 & 5.8 & 4.9 & 5.6 \\\hdashline
    LwF \cite{li2018learning}  & 95.1 & 12.0 & 47.8 & 54.0 & 15.0 & 23.2 & 18.4 & 27.4 & 57.3 & 35.1 & 62.8 & 8.3 & 18.0 & 12.7 & 15.7 & 8.3 & 0.0 & 22.0 & 35.8 & 47.9 & 70.7 & 26.2 & 15.1 & 23.6 \\
    LwF-MC \cite{rebuffi2017icarl} & 99.9 & 0.1 & 0.0 & 8.0 & 0.1 & 7.4 & 0.0 & 0.1 & 8.1 & 0.0 & 6.6 & 0.0 & 8.0 & 1.7 & 0.3 & 0.1 & 0.0 & 0.0 & 0.0 & 2.9 & 8.2 & 6.9 & 2.1 & 5.7 \\\hdashline
    ILT \cite{michieli2019incremental}  & 20.9 & 0.0 & 73.2 & 0.0 & 0.0 & 0.0 & 2.3 & 89.3 & 19.0 & 16.3 & 14.8 & 1.4 & 48.3 & 0.0 & 23.2 & 0.4 & 4.6 & 0.0 & 0.0 & 1.8 & 4.9 & 6.7 & 1.2 & 5.4 \\
\hline
    \end{tabular}%
  \label{tab:voc_15_1_PA}%
\end{table*}%

\end{document}